\documentclass{article}

\usepackage{arxiv}

\usepackage[utf8]{inputenc}
\usepackage[T1]{fontenc}
\usepackage{amsmath}
\usepackage{amssymb}
\usepackage{amsfonts}
\usepackage{mathtools}
\usepackage{booktabs}
\usepackage{array}
\usepackage[table]{xcolor}
\usepackage{graphicx}
\usepackage{placeins}
\usepackage{microtype}
\usepackage{siunitx}
\usepackage{hyperref}
\usepackage{cleveref}
\usepackage[numbers,sort&compress]{natbib}
\usepackage{url}

\title{From Preimage Search to Source-Grounded Feature Inversion%
\thanks{Code: \url{https://github.com/k-Jayus/source-grounded-feature-inversion}}}

\author{
  Kaixiang Shu\\
  Independent Researcher\\
  \texttt{614729197@qq.com}
}

\renewcommand{\shorttitle}{Source-Grounded Feature Inversion}

\hypersetup{
hidelinks,
pdftitle={From Preimage Search to Source-Grounded Feature Inversion},
pdfsubject={Machine Learning},
pdfauthor={Kaixiang Shu},
pdfkeywords={feature inversion, visualisation, Wiener filtering, computational graphs, closed-form learning}
}


\newcommand{\R}{\mathbb{R}}
\newcommand{\CC}{\mathbb{C}}
\newcommand{\E}{\mathbb{E}}
\newcommand{\calD}{\mathcal{D}}
\newcommand{\calF}{\mathcal{F}}
\newcommand{\calG}{\mathcal{G}}
\newcommand{\calE}{\mathcal{E}}
\newcommand{\calV}{\mathcal{V}}
\newcommand{\Rsrc}{R}
\newcommand{\Hv}{H}
\newcommand{\Gv}{G}
\newcommand{\SHR}{S^{HR}}
\newcommand{\SRR}{S^{RR}}

\begin{document}
\maketitle

\begin{abstract}
Interpreting a neural network requires understanding what its internal features
extract from a particular input.  Feature inversion seeks to express a selected
feature in the input domain, but canonical iterative methods search for an
input whose re-encoded representation matches the target.  Because many inputs
can satisfy this constraint, target matching alone does not specify the inverse
associated with the sample that generated the feature.  We therefore recast
feature inversion from target-matching preimage search to source-grounded
upstream-state estimation along the target-generating computational DAG,
thereby conditioning the inverse on the source-local network geometry.  At
each boundary of the computational DAG, backpropagation provides the correct
reverse dependencies but transports an adjoint signal rather than an
upstream-state estimate.  We locally repair this signal with a closed-form
matrix Wiener map from a mean-seed VJP to an upstream-state estimate.  A second
Wiener map corrects the JVP forward-consistency residual, and the repaired
states are composed through the same DAG in one finite reverse pass.  After
calibration, the resulting zero-intercept map families support new inputs,
depths, and tested channel--coordinate queries without query-specific
optimisation.
Across CNNs and vision Transformers, the same calibrated formulation spans
diverse tensor components and visual distributions, while target--operator
controls show that architecture-dependent inverses are jointly determined by
the selected feature and source-local operators.  Beyond vision, for
terminal-token queries in GPT-2 small, the same source-grounded closed-form
construction produces token-resolved continuous input-embedding inverses
without a text decoder.  At the exploratory layer-6 contextual peak, the
resulting token-position energy profiles show a cue-minus-random effect of
0.190 (95\% word-cluster bootstrap interval [0.096, 0.297], conditional on the
selected layer); relative to Raw VJP, Final improves
agreement-minus-attractor routing by 0.046 (95\% template-cluster bootstrap
interval [0.027, 0.066] under the same layer condition).  Across
both modalities, prediction-conditioned feature atlases pair inversion with
independent interventions on the corresponding internal features, linking the
revealed input-domain structures to the model's current decision without
treating an inverse as a causal stimulus.  Together, these results establish
source-grounded feature inversion as a reusable, modality-native interface for
revealing what selected internal representations identify in the input domain
and how the corresponding evidence shapes model decisions.
\end{abstract}


\section{Introduction}
\label{sec:introduction}

Modern neural networks form predictions through a hierarchy of internal
representations, yet these representations remain difficult to inspect.  A
tensor activation records a state in feature space, but it does not directly
show which structures that state extracts from a particular input.  Output
explanations and sensitivity maps can identify influential input regions, but
they do not provide an input-domain view of a selected internal state at an
arbitrary depth or channel.  A representation-level interpretation should
therefore answer a sample-specific question: what does this feature, at this
layer and channel, extract from this input?  Expressing internal features in the
input domain would place heterogeneous layers and architectures in a common
coordinate system and make their progression through the network observable.

Feature inversion provides a natural interface for this purpose.  Given a
frozen model $f$, an input $x$, a target node $T$, a selected channel set $S$,
and a selected coordinate set $Q$, we seek an input-domain state associated
with the masked target feature $P_{S,Q}H_T(x)$.  This state is intended to
reveal the structure identified by the selected feature in the computation
generated by $x$.  It is not posed as
reconstruction of the original input from a compressed code, and similarity to
$x$ is not the definition of correctness.  The source input specifies the
forward operating point; it is not treated as a unique inverse label.  The
desired inverse is therefore sample-specific: it should explain the selected
feature as it occurs in the forward computation of the input being interpreted.

Canonical iterative inversion instead poses a target-preimage problem of the
form \citep{mahendran2015understanding,olah2017feature}
\[
  z^\star
  =
  \arg\min_z
  d\bigl(H_T(z),H_T(x)\bigr)+\mathcal R(z).
\]
The optimiser searches for an input whose re-encoded representation matches the
target.  When the representation is non-injective, many inputs can satisfy this
constraint, and the returned preimage depends on the initialisation,
regularisation, and input prior used to select among them.  Although the frozen
network participates in this search, its differential operators are evaluated
along the candidate trajectory rather than used to condition the inverse on the
forward point that generated $H_T(x)$.  Learned decoders and gradient
visualisations address related objectives
\citep{dosovitskiy2016inverting,nash2019inverting,simonyan2013deep,
zeiler2014visualizing,springenberg2014striving,selvaraju2017grad}, but they
respectively introduce a separate learned mapping or expose an adjoint signal
rather than a sample-specific upstream state.  The missing condition is
therefore not another input prior, regulariser, or learned decoder, but the
local computation that produced the selected feature.

An internal feature does not exist independently of that computation.
The input $x$ determines the forward activations, nonlinear operating regions,
branch states, attention routes, and local Jacobians along the computational
DAG that produces $H_T(x)$.  We therefore recast feature inversion from
target-matching preimage search to source-grounded upstream-state estimation
along the target-generating computational DAG, thereby conditioning the inverse
on the source-local network geometry.  Accordingly, we treat a
feature-inversion query as $(f,x,T,S,Q)$ rather than as a target tensor alone.  We
call the resulting object \emph{source-grounded feature inversion}: the selected
feature is traced through the source-local network geometry at the
target-generating forward point.  This input conditioning does not provide the
answer through an input shortcut.  The input is used only to instantiate the
frozen forward activations and local differential operators, while the reverse
recursion is seeded by the selected terminal state.  The output is an estimated
sequence of upstream states along this sample-specific local inverse branch.
This definition turns the problem into a sequence of local upstream-state
estimation tasks on the same DAG that generated the feature.  The formulation
is defined on frozen finite differentiable computational DAGs exposed by
autograd.

Computing these local inverses requires more than raw backpropagation.  A VJP
follows the correct reverse dependencies, but it transports an adjoint signal
$J^*q$ rather than an upstream-state estimate.  Under parent and child metrics
$M_v$ and $M_u$, the corresponding natural adjoint is
$J^\sharp=M_v^{-1}J^*M_u$.  Under the centred local linear-Gaussian reference
model, covariance-induced metrics make this operator coincide with local Wiener
inversion.  In a nonlinear network, however, the
relevant metrics and local Jacobians depend on the boundary and source operating
point.  Explicit online estimation of $M_v$ and $M_u$ is therefore impractical.
We instead use a mean-seed VJP as an observable and calibrate a zero-intercept
full-channel matrix Wiener map $G_v$ from this observable to the upstream
activation.  This first projection amortises the missing metric
correction over the calibration distribution while retaining the source-local
Jacobian in its VJP input.

The first projection can capture the dominant correction when the local
operator is fixed or weakly sample dependent.  Normalisation, attention, and
data-dependent gating or routing can induce stronger sample dependence, for
which a shared $G_v$ can leave an operating-point-specific mismatch.  A JVP
evaluates the first-stage estimate at the actual source forward point, and a
second Wiener map $D_v$ predicts the remaining state error from the pulled-back
child-consistency residual.  The repaired states are composed through the
selected boundary DAG in one finite reverse pass, as summarised in
Figure~\ref{fig:method_paradigm}.  The theoretical development connects the
MSE-optimal conditional mean to the covariance-induced natural adjoint, the
deployed distributional projection, and finite-DAG error propagation.  Once
calibrated, the same local map family supports new inputs, depths, and tested
channel--coordinate queries without query-specific optimisation.

Across diverse CNN and vision Transformer families, tensor components, and
visual distributions, the resulting inverses reveal architecture-specific
spatial and frequency structure.  The outputs do not collapse to an image
template: zero targets yield zero inverses, channel selections produce distinct
structures, and operating-point swaps redirect the inverse to the new
source-local branch.
These controls show that each visualisation is jointly determined by the
selected feature and the local geometry that generated it.  Layer-wise atlases
reveal architecture-dependent progressions of extracted structure, while
channel-wise queries separate distinct internal evidence within the same
representation.  Coordinate queries provide a complementary test: one ViT
calibration supports all 49 patch queries without refitting, and the resulting
normalised inverse-magnitude support changes from locally aligned to distributed
with depth.  Independent internal interventions distinguish this descriptive
support from the decision effect of the corresponding patch-token states.  For
prediction-conditioned groups, top-positive feature groups
produce larger prediction-margin drops than matched random controls across all
48 tested model--layer--group-size settings.  Independent intervention
therefore grounds the visualised feature groups in the current decision without
treating the inverse image as a causal stimulus.  These results expose what
individual layers and channels extract from an input and which internal feature
groups shape the model's decision.

Beyond vision, we test the same closed-form construction in a controlled GPT-2
small terminal-token setting \citep{radford2019language}.  A calibrated
source-local reverse path maps a selected attention-output representation to a
token-resolved continuous input-embedding inverse without a text decoder.
Controlled semantic and number-agreement queries, together with
source-geometry, zero-target, and reproducibility controls, show that the
inverse exposes context-specific evidence carried by the selected
representation.  Independent interventions connect the corresponding internal
feature groups to GPT-2's current decision.  This provides constructive
evidence that source-grounded feature inversion is not intrinsically tied to
vision.

The contributions of this work are threefold:
\begin{itemize}
  \item We recast feature inversion from target-matching preimage search to
  source-grounded upstream-state estimation along a target-generating finite
  differentiable computational DAG, making its source-local geometry part of
  the inverse object.
  \item We derive a two-stage local closed-form matrix Wiener construction that
  converts VJP adjoints into upstream-state estimates, corrects predictable
  operating-point-specific residuals, and composes the repaired states through
  the computational DAG in one finite reverse pass.
  \item We validate the same closed-form construction across CNNs, vision
  Transformers, and GPT-2 small, where it produces modality-native inverses
  across tested channel--coordinate queries without query-specific
  optimisation.  Target--operator controls verify source grounding, while
  independent interventions connect the corresponding internal feature groups
  to model decisions.
\end{itemize}
Together, these contributions establish source-grounded feature inversion as a
reusable, modality-native interface for revealing what selected internal
representations identify in the input domain and how the corresponding evidence
shapes model decisions.

\section{Related Work}
\label{sec:related_work}

\subsection{Iterative Preimage Optimisation}

The feature-inversion literature has conventionally formulated inversion as a
preimage search.  Given a target representation, iterative methods optimise an
input whose re-encoded feature matches that target, typically together with
hand-designed natural-image regularisers
\citep{mahendran2015understanding,olah2017feature}.  This formulation is useful
for probing which inputs are compatible with a representation, but the matching
constraint generally defines an equivalence class rather than a unique inverse.
Initialisation, regularisation, and the optimisation trajectory therefore select
one member of that class.  Moreover, the network derivatives used by the
optimiser are evaluated along the evolving candidate input.  We distinguish
this target-consistent preimage search from source-grounded inversion, whose
query also includes the forward operating point that generated the selected
feature and whose reverse computation follows that source-local branch.

\subsection{Learned Reverse Models}

Another line of work learns representation-to-input mappings from paired data.
In vision, convolutional decoders can invert fixed representations, while
autoregressive density models can represent a conditional distribution of
images given supervised features
\citep{dosovitskiy2016inverting,nash2019inverting}.  In language, trained
embedding-inversion attacks partially recover input words from sentence
embeddings, and encoder--decoder correctors can generate, re-embed, and
iteratively revise text from dense embeddings
\citep{song2020information,morris2023text}.  These results show that learned
representations can retain substantial input information, but the recovered
input is produced by an independently trained reverse model and is therefore
jointly determined by the target representation and the decoder's architecture
and data prior.  Such methods also commonly require a separate training problem
for a particular representation family or output domain.  In our formulation,
the conditional mean defines a local upstream-state estimation target.  The
deployed estimator is a closed-form boundary repair conditioned on source-local
VJP and JVP observables, rather than a separately trained end-to-end
representation-to-input decoder.  The repaired states are then composed through
the frozen model's actual DAG.

\subsection{Gradient, Attribution, and Reverse Visualisation}

Gradient, attribution, and reverse-visualisation methods answer a different
family of questions.  Input gradients expose local sensitivity
\citep{simonyan2013deep}; deconvolutional visualisation and Guided
Backpropagation modify the reverse rules used to display activation patterns
\citep{zeiler2014visualizing,springenberg2014striving}; and Grad-CAM aggregates
class gradients into a spatial localisation map
\citep{selvaraju2017grad}.  For language models, attention weights provide an
internal routing reference, but whether they support an explanation depends on
the claim and evaluation criterion
\citep{jain2019attention,wiegreffe2019attention}.  Gradient-times-input and
Integrated Gradients instead produce token-level sensitivity or attribution
scores \citep{ancona2018towards,sundararajan2017axiomatic}.  We use these
signals as interpretability references rather than feature-inversion baselines:
they neither estimate a continuous upstream state for a selected internal
representation nor recursively compose such estimates through the source-local
DAG.  More generally, ordinary backpropagation transports a cotangent through
the adjoint of the local Jacobian rather than estimating the corresponding
upstream forward state.  Sanity checks have also shown that some saliency
procedures can remain visually similar after model or data randomisation,
motivating method-specific tests of model dependence rather than treating visual
sharpness as sufficient evidence \citep{adebayo2018sanity}.  Source-grounded
feature inversion retains the true VJP dependencies, but statistically repairs
the adjoint signal into an upstream-state estimate before continuing through the
graph.

\subsection{Representation Interpretation and Decision Grounding}

Representation-level interpretability has also been pursued by assigning
semantic meaning to units and feature directions.  Network Dissection measures
the alignment between individual units and a labelled concept vocabulary
\citep{bau2017network}; TCAV tests the directional sensitivity of predictions to
user-defined concepts \citep{kim2018interpretability}; and ACE discovers
concepts automatically before measuring their importance
\citep{ghorbani2019towards}.  In language models, probing asks whether linguistic
structure is decodable from hidden representations; structural probing, for
example, fits a supervised map from word representations to syntactic geometry
\citep{hewitt2019structural}.  Intervention and mediation analyses ask the
distinct question of whether an internal component affects model behaviour
\citep{vig2020investigating}.  These evidence types are not interchangeable:
decodability does not establish model use, and an inverse is not itself a causal
intervention.  Target--operator controls in our framework test whether the
inverse depends jointly on the selected feature and source-local geometry.  Our
prediction-conditioned atlases then use a parallel inversion branch to
visualise the input-domain structure selected by a feature group and an
independent intervention branch to measure the original internal group's effect
on the current decision; the inverse is never re-input as a causal stimulus.
The lines reviewed above therefore search for target-consistent inputs, learn
separate representation-to-input mappings, propagate attribution signals, or
analyse the information and causal roles of representations.  Source-grounded
feature inversion instead estimates the input-domain state associated with a
selected feature along the target-generating sample's local computational
geometry, then traces that estimate through the actual DAG using reusable
closed-form repairs.

\section{Method: Closed-Form Matrix Wiener Repair}
\label{sec:method}

We use feature inversion as a representation-level interface for model
interpretation: the goal is to expose which input-domain structures are
identified by an internal representation selected at a particular boundary by
a terminal channel--coordinate query.
The output is not posed as reconstruction of the complete source input from a
compressed code.  It is the input-domain state obtained by tracing the selected
representation along the target-generating sample's own local inverse branch.
The known input instantiates that branch through the frozen model's actual
forward activations and differential operators.
Similarity to the source input, where defined in modality-native coordinates,
is a descriptive diagnostic rather than the definition of a correct feature
inverse.

\begin{figure*}[t]
  \centering
  \includegraphics[width=\textwidth]{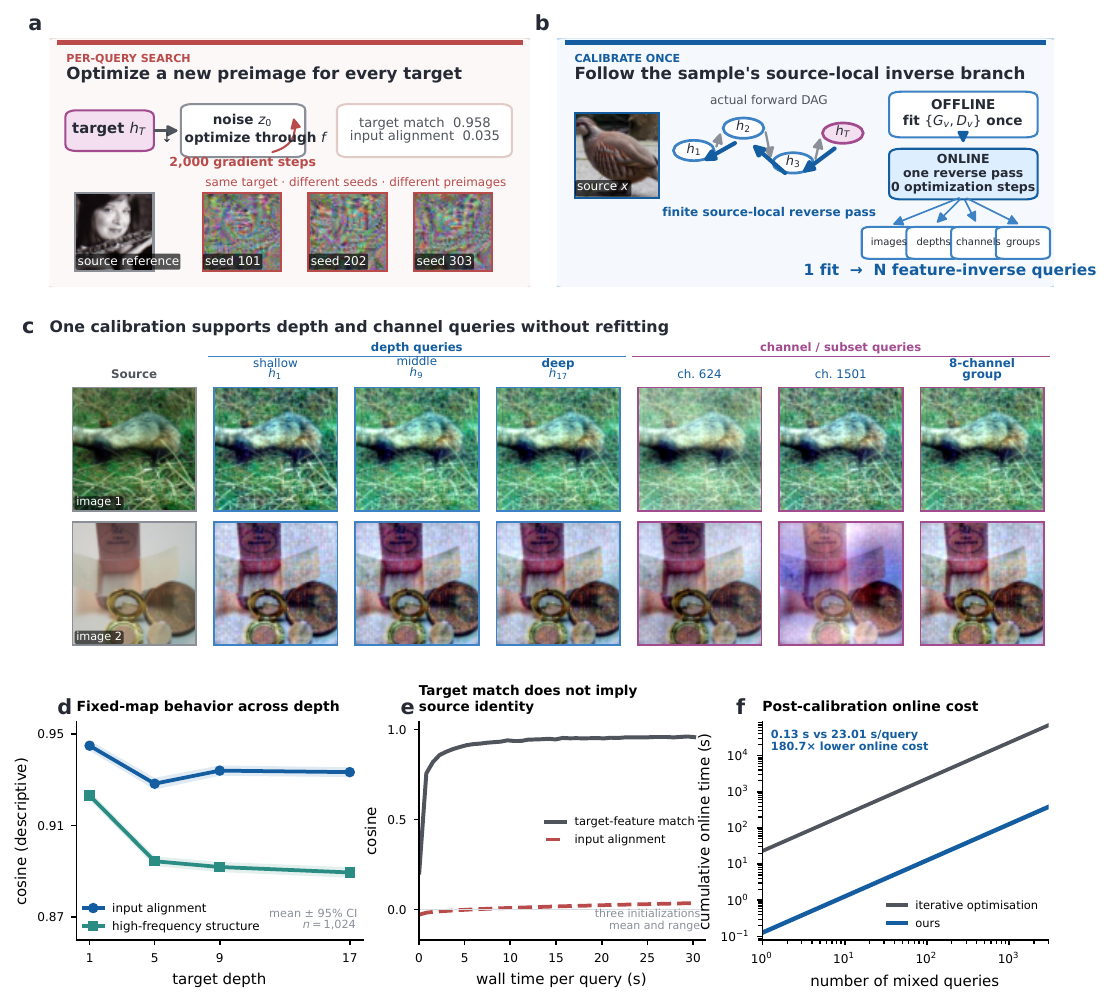}
  \caption{\textbf{From target-preimage search to source-grounded feature
  inversion.}
  \textbf{a}, Canonical iterative inversion starts from noise and repeats
  target-matching optimisation for every query; three seeds reach similar
  target-feature match but weak input-instance alignment.
  \textbf{b}, The proposed regime calibrates reusable local repairs and applies
  one finite reverse pass through the target-generating sample's source-local
  boundary DAG.
  \textbf{c}, One ResNet50 map family serves image, depth, channel, and channel-group
  queries without refitting.
  \textbf{d}, Input-instance alignment and high-frequency structure across
  target depths (mean and 95\% confidence interval, $n=1{,}024$).
  \textbf{e}, Optimisation target-feature match and input-instance alignment
  versus wall time (mean and range over three initialisations).
  \textbf{f}, Post-calibration cumulative online time for repeated mixed
  queries; calibration time is excluded to isolate deployment-time scaling.
  Source images are denormalised for display, and inverse tensors use one
  global per-image 1st--99th percentile scaling.}
  \label{fig:method_paradigm}
\end{figure*}

We formulate feature inversion as \emph{local cotangent repair} on a frozen
model's finite differentiable tensor-level computational DAG exposed by
autograd.
Raw backpropagation already follows the correct target-to-input topology, but
its vector-Jacobian products (VJPs) are adjoint transport signals rather than
inverse representations
\citep{rumelhart1986learning,baydin2018automatic}.
Our method calibrates closed-form local repair maps at selected tensor
boundaries and composes them from the target toward the input.
The final estimator has two closed-form stages.
First, a local matrix Wiener map $G_v$ transforms a mean-seed VJP into an
upstream-state estimate and amortises the missing covariance correction over
the calibration distribution \citep{wiener1949extrapolation}.
Because the VJP is evaluated at the current forward point, this stage already
retains source-local differential geometry and is often strong when the local
operator is fixed or weakly sample dependent.
Second, a Jacobian-vector product (JVP) at the same forward point exposes the
mismatch left by the shared first-stage map, and a second Wiener map $D_v$
predicts the remaining local state error.
This correction applies to any predictable first-stage residual.
Evaluating the JVP at the actual source forward point is especially relevant
for sample-dependent local operators.
The setting is input-conditioned: the input $x$ remains available to instantiate
the frozen forward point and its JVP/VJP operators.
The reverse recursion is seeded by the selected target state and does not use
the true intermediate activations as online inverse targets.
There is no direct additive pathway from $x$ to the reported inverse; $x$ enters
only through the selected target state and the local operators evaluated at its
forward point.
Figure~\ref{fig:method_paradigm} contrasts this reusable source-grounded pass
with per-query target-preimage optimisation.
The section first defines the graph-local observable, then derives its
closed-form repair class and calibration.  It next introduces JVP-FC, composes
both stages through the boundary DAG, and closes with deployment conventions.

\subsection{Feature Inversion as Local DAG Cotangent Repair}
\label{sec:method_setup}

Let $f$ be a frozen differentiable model and let $\calG=(\calV,\calE)$ denote
its finite tensor-level computational DAG exposed by autograd.
For a selected boundary node $v\in\calV$, its forward activation on input $x$ is
denoted $\Hv_v(x)$.
Let $\mathcal C_v$ be the selected non-overlapping child frontier of $v$ on
paths to the target.
If two candidate children are nested, the upstream candidate is removed so the
same downstream computation is not counted twice.
The precise frontier construction is given in \cref{app:notation_axes}.
For each child $u\in\mathcal C_v$, fix the input $x$ and hold any external
subgraph inputs that are not descendants of $v$ at their forward values.
The intervening subgraph then defines a local partial map
\[
  \Hv_u(x)=f_{u\leftarrow v,x}(\Hv_v(x)),
  \qquad
  J_{u\leftarrow v}(x)
  =
  \frac{\partial f_{u\leftarrow v,x}}{\partial \Hv_v}
  \bigg|_{\Hv_v=\Hv_v(x)} .
\]
This Jacobian includes all directed paths from $v$ to $u$ inside the local
subgraph and is the derivative applied by autograd.

The local source used to calibrate node $v$ is the VJP induced by a
channel-averaged child energy.
If child $u$ has $C_u$ channels, define
\begin{equation}
  s_u(x)=\frac{1}{C_u}\Hv_u(x).
  \label{eq:mean_seed}
\end{equation}
Equivalently, $s_u$ is the gradient of the scalar local functional
$\Psi_u(\Hv_u)=\|\Hv_u\|_F^2/(2C_u)$.
Since the VJP is linear in its seed, the resulting VJP source is the average of
the channel-energy cotangents of the child tensor.
The factor $1/C_u$ is part of the seed definition and is applied at the child
where the seed is created; coordinate positions are not averaged in this gauge.
For a node with multiple selected children, autograd sums the child VJPs in its
native multi-output gauge; no additional division by $|\mathcal C_v|$ is used.
The local mean-seed VJP source is therefore
\begin{equation}
  \Rsrc_v(x)
  =
  \sum_{u\in\mathcal C_v}
  J_{u\leftarrow v}(x)^* s_u(x)
  =
  \sum_{u\in\mathcal C_v}
  J_{u\leftarrow v}(x)^*
  \frac{\Hv_u(x)}{C_u}.
  \label{eq:local_mean_raw}
\end{equation}
The calibration target is the forward activation at the same node:
\begin{equation}
  \Gv_v(\Rsrc_v(x)) \approx \Hv_v(x).
  \label{eq:local_target}
\end{equation}
This fit is strictly graph-local: $\Rsrc_v$ depends only on the current boundary
node, its selected child boundaries, the local Jacobians between them, and the
child activations on the same input.
Here local refers to adjacency in the selected boundary DAG, not necessarily to
spatial locality; an adjacent subgraph may contain global attention or pooling.

\subsection{Local Matrix Wiener Repair}
\label{sec:local_matrix_wiener}

After identifying the batch and channel axes, write each tensor, up to a
notational axis permutation, as
\begin{equation}
  \Hv_v \in \R^{B\times C\times d_1\times\cdots\times d_k},
  \label{eq:tensor_shape}
\end{equation}
where $B$ is the batch dimension, $C$ is the channel or feature dimension, and
$d_1,\ldots,d_k$ are coordinate dimensions.
The coordinate dimensions may be spatial positions, token positions,
query-key coordinates, or absent.
The method uses one tensor rule: Fourier-transform the coordinate axes and use
a full matrix over channels at each stored coordinate frequency.  A declared
stored-frequency partition may tie this matrix across selected frequency bins;
this changes the operator capacity but not the local source, closed-form solve,
or DAG recursion.

Let $\calF_k$ be the $k$-dimensional Fourier transform over coordinate axes,
with $\calF_0$ defined as the identity.
For the mathematical derivation we use the unitary convention for $\calF_k$, so
that Parseval's identity holds exactly.
The implementation uses a real fast Fourier transform (FFT) and stores only the
non-redundant spectrum, which represents the same real frequency-diagonal
operator family.
The relative ridge is defined over these stored bins; a change in FFT
normalisation rescales the stored second moments and effective ridge
consistently.
We denote
\[
  \Hv'_v=\calF_k\Hv_v,\qquad
  \Rsrc'_v=\calF_k\Rsrc_v .
\]
Let $\Omega_v^{\mathrm{st}}$ be the stored non-redundant frequency-bin set and
let $\Pi_v$ be a partition of this set.  For $\omega\in g$ and
$g\in\Pi_v$, the tied operator satisfies
\begin{equation}
  \Gv_v(\omega)=\Gv_{v,g}.
  \label{eq:frequency_partition}
\end{equation}
The singleton partition recovers an independent matrix at every stored
frequency, whereas a single group ties one matrix across all stored bins.
At each coordinate frequency $\omega$, the repair applies a full $C\times C$
matrix:
\begin{equation}
  \Gv_v(\Rsrc_v)
  =
  \calF_k^{-1}
  \left[
    \Gv_v(\omega)
    (\calF_k\Rsrc_v)(\omega)
  \right],
  \qquad
  \Gv_v(\omega)\in\CC^{C\times C}.
  \label{eq:local_operator}
\end{equation}
Thus channels may mix arbitrarily at a fixed coordinate frequency, while
different coordinate frequencies are not mixed by the repair operator.
Spectral tying constrains selected frequencies to share parameters but does not
remove the frequency content already present in the source observable.
Frequency diagonality is the only cross-coordinate coupling restriction on the
local map.
It is introduced for statistical tractability and computational efficiency, not
as a claim that the original network is translation equivariant.
The estimator additionally remains graph-local, linear, and zero-intercept.
Table~\ref{tab:tensor_rules} summarises the resulting axis rule for the tensor
families considered in this paper.

\begin{table}[t]
  \centering
  \caption{One tensor rule for different node types.}
  \label{tab:tensor_rules}
  \small
  \begin{tabular}{llll}
    \toprule
    Node type & Tensor shape & Channel axis & Coordinate transform \\
    \midrule
    CNN feature & $(B,C,H,W)$ & $C$ & 2D Fourier over $(H,W)$ \\
    Token stream & $(B,N,D)$ & $D$ & 1D Fourier over $N$ \\
    Vector / GAP / MLP & $(B,D)$ & $D$ & None; one matrix \\
    Attention tensor & $(B,h,N,N)$ & $h$ & 2D Fourier over query-key axes \\
    \bottomrule
  \end{tabular}
\end{table}

\subsection{Closed-Form Local Calibration}
\label{sec:local_calibration}

For each node $v$ and stored-frequency group $g\in\Pi_v$, the map
$\Gv_{v,g}$ is obtained from a ridge-regularised empirical least-squares
problem on a calibration set $\calD_{\mathrm{cal}}$.  The evaluated
implementation defines this objective by weighting the stored non-redundant
rFFT bins equally:
\begin{equation}
  \Gv_{v,g}^\star
  =
  \arg\min_{\Gv}
  \frac{1}{|g|}
  \sum_{\omega\in g}
  \E_{x\sim\calD_{\mathrm{cal}}}
  \left\|
    \Gv \Rsrc'_v(x,\omega)-\Hv'_v(x,\omega)
  \right\|_2^2
  +
  \lambda \|\Gv\|_F^2 .
  \label{eq:wiener_objective}
\end{equation}
Define the cross- and auto-spectral matrices
\begin{align}
  \SHR_v(\omega)
  &=
  \E\!\left[
    \Hv'_v(\omega)\Rsrc'_v(\omega)^*
  \right],
  \label{eq:shr} \\
  \SRR_v(\omega)
  &=
  \E\!\left[
    \Rsrc'_v(\omega)\Rsrc'_v(\omega)^*
  \right].
  \label{eq:srr}
\end{align}
All expectations are empirical averages over calibration inputs.
For each calibration example and stored frequency, $\Hv'_v(\omega)$ and
$\Rsrc'_v(\omega)$ are $C$-dimensional channel vectors; the batch dimension
therefore contributes independent calibration samples and is not a matrix axis
of $\Gv_v(\omega)$.
This convention assumes evaluation-mode models whose forward operators do not
couple different batch elements, as in all experiments reported here.
Define the equally pooled stored-bin moments
\begin{equation}
  \overline{\SHR}_{v,g}
  =
  \frac{1}{|g|}\sum_{\omega\in g}\SHR_v(\omega),
  \qquad
  \overline{\SRR}_{v,g}
  =
  \frac{1}{|g|}\sum_{\omega\in g}\SRR_v(\omega).
  \label{eq:grouped_spectral_moments}
\end{equation}
For $\lambda>0$, the unique equality-constrained solution to the declared
stored-spectrum objective is the grouped matrix Wiener map
\begin{equation}
  \boxed{
  \Gv_{v,g}^\star
  =
  \overline{\SHR}_{v,g}
  \left(
    \overline{\SRR}_{v,g}+\lambda I
  \right)^{-1}
  },
  \qquad
  \Gv_v^\star(\omega)=\Gv_{v,g}^\star
  \quad(\omega\in g).
  \label{eq:matrix_wiener}
\end{equation}
For the singleton partition, this reduces to the independent per-frequency
solution.  Because non-redundant rFFT bins have unequal Parseval multiplicities,
\cref{eq:wiener_objective} is specifically the equally weighted stored-spectrum
objective; it is not presented as the coordinate-domain MSE with full-spectrum
multiplicity weights.
No gradient-based parameter training is used for these maps.
They are determined entirely by second-order calibration statistics and a
scalar ridge parameter.
In experiments we set the effective ridge by the relative rule
\begin{equation}
  \lambda_{\mathrm{eff}}
  =
  \rho\,
  \max\!\left\{
    \mathrm{mean}_{\omega,c}\SRR_{v,cc}(\omega),
    \epsilon_{\mathrm{ridge}}
  \right\},
  \qquad
  \rho=0.01,
  \quad
  \epsilon_{\mathrm{ridge}}=10^{-30},
  \label{eq:relative_ridge}
\end{equation}
and use $\lambda=\lambda_{\mathrm{eff}}$ in \cref{eq:matrix_wiener}.
The diagonal entries of $\SRR_v(\omega)$ are real and nonnegative because
$\SRR_v(\omega)$ is Hermitian positive semidefinite; numerically, the real
diagonal is used for this average.
The numerical floor only protects a degenerate zero-source case and is
negligible at ordinary source scales.

The calibration of $\Gv_v$ is independent across selected nodes once the
boundary DAG is fixed.
Each node uses its own local child set $\mathcal C_v$ and the local source in
\cref{eq:local_mean_raw}.
The maps may therefore be solved in reverse order, shallow-to-deep order, or in
parallel, provided that each node is paired with the same local source and
target variables.

\subsection{JVP Forward-Consistency Correction}
\label{sec:jvp_fc}

The first-stage source $\Rsrc_v(x)$ already contains the local Jacobians at the
current source operating point through its VJP construction.
However, the map $\Gv_v$ is shared across calibration and deployment samples and
therefore represents a distribution of local source-to-state relations.
This distributional projection can be accurate when the local operator is fixed
or varies weakly across samples.
When normalisation, attention, or data-dependent gating or routing makes the
local operator strongly sample dependent, the ideal correction also changes
with the operating point.
JVP-FC constructs a second observable by testing the first-stage estimate under
the actual differential at the source forward point.
Its formal regression can correct any predictable first-stage residual, while
its principal motivation is the operating-point-specific mismatch left by a
shared first-stage map.

The first-stage estimate
\begin{equation}
  y_v^0(x)=\Gv_v(\Rsrc_v(x))
  \label{eq:first_stage_state}
\end{equation}
is a zero-intercept state estimate in the activation coordinates of node $v$.
The remaining error is measured in the same local linear gauge used by
autograd.
Although $y_v^0$ is stored in activation coordinates, it is supplied to the JVP
as a tangent vector at the fixed forward point.
We ask whether the differential of the current local subgraph maps this vector
to the child state observed at the adjacent boundary.
We then repair the induced residual by a second closed-form map.
For each child $u\in\mathcal C_v$, compute the JVP at the actual forward point:
\begin{equation}
  \widehat \Hv_{u\mid v}(x)
  =
  J_{u\leftarrow v}(x)\,y_v^0(x),
  \qquad
  e_{u\mid v}(x)
  =
  \Hv_u(x)-\widehat \Hv_{u\mid v}(x).
  \label{eq:jvp_child_residual}
\end{equation}
This residual is well defined in the child coordinate space, but it is not
assumed to vanish when $y_v^0=\Hv_v$. Writing
$\Delta_v=\Hv_v-y_v^0$ gives the exact decomposition
\begin{equation}
  e_{u\mid v}
  =
  \underbrace{\Hv_u-J_{u\leftarrow v}\Hv_v}_{b_{u\mid v}}
  +
  J_{u\leftarrow v}\Delta_v.
  \label{eq:jvp_residual_decomposition}
\end{equation}
The base term $b_{u\mid v}$ is generally nonzero for maps with offsets or
non-homogeneous nonlinearities.
Equality $\Hv_u=J_{u\leftarrow v}\Hv_v$ does hold when the local map is
positively one-homogeneous at the forward point, including the bias-free linear
case, but it does not hold in general.
The second term $J_{u\leftarrow v}(x)\Delta_v$ transports the first-stage state
error through the current sample's local differential and therefore exposes the
source-specific response of that error.
JVP-FC therefore learns a distributional residual-to-error relation rather than
enforcing an exact nonlinear forward equation.
The correction source is the VJP of this child residual:
\begin{equation}
  C_v(x)
  =
  \sum_{u\in\mathcal C_v}
  J_{u\leftarrow v}(x)^* e_{u\mid v}(x).
  \label{eq:jvp_fc_source}
\end{equation}
Holding the forward point and child states fixed, this source is the negative
gradient of the local linear child-consistency energy
\[
  \Phi_v(y)=\frac12\sum_{u\in\mathcal C_v}
  \left\|\Hv_u-J_{u\leftarrow v}y\right\|_F^2,
  \qquad
  C_v=-\nabla_y\Phi_v(y)\big|_{y=y_v^0}.
\]
The first-stage seed is the differential of channel-averaged child energy,
whereas the JVP-FC source pulls back a Euclidean child-state residual.
No additional channel averaging is applied to $e_{u\mid v}$; the same residual
gauge is used during calibration and online inversion.
Let $\Delta_v(x)=\Hv_v(x)-y_v^0(x)$, $C'_v=\calF_k C_v$, and
$\Delta'_v=\calF_k\Delta_v$.
For each group $g\in\Pi_v$, we fit a second matrix Wiener map in the same
frequency-diagonal, full-channel class:
\begin{equation}
  D_{v,g}^\star
  =
  \arg\min_D
  \frac{1}{|g|}\sum_{\omega\in g}
  \E_{x\sim\calD_{\mathrm{cal}}}
  \left\|D C'_v(\omega)-\Delta'_v(\omega)\right\|_2^2
  +
  \lambda_D\|D\|_F^2 .
  \label{eq:jvp_fc_objective}
\end{equation}
Define
$S_v^{\Delta C}(\omega)=\E[\Delta'_v(\omega)C'_v(\omega)^*]$ and
$S_v^{CC}(\omega)=\E[C'_v(\omega)C'_v(\omega)^*]$, together with the
equally pooled group moments
\[
  \overline S_{v,g}^{\Delta C}
  =\frac{1}{|g|}\sum_{\omega\in g}S_v^{\Delta C}(\omega),
  \qquad
  \overline S_{v,g}^{CC}
  =\frac{1}{|g|}\sum_{\omega\in g}S_v^{CC}(\omega).
\]
The grouped correction is
\begin{equation}
  D_{v,g}^\star
  =
  \overline S_{v,g}^{\Delta C}
  \left(
    \overline S_{v,g}^{CC}+\lambda_D I
  \right)^{-1},
  \qquad
  D_v^\star(\omega)=D_{v,g}^\star
  \quad(\omega\in g).
  \label{eq:jvp_fc_wiener}
\end{equation}
The correction ridge $\lambda_D$ is set by the same relative rule as
\cref{eq:relative_ridge}, but using the diagonal mean of the correction-source
second moment $S_v^{CC}$.
The repaired local state is
\begin{equation}
  y_v(x)=y_v^0(x)+D_v^\star(C_v(x)).
  \label{eq:jvp_fc_state}
\end{equation}
This stage is still closed-form: $D_v$ is solved by the same spectral
second-order estimator as \cref{eq:matrix_wiener}, with $C_v$ as source and
$\Hv_v-y_v^0$ as target.
The JVP is the differential of the current local subgraph at the actual forward
activation.
It defines the child response in the zero-intercept linear gauge of the repair
problem; the method does not run the nonlinear subgraph on $y_v^0$ as if it
were a new forward activation, nor does it require that
$J_{u\leftarrow v}(x)y_v^0$ equal $f_{u\leftarrow v}(y_v^0)$.

\subsection{Topology-Respecting Feature Inversion}
\label{sec:topology_respecting_inversion}

At inversion time, the solved maps are composed from the target toward the
input.
Choose a nonempty terminal channel set
$S\subseteq\{1,\ldots,C_T\}$ and a nonempty terminal coordinate set $Q$.
For tensors with multiple coordinate axes, $Q$ is a set of coordinate tuples;
when no coordinate axis is present, it contains the single trivial coordinate.
Let $P_S^{\mathrm{ch}}$ and $P_Q^{\mathrm{coord}}$ be shape-preserving masks and
define the orthogonal mask projector
\begin{equation}
  P_{S,Q}
  =P_S^{\mathrm{ch}}P_Q^{\mathrm{coord}},
  \qquad
  P_{S,Q}^*=P_{S,Q}=P_{S,Q}^2,
  \qquad
  C_S=|S|.
  \label{eq:terminal_mask_projector}
\end{equation}
The projector zeros unselected channel-coordinate entries without deleting a
tensor axis.  The pair $(S,Q)$ defines the terminal channel--coordinate query.
The terminal state and its online cotangent seed are
\begin{equation}
  y_T=P_{S,Q}\Hv_T,
  \qquad
  s_T^{\mathrm{on}}=\frac{y_T}{C_S}
  =\frac{P_{S,Q}\Hv_T}{C_S}.
  \label{eq:masked_terminal_seed}
\end{equation}
This seed is the differential of the channel-averaged masked terminal energy
\begin{equation}
  \Psi_T^{(S,Q)}(\Hv_T)
  =
  \frac{1}{2|S|}\left\|P_{S,Q}\Hv_T\right\|_F^2.
  \label{eq:masked_terminal_energy}
\end{equation}
The coordinate-set cardinality therefore does not enter the seed divisor.
The full target uses $P_{S,Q}=I$ and $C_S=C_T$.
A single-channel, all-coordinate query has $C_S=1$ and
$Q=Q_{\mathrm{all}}$, where $Q_{\mathrm{all}}$ is the complete coordinate set.
If $Q$ is restricted separately, only the selected coordinates are retained.
The reported GPT-2 terminal-token query retains all hidden channels and uses
$Q=\{L-1\}$, so its divisor is the hidden width rather than the number of
selected token positions.
For a shallower node $v$, assume that online child states
$\{y_u:u\in\mathcal C_v\}$ have already been inverted to that boundary.
The seed attached to each online child is
\begin{equation}
  s_u^{\mathrm{on}}
  =
  \begin{cases}
    y_T/C_S, & u=T,\\
    y_u/C_u, & u\neq T.
  \end{cases}
  \label{eq:online_child_seed}
\end{equation}
Thus only the terminal child uses the selected-channel count $C_S$.
Every inverted nonterminal child uses its physical channel count $C_u$.
The online local source is
\begin{equation}
  \Rsrc_v^{\mathrm{on}}(x)
  =
  \sum_{u\in\mathcal C_v}
  J_{u\leftarrow v}(x)^*
  s_u^{\mathrm{on}}(x).
  \label{eq:online_raw}
\end{equation}
The first-stage online estimate is $y_v^0=\Gv_v(\Rsrc_v^{\mathrm{on}})$.
The online child residual uses the repaired child states, not the true child
activations or their normalised seeds:
\begin{equation}
  C_v^{\mathrm{on}}(x)
  =
  \sum_{u\in\mathcal C_v}
  J_{u\leftarrow v}(x)^*
  \left(
    y_u(x)-J_{u\leftarrow v}(x)y_v^0(x)
  \right).
  \label{eq:online_jvp_fc_source}
\end{equation}
The final online state is
\begin{equation}
  y_v(x)
  =
  y_v^0(x)+D_v^\star(C_v^{\mathrm{on}}(x)).
  \label{eq:online_jvp_fc_state}
\end{equation}
When $v$ is the input node, $y_v$ is the reported input-domain feature inverse.
Changing $(S,Q)$ changes only the terminal initialisation and the states induced
by the ensuing recursion; the calibrated local maps remain fixed.

Branches and multi-use tensors are handled by native autograd semantics.
Residual additions distribute cotangents to branches, concatenations slice them
back to their inputs, and one-to-many uses sum incoming cotangents at the shared
tensor.
Our method repairs the cotangent or state at selected tensor boundaries; it does
not introduce special component-specific inverse rules. The estimator is
component-agnostic once architecture-aware boundary and axis metadata have been
specified.

\subsection{Boundary Modes, Tensor Axes, and Spectral Tying}
\label{sec:boundary_modes}

The theoretical maximal form assigns a local matrix Wiener repair to every
captured differentiable floating tensor node whose channel and coordinate axes
are specified.
Scalar nodes, non-floating tensors, and bookkeeping operations without a
meaningful channel axis are traversed by autograd but are not instantiated as
repair nodes.
The implementation supports three boundary modes:
\begin{itemize}
  \item \textbf{Full-DAG}: instantiate repairs at all captured differentiable floating tensor nodes with a specified channel axis.
  \item \textbf{Coarse/stage block}: instantiate repairs at the input, stem or patch embedding, and architecture block or stage edges; this is the default efficient mode for cross-model evaluation.
  \item \textbf{Custom list}: capture the full DAG but instantiate only user-specified nodes.
\end{itemize}
These are instantiation choices, not different mathematical methods.
All modes use the same local source, Wiener repair, and JVP-FC correction in
\cref{eq:local_mean_raw,eq:matrix_wiener,eq:jvp_fc_objective}.
The formulation requires that the intervening frozen subgraphs expose the
matrix-free VJP and JVP operations used by these equations.  Architecture
onboarding therefore consists of exposing meaningful floating-tensor
boundaries and declaring their axes; it is distinct from the closed-form
estimator itself.

The axis rule is architecture-agnostic.
The batch axis is fixed to dimension zero.
The channel axis is either inferred or specified by the user.
All remaining axes are coordinate axes and are transformed by $\calF_k$.
Singleton coordinate axes that only record global pooling output are treated as
trivial and may be collapsed before this rule is applied.
For ViT token streams, this gives a 1D Fourier transform over token positions
and a full matrix over hidden features.
The stored-frequency partition $\Pi_v$ is also declared per node.  The reported
vision configurations use the singleton partition, whereas the GPT-2 small
configuration ties one complex full-channel gain matrix across all stored
token-frequency bins for both $G_v$ and $D_v$ under the declared rFFT/irFFT
convention.  This spectral tying does not itself guarantee causal support;
causal dependencies enter through the source-local GPT-2 computational DAG and
its VJP/JVP operators.  The formal semantic and syntactic language-routing
claims consequently use terminal-token queries.  The supplementary joint
full-coordinate channel-subset control does not validate isolated
nonterminal-position queries.
For attention weights or logits, the repaired object is a cotangent or local
state estimate, not a probability matrix that is fed through the forward model.
Softmax simplex constraints are therefore not explicitly enforced by the
reverse estimator.  The same closed-form rule is algebraically applicable, but
the adequacy of this relaxation for internal attention tensors remains an
empirical question.

\subsection{Practical Details}
\label{sec:practical_details}

We use zero-intercept matrix maps throughout.
There is no additive bias term in either $G_v$ or $D_v$.
Consequently, the fitted maps cannot inject a fixed target-independent template;
all inverted structure must enter through the selected target state, the
model Jacobians, and the calibrated linear repairs.
Full matrix calibration requires sufficient effective samples for each
$C\times C$ second-moment estimate.
In high-dimensional token nodes, calibration size is a dominant practical
factor; our ViT token-stream runs use the larger 4096-image calibration setting
for this reason, and the reported GPT-2 map families use 4,096 calibration
windows.
Unless stated otherwise, all reported default configurations use mean-full
calibration seeds,
meaning that all child channels are retained and divided by their channel count
as in \cref{eq:mean_seed}.
These default configurations also use no source normalisation, full-channel
zero-intercept solves, the relative ridge coefficient $\rho=0.01$ for both
stages, and unit JVP-FC residual and correction weights.
The headline vision configuration uses coarse/stage
boundaries and the singleton per-frequency partition.  The reported GPT-2
configuration uses selected custom repair boundaries and all-shared frequency
tying.  These are declared boundary and operator-capacity instantiations of the
same two-stage estimator.
These definitions complete the deployed algorithm.
The next section derives its statistical and geometric motivation and separates
the exact closed-form statements from the empirical proxy assumptions.

\section{Theory: From Optimal Inversion to Closed-Form DAG Repair}
\label{sec:theory}

We derive the deployed estimator through a sequence of explicit relaxations.
Throughout this section, inversion means upstream-state estimation along the
source sample's local network geometry.  The estimated objects are tensor
states on this inverse branch, not generic reconstructions of the complete
source input from a compressed code.
The derivation is modality neutral.
Modality-specific structure enters through the frozen tensor-level DAG and its
source-local VJP/JVP operators, the declared channel and coordinate axes, the
stored-frequency partition $\Pi_v$, and the terminal projector $P_{S,Q}$.
The starting point is the minimum-mean-square statistical inverse, namely the
conditional mean of an upstream state given a downstream observation.
Under a centred local linear-Gaussian reference model, the conditional mean of
the local deviations reduces to a covariance-weighted Wiener operator.
The same operator is a natural adjoint under covariance-induced metrics.
Directly constructing these boundary- and operating-point-dependent metrics is
impractical.
We therefore construct a graph-local Euclidean VJP observable and estimate its
ridge-regularised closed-form distributional projection to the forward state.
The VJP observable retains the current source-local Jacobian, but one fitted
projection must interpret these observables across the calibration distribution.
It can capture the dominant correction for fixed or weakly sample-dependent
local operators, while strongly sample-dependent operators can leave an
operating-point-specific error.
A second Wiener projection uses a pulled-back JVP residual at the actual forward
point to repair the predictable part of this remaining error.
The resulting maps are then composed in one finite target-to-input DAG pass.

\subsection{Optimal Statistical Inversion under Mean-Squared Error}
\label{sec:optimal_statistical_inversion}

Let $H_v$ be an upstream state and $H_u$ a downstream observation generated by
a frozen model on the data distribution.
Among all measurable estimators $\mathcal T$, the MSE-optimal
statistical inverse satisfies
\begin{equation}
  \mathcal T^\star
  \in
  \arg\min_{\mathcal T}
  \E\left\|\mathcal T(H_u)-H_v\right\|_2^2,
  \qquad
  \mathcal T^\star(h_u)
  =
  \E[H_v\mid H_u=h_u].
  \label{eq:conditional_mean_inverse}
\end{equation}
Thus the conditional mean is the ideal MSE estimator of the upstream state from
the downstream feature alone.
It is generally unavailable because it requires the full conditional law of
$H_v$ given $H_u$.

Our setting is additionally input-conditioned.
The input $x$ being explained remains available to instantiate the actual
forward activations and local derivatives of the frozen model.
We do not replace \cref{eq:conditional_mean_inverse} by
$\E[H_v\mid H_u,x]$: in a deterministic frozen model, conditioning on the full
input makes $H_v$ known and no longer describes inversion from the downstream
feature.
Instead, $x$ selects the forward point and local operators used to replace one
global nonlinear inverse by a sequence of local inverse problems.

\subsection{Local Linearisation and the Adjoint--Inverse Gap}
\label{sec:adjoint_not_inverse}

Let a local downstream map $y=f(h)$ be linearised at the current activation
$h_0$ as
\[
  \delta y=J(h_0)\,\delta h .
\]
Backpropagation of a seed $q$ through this mapping gives
\begin{equation}
  r=J(h_0)^\top q.
  \label{eq:adjoint_signal}
\end{equation}
This is the Euclidean adjoint of the local Jacobian.
We use $\top$ for real-domain Jacobians and $*$ for conjugate transpose in
complex or Fourier coordinates.
In general,
\begin{equation}
  J^\top\neq J^\dagger,
  \label{eq:adjoint_not_pinv}
\end{equation}
except in special partial-isometry cases, where every nonzero singular value of
$J$ equals one.
Raw VJPs therefore follow the correct computational dependencies but carry the
geometry of adjoint transport rather than that of an upstream-state inverse.

To express the missing geometry, equip the parent and child feature spaces with
Hermitian positive-definite metric operators $M_v$ and $M_u$.
We use the weighted inner products
\begin{equation}
  \langle a,b\rangle_{M_v}:=a^*M_vb,
  \qquad
  \langle c,d\rangle_{M_u}:=c^*M_ud.
  \label{eq:weighted_inner_products}
\end{equation}
The unique Riesz adjoint of $J_{u\leftarrow v}$ under these metrics is
\begin{equation}
  J_{u\leftarrow v}^{\sharp}
  =M_v^{-1}J_{u\leftarrow v}^*M_u.
  \label{eq:natural_adjoint}
\end{equation}
Indeed, equality
\[
  \langle J_{u\leftarrow v}\delta_v,s_u\rangle_{M_u}
  =
  \langle\delta_v,J_{u\leftarrow v}^{\sharp}s_u\rangle_{M_v}
\]
for all $\delta_v$ and $s_u$ implies
$M_vJ_{u\leftarrow v}^{\sharp}=J_{u\leftarrow v}^*M_u$.
The operators $M_u$ and $M_v$ are the metrics themselves.
The factor $M_v^{-1}$ appears when the adjoint is represented in parent
coordinates.

\subsection{Wiener Inversion as a Covariance-Induced Natural Adjoint}
\label{sec:wiener_natural_adjoint}

The statistical and geometric views coincide in a local linear-Gaussian model.
Fix a forward point and hold $J_{u\leftarrow v}$ fixed in the following
idealised local ensemble.
Vectorise the local tensors for this unrestricted reference calculation, and
write $\widetilde H_v$ and $\widetilde H_u$ for centred parent and child
deviations.
Assume
\begin{equation}
  \widetilde H_u=J_{u\leftarrow v}\widetilde H_v+\eta,
  \qquad
  \E[\eta \widetilde H_v^*]=0.
  \label{eq:local_linear_gaussian_model}
\end{equation}
For multiple children, stack their states and local Jacobians; the same formulas
then apply to the resulting block operator.
Define the parent covariance
$\Sigma_v=\E[\widetilde H_v\widetilde H_v^*]$, the residual covariance
$\Sigma_\eta=\E[\eta\eta^*]$, and the child covariance
\begin{equation}
  \Sigma_u
  =
  \E[\widetilde H_u\widetilde H_u^*]
  =
  J_{u\leftarrow v}\Sigma_vJ_{u\leftarrow v}^*
  +\Sigma_\eta.
  \label{eq:local_child_covariance}
\end{equation}
Assuming $\Sigma_u$ is nonsingular, the linear minimum-mean-square-error
(LMMSE) estimator is
\begin{equation}
  \mathcal T_{u\to v}^{\mathrm{LMMSE}}
  =
  \Sigma_vJ_{u\leftarrow v}^*\Sigma_u^{-1}.
  \label{eq:local_wiener_inverse}
\end{equation}
For jointly Gaussian centred deviations, this linear estimator equals
$\E[\widetilde H_v\mid \widetilde H_u]$, the zero-mean specialisation of
\cref{eq:conditional_mean_inverse}.

If $\Sigma_v$ is also nonsingular, choose covariance-induced precision metrics
\begin{equation}
  M_v=\Sigma_v^{-1},
  \qquad
  M_u=\Sigma_u^{-1}.
  \label{eq:covariance_metrics}
\end{equation}
Substitution into \cref{eq:natural_adjoint} gives
\begin{equation}
  J_{u\leftarrow v}^{\sharp}
  =
  \Sigma_vJ_{u\leftarrow v}^*\Sigma_u^{-1}
  =
  \mathcal T_{u\to v}^{\mathrm{LMMSE}}.
  \label{eq:wiener_natural_equivalence}
\end{equation}
Hence, after vectors and covectors are identified through the stated metrics,
the local Wiener inverse and the natural adjoint are the statistical and
geometric forms of the same covariance-corrected local operator.
Centring is used only for this ideal reference identity.
The deployed maps do not subtract means or fit additive biases; their guarantees
refer to the zero-intercept objectives stated in the Method.

This identity is an ideal reference rather than our deployment formula.
In a nonlinear network, the Jacobian and relevant local geometry depend on the
input and boundary.
Direct deployment would require estimating and applying nontrivial parent and
child metrics at every boundary for every online sample.
A projected local-oracle experiment validates the stated identity under its
controlled assumptions (\cref{app:theory_bridge_experiments}).
The next step replaces explicit metric estimation by a fixed, observable
Euclidean cotangent and its ridge-regularised distributional repair.

\subsection{Observable Mean-Seed Cotangent and the Proxy Bridge}
\label{sec:mean_seed_theory}
\label{sec:proxy_bridge}

For a child boundary $u$ with $C_u$ channels, define
\begin{equation}
  s_u=\frac{H_u}{C_u}
  =
  \nabla_{H_u}
  \left(
    \frac{1}{2C_u}\|H_u\|_F^2
  \right).
  \label{eq:theory_mean_seed}
\end{equation}
The graph-local Euclidean VJP observable is
\begin{equation}
  R_v
  =
  \sum_{u\in\mathcal C_v}
  J_{u\leftarrow v}^*\frac{H_u}{C_u}.
  \label{eq:theory_observable_source}
\end{equation}
It is exactly the adjoint pullback of the summed channel-averaged child
energies.
For multiple selected children, native multi-output VJP semantics sum the child
cotangents; no additional division by $|\mathcal C_v|$ is introduced.
During calibration, $R_v(x)$ is jointly determined by the forward child states
and the source-local Jacobians.
During deployment, $R_v^{\mathrm{on}}(x)$ replaces these child states by the
supplied repaired states, as in \cref{eq:online_raw}.
Target selection therefore propagates through the online child states, whereas
changing the operating point changes the local Jacobians.

The observable in \cref{eq:theory_observable_source} retains the actual local
Jacobian and graph topology, but it does not explicitly apply the covariance
metrics in \cref{eq:wiener_natural_equivalence}.
The ideal child-side metric would act before the pullback, while the ideal
parent-side metric would act after it.
Estimating both online is precisely the deployment obstacle identified above.

We therefore use $R_v$ as a graph-local proxy source and $H_v$ as its desired
forward-state target.
This substitution creates the main empirical bridge in the derivation.
Jacobian sparsity guarantees that the VJP follows computation dependency, but
it does not imply pointwise equality between cotangent support and activation
support.
The weaker assumption is that the organisation of $R_v$ remains predictive of
the coordinate or semantic organisation carried by $H_v$.
No-refit transfer across target depths and channel--coordinate masks tests
whether this relation is node-local, while cross-architecture and component
experiments test the same estimator class and public recipe
(\cref{sec:feature_inversion_scale,sec:cross_architecture,app:large_scale_atlas,app:component_zoo}).

\subsection{Closed-Form Distributional Wiener Repair}
\label{sec:wiener_best_linear}

Once $(R_v,H_v)$ is chosen as the observable source-target pair, the
MSE-optimal nonlinear predictor would be $\E[H_v\mid R_v]$.
We instead restrict the repair to the frequency-diagonal, full-channel linear
operator class and stored-frequency partition in
\cref{eq:local_operator,eq:frequency_partition}.
Let $R'_v=\calF_kR_v$ and $H'_v=\calF_kH_v$.
For each group $g\in\Pi_v$, calibration solves the equally weighted
stored-spectrum objective
\begin{equation}
  G_{v,g}^\star
  =
  \arg\min_G
  \frac{1}{|g|}\sum_{\omega\in g}
  \E\left\|G R'_v(\omega)-H'_v(\omega)\right\|_2^2
  +\lambda\|G\|_F^2.
  \label{eq:theory_first_projection}
\end{equation}
For $\lambda>0$, grouped ridge regression gives the unique
ridge-regularised matrix Wiener solution
\begin{equation}
  G_{v,g}^\star
  =
  \overline{\SHR}_{v,g}
  \left(
    \overline{\SRR}_{v,g}+\lambda I
  \right)^{-1},
  \qquad
  G_v^\star(\omega)=G_{v,g}^\star
  \quad(\omega\in g).
  \label{eq:theory_wiener_projection}
\end{equation}
This is the same estimator defined in
\cref{eq:wiener_objective,eq:grouped_spectral_moments,eq:matrix_wiener}.
The singleton partition recovers an independent per-frequency solution,
whereas the all-shared partition imposes one equality-constrained matrix across
all stored bins.
It requires only the joint second moments of the observable and target.
Gaussianity is not required for optimality of the stated ridge objective within
the chosen linear class.

The role of $G_v$ is now precise.
It is the unique ridge-regularised distributional projection from the Euclidean
VJP observable to the upstream forward state in the selected operator family.
It does not identify $J_{u\leftarrow v}$, $M_u$, or $M_v$.
Nor is it generally algebraically equal to the ideal natural adjoint.
Rather, it amortises the missing covariance correction over the calibration
distribution in a tractable spectral class.
The source $R_v(x)$ retains the actual $J_{u\leftarrow v}(x)$ through the VJP,
but the same fitted $G_v$ is applied to every sample at node $v$.
When the local operator and covariance geometry are fixed or vary weakly, this
shared projection can account for most of the required correction.
When they vary strongly with the source operating point, $G_v$ can leave a
sample-specific state error even though its input remains source grounded.

\subsection{JVP Residual Correction as a Second Wiener Projection}
\label{sec:jvp_fc_theory}

The second projection is designed to expose this operating-point-specific
mismatch, while remaining applicable to any predictable first-stage error.
Let
\[
  y_v^0=G_vR_v.
\]
At the actual forward point, the local JVP maps this estimate to the child
differential coordinates.
Here $y_v^0$ is used as a tangent vector in the parent coordinate space, not as
a replacement forward activation.
For each child, define
\begin{equation}
  e_{u\mid v}
  =
  H_u-J_{u\leftarrow v}y_v^0.
  \label{eq:theory_child_residual}
\end{equation}
The residual has the exact decomposition
\begin{equation}
  e_{u\mid v}
  =
  \underbrace{H_u-J_{u\leftarrow v}H_v}_{b_{u\mid v}}
  +
  J_{u\leftarrow v}(H_v-y_v^0).
  \label{eq:theory_jvp_residual_decomposition}
\end{equation}
The base term $b_{u\mid v}$ records offsets and non-homogeneous components absent
from the zero-intercept differential.
Consequently, the residual need not vanish even when $y_v^0=H_v$.
It does vanish at that point for a positively one-homogeneous local map, as
follows from Euler's homogeneous-function identity.
The second term
$J_{u\leftarrow v}(x)(H_v-y_v^0)$ evaluates the first-stage state error under
the current sample's differential.
It therefore carries the response of the shared first-stage error under the
present operating point.

Pulling the child residuals back gives a second observable
\begin{equation}
  C_v
  =
  \sum_{u\in\mathcal C_v}
  J_{u\leftarrow v}^*e_{u\mid v}.
  \label{eq:theory_correction_source}
\end{equation}
Equivalently, with
$\Phi_v(y)=\frac12\sum_{u\in\mathcal C_v}
\|H_u-J_{u\leftarrow v}y\|_F^2$, one has
$C_v=-\nabla\Phi_v(y_v^0)$.
Thus the second observable is a local child-consistency descent cotangent, not
an arbitrarily chosen residual summary.
Let $\Delta_v=H_v-y_v^0$, $C'_v=\calF_kC_v$, and
$\Delta'_v=\calF_k\Delta_v$.
For each group $g\in\Pi_v$, the correction map is the second closed-form
projection
\begin{equation}
  D_{v,g}^\star
  =
  \arg\min_D
  \frac{1}{|g|}\sum_{\omega\in g}
  \E\left\|D C'_v(\omega)-\Delta'_v(\omega)\right\|_2^2
  +\lambda_D\|D\|_F^2.
  \label{eq:theory_second_projection}
\end{equation}
Thus $D_v$ has the same ridge-regularised Wiener optimality as $G_v$, but for a
residual-to-error regression problem.
Its exact pooled solution is given in \cref{eq:jvp_fc_wiener}; this is the
objective in \cref{eq:jvp_fc_objective}.
The guarantee is conditional on the observable $C_v$.
Child-residual components annihilated by the summed pullback do not enter
$C_v$, and parent-error components that are not statistically predictable from
$C_v$ cannot be estimated by this stage.

The child prediction $J_{u\leftarrow v}y_v^0$ is a JVP response at the current
forward point.
It is not a new nonlinear evaluation of $f_{u\leftarrow v}$ on $y_v^0$.
JVP-FC is therefore a second distributional projection in a specified local
linear gauge, not an assertion that the nonlinear subgraph has no offsets or
higher-order terms.
Matched cross-architecture ablations test both the aggregate residual reduction
and the non-monotonic boundary-level behaviour predicted by this distributional,
rather than pointwise, guarantee
(\cref{sec:mechanism_experiments,app:mechanism_ablations}).

\subsection{Finite-DAG Composition and Stability}
\label{sec:finite_dag_stability}

At deployment, the calibrated maps are composed from the selected terminal
state toward the input.
The input $x$ fixes all forward activations and Jacobians, so each local online
update is linear in its supplied inverted child states.
For the concrete forward-state error statement in this subsection, consider
full-target inversion, where $y_T=H_T$ and the terminal error is zero.
Define
\[
  \delta_v=y_v-H_v.
\]
Equip every selected node space with a fixed norm and every inter-node block
with its induced operator norm.
The supplementary derivation constructs an effective two-stage reverse
operator $\mathcal K_v$ and gives the exact recursion
\begin{equation}
  \delta_v
  =
  \epsilon_v+\mathcal K_v\delta_{\mathcal C_v},
  \label{eq:main_error_recursion}
\end{equation}
where $\epsilon_v$ is the local defect obtained when the true child states are
supplied.
This identity separates local approximation error from error propagated by
imperfect online children.

If every local defect is zero, reverse topological induction gives exact input
state inversion on the selected boundary DAG.
More generally, writing $\kappa_{vu}$ for the operator norm of the block mapping
child $u$ to parent $v$, the sufficient condition
\[
  \sup_{v\ne T}
  \sum_{u\in\mathcal C_v}
  \kappa_{vu}
  \le q<1
\]
implies
\[
  \max_v\|\delta_v\|
  \le
  \frac{\max_{v\ne T}\|\epsilon_v\|}{1-q}.
\]
The pathwise bound remains informative when this uniform condition is not
satisfied.
Because the algorithm is one finite reverse pass, this is a stability result
rather than convergence of an iterative fixed-point method.
The channel--coordinate-conditioned general form for an arbitrary terminal
query $(S,Q)$ is given in \cref{app:finite_dag_stability}.
The empirical path-gain study in \cref{app:theory_bridge_experiments} does not
provide a stable global error ranking, so it is used only as a conservative
diagnostic and not as evidence of contraction or global convergence.

\subsection{Guarantees, Assumptions, and Boundaries}
\label{sec:no_hallucination}
\label{sec:assumption_summary}

Both calibrated maps are zero-intercept:
\[
  y_v^0=G_vR_v,
  \qquad
  y_v=y_v^0+D_vC_v.
\]
If the selected terminal state satisfies $P_{S,Q}H_T=0$, induction through the
online recursion gives zero mean seeds, zero VJP sources, zero residual sources,
and zero inverted states.
The estimator therefore has no additive target-independent template.
This is not a support non-expansion theorem because spectral multiplication can
spread energy in coordinate space.

The main structural restriction is coordinate-frequency diagonality.
Channels may mix through a full matrix at each coordinate frequency, but
different frequencies do not mix inside a repair map.
The partition $\Pi_v$ may additionally constrain selected stored-frequency
blocks to share one matrix.
This restriction makes full-channel calibration statistically and
computationally tractable.
Its adequacy for a given tensor family remains an empirical modelling question.
Jacobian sparsity constrains the dependency support of the raw VJP observables,
but the calibrated spectral repairs need not preserve that support in
coordinate space.
In particular, no causal-support guarantee is claimed for nonterminal queries
in causal language models.  The formal semantic and syntactic GPT-2 claims use
terminal-token queries; the supplementary joint full-coordinate
channel-subset control does not validate isolated nonterminal-position
queries.
Being defined on frozen finite differentiable DAGs exposed by autograd is a
computational-scope statement.
It does not guarantee observable sufficiency, well-conditioned calibration, or
accurate inversion for every model.
The derivation therefore separates three levels of claim.
The conditional-mean, natural-adjoint, Wiener, and finite-DAG statements are
mathematical results under their stated conditions.
The replacement of explicit sample-dependent metrics by a fixed distributional
projection is a modelling decision.
JVP-FC adds a second observable evaluated at the actual operating point, but it
does not identify the online metrics or guarantee exact removal of every
sample-dependent mismatch.
Its approximation quality, the observability of the two regression sources,
and the adequacy of the frequency-diagonal restriction are evaluated
empirically rather than guaranteed by algebraic identity.
These distinctions determine the experimental sequence: reuse and structural
scope test the proxy bridge, matched ablations test both projections, and
information-flow controls test the meaning and limits of the composed inverse.

\section{Experiments}
\label{sec:experiments}

The experiments test a single claim--evidence chain.
We first ask whether one calibrated map family supports many feature-inverse
queries without per-target optimisation.
We then test whether the same public recipe applies across architectures,
tensor components, and visual distributions.
Matched ablations isolate the roles of local DAG composition, the two
closed-form stages, frequency dependence, and cross-channel coupling.
Finally, target--operator controls define what determines a feature inverse,
and representation interventions connect the same visualised feature groups to
the model decision.
Calibration, boundary, runtime, and memory audits delimit the practical visual
deployment regime.
We close by testing the same source-grounded construction on controlled
semantic, syntactic, source-geometry, reproducibility, and intervention
experiments in GPT-2 small.

\subsection{Experimental Protocol and Evaluation Axes}
\label{sec:experimental_protocol}

\paragraph{Final method.}
Unless stated otherwise, every experiment used the deployed configuration in
\cref{sec:method}: local mean seeds, zero-intercept full-channel Wiener maps for
both $G_v$ and $D_v$, declared architecture-aware boundaries, and repaired
online child states.
The vision experiments used the singleton per-frequency partition, whereas the
reported GPT-2 experiment used all-shared spectral tying over stored token
frequencies.
No source, activation, or cotangent pre-normalisation was applied.
The relative ridge was fixed to $\rho=0.01$ for both stages.
Raw VJP, $G_v$-only, global-map, and restricted-operator variants were used
only as explicitly labelled controls.

\paragraph{Data and evaluation.}
The ImageNet protocol used 4,096 randomly sampled training images for
calibration and 1,024 disjoint validation images for evaluation
\citep{deng2009imagenet,russakovsky2015imagenet}.
Calibration and evaluation seeds were fixed to 123 and 456, respectively.
The representative distribution study used dataset-trained checkpoints on
ImageNet, Oxford-IIIT Pets, and CUB-200-2011
\citep{parkhi2012cats,wah2011caltech}.
Pets used all 3,680 available training images for calibration, while the other
dataset-specific runs used 4,096 images.
The decision-conditioned study used 16 fixed correctly classified CUB test
images per model and CUB-trained ResNet18, ConvNeXt-B, and ViT-B/32
checkpoints.
The ViT patch-coordinate study used a separate c4096 map family and 128 fixed,
correctly classified ImageNet validation images; the same maps were reused for
all 49 non-CLS patch queries.
Exact headline model, terminal-tensor, boundary, and axis records are reported
in \cref{app:experiment_protocol,tab:supp_headline_setups}; calibration/evaluation indices and
preprocessing identifiers are retained in the run artifacts.
We abbreviate a run with $N$ calibration images and $M$ evaluation images as
c$N$/e$M$.

\paragraph{Metrics and interpretation.}
Pixel cosine, low-pass luminance cosine, high-pass luminance cosine, chroma
cosine, SSIM, and LPIPS describe alignment between a feature inverse and the
known target-generating input
\citep{wang2004image,zhang2018unreasonable}.
They are \emph{input-instance alignment diagnostics}, not ground-truth
correctness measures for an inverse.
This distinction is essential for channel- and coordinate-selected targets,
whose terminal state is $P_{S,Q}H_T$ rather than the complete input.
Re-encoding cosine is reported separately as a representation-consistency
diagnostic.
For GPT-2, token-position inverse energy is defined directly from the
continuous input-embedding inverse in \cref{sec:language_extension}; it is not
an image metric or decoded-token probability.
All main ImageNet means use 1,024 evaluation images; per-image distributions
and uncertainty summaries are provided in the supplement.

\subsection{Feature Inversion at Scale: One Calibration, Many Queries}
\label{sec:feature_inversion_scale}

We first tested whether feature inversion could be used as a reusable
interpretation interface rather than a separate optimisation problem for each
query.
A single ResNet50 calibration was reused across 1,024 images, four target
depths, four individual terminal channels, and one eight-channel terminal
subset.
Changing the image, target depth, or channel mask did not refit $G_v$, $D_v$,
the boundary DAG, or the ridge.
Across the four full-state depths, pixel cosine remained between 0.928 and
0.945.
The selected-channel targets produced distinct outputs under the same map
family, confirming that the public terminal-subset gauge supports no-refit
channel queries.

\begin{figure*}[t]
  \centering
  \includegraphics[width=\textwidth]{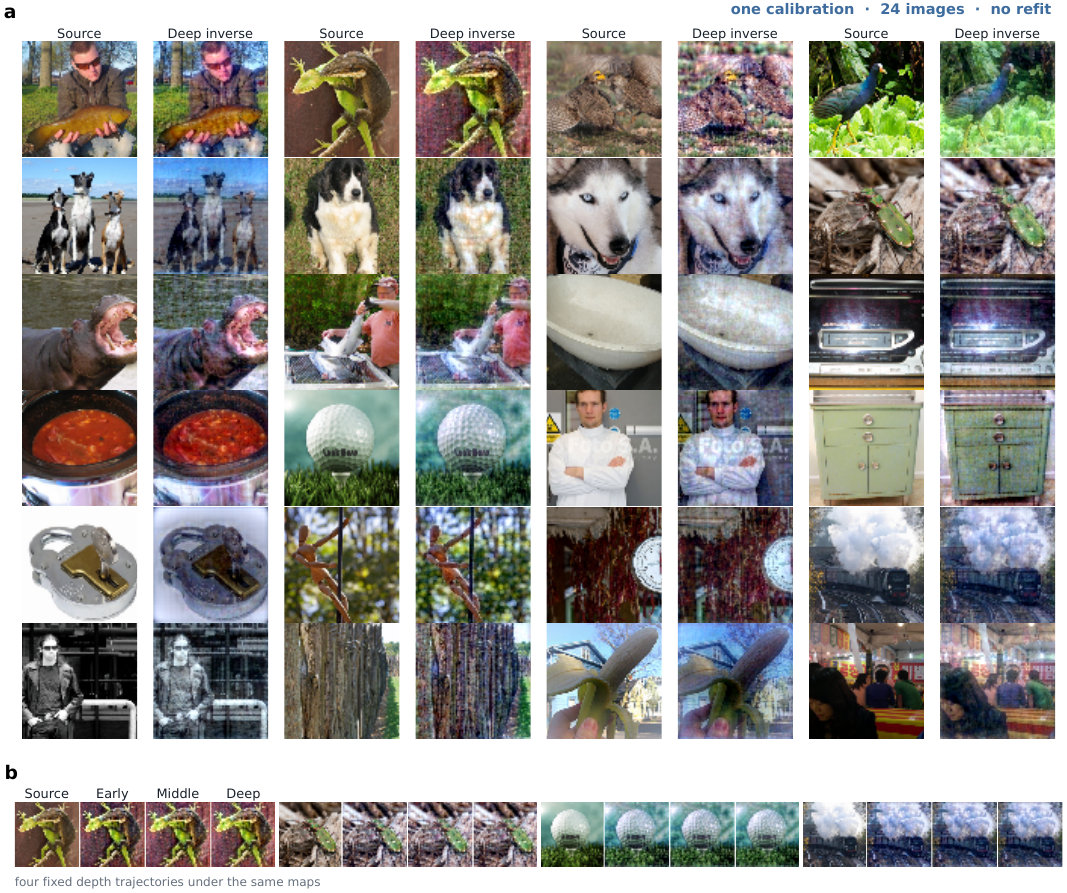}
  \caption{\textbf{Feature inversion at scale under one calibration.}
  \textbf{a}, The dominant plate contains 24 predeclared ImageNet source references paired
  with deep ResNet50 feature inverses.
  \textbf{b}, The lower strip shows four fixed examples across early, middle, and deep
  targets.
  All outputs use one calibrated map family, one public inversion path, and no
  query-specific optimisation or refitting.
  The source images are references for input-instance alignment, not unique
  inverse labels.}
  \label{fig:feature_inversion_scale}
\end{figure*}

Figure~\ref{fig:feature_inversion_scale} therefore emphasises scale and reuse
before any favourable case study.
The sample indices, ordering, target nodes, and display transform were fixed
before final rendering.
The complete fixed list, additional target depths, channel queries, and
per-image measurements are provided in
\cref{app:large_scale_atlas}.

\paragraph{Channel--coordinate queries under one calibration.}
The preceding reuse study changes the channel set $S$ while retaining every
coordinate.  We next held all 768 hidden channels fixed and changed the
coordinate set $Q$ in ViT-B/32.  At each of layers 2, 8, and 14, every one of
the 49 non-CLS patch positions was queried on 128 fixed images using one
isolated c4096 map family, for 18,816 inversions without patch-specific
refitting.  For inverse $\widehat x_{l,q}$ from queried patch $q$, we measured
the descriptive input-domain support at input patch $r$ as
\begin{equation}
  A_l(q,r)
  =
  \frac{M_r(\widehat x_{l,q})}
       {\sum_{r'}M_{r'}(\widehat x_{l,q})},
  \qquad
  M_r(\widehat x)
  =
  \sum_{(h,w)\in r}\|\widehat x_{:,h,w}\|_2 .
  \label{eq:vit_patch_support}
\end{equation}
We call $A_l$ the \emph{patch-query input-domain support matrix} and its
entries \emph{normalised inverse-magnitude support mass}.  This quantity is
neither squared inverse energy nor an attention, attribution, or causal score.

\begin{figure*}[t]
  \centering
  \includegraphics[width=\textwidth]{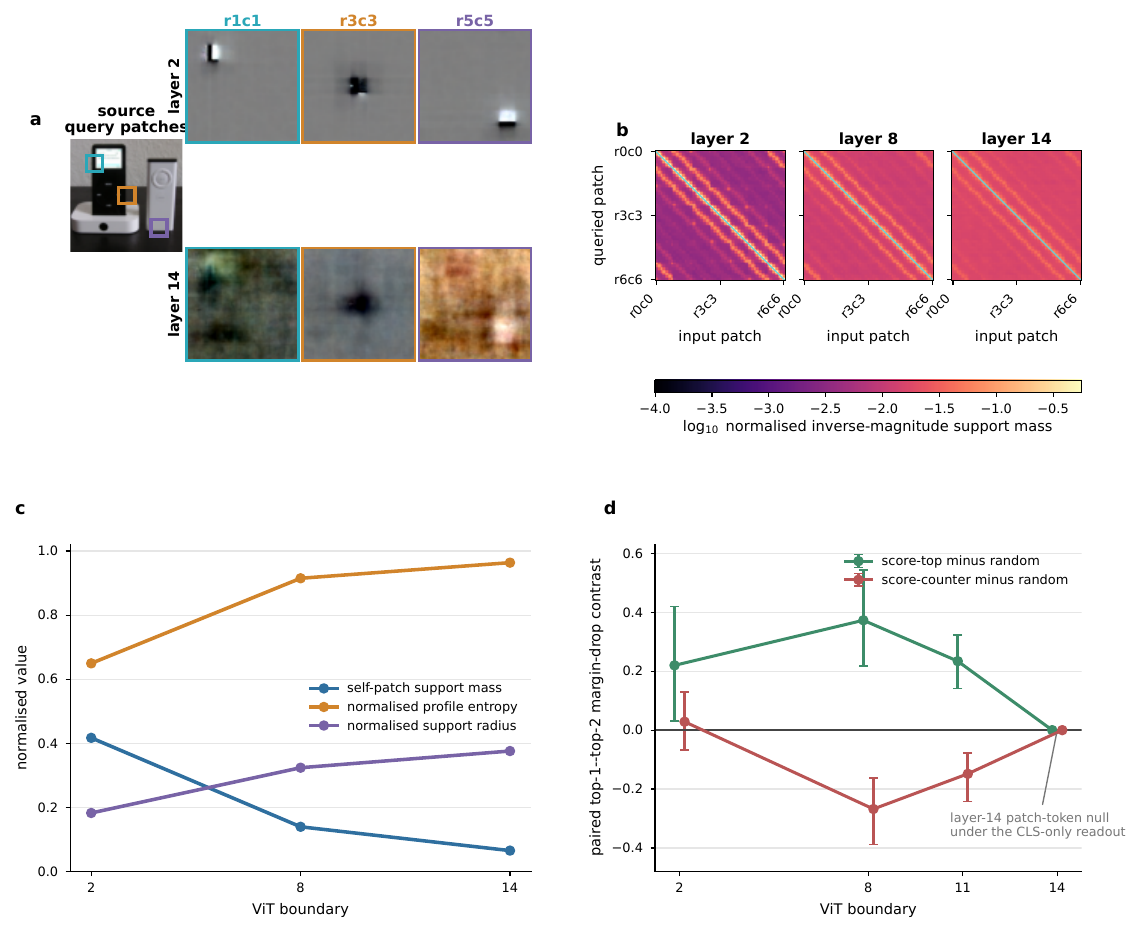}
  \caption{\textbf{One calibration supports all ViT patch-coordinate queries.}
  \textbf{a}, Representative inverses for three fixed patch queries at layers
  2 and 14 on the first fixed evaluation image.  Coloured boxes mark the query
  coordinates; the same signed visualisation rule is applied to each inverse.
  \textbf{b}, Patch-query input-domain support matrices averaged over 128
  correctly classified ImageNet validation images.  Rows index queried
  internal patches and columns index input patches; the cyan line marks equal
  coordinates.
  \textbf{c}, Mean self-patch support mass, normalised profile entropy, and
  normalised support radius; shading gives 95\% image-bootstrap intervals.
  \textbf{d}, Paired exact top-1--top-2 margin-drop contrasts after independent
  internal zero intervention on each of the 49 patch-token states.  Points are
  means and bars are 95\% image-bootstrap intervals ($n=128$).  Patch ranking uses
  gradient times activation on the original representation; the exact
  intervention is a separate forward evaluation.  Layer 14 is a patch-token
  null control under the CLS-only readout; no CLS intervention is shown.}
  \label{fig:channel_coordinate_queries}
\end{figure*}

The self-patch support mass decreased from 0.417 (95\% interval
0.413--0.421) at layer 2 to 0.139 (0.136--0.143) at layer 8 and 0.065
(0.063--0.067) at layer 14.  In parallel, normalised profile entropy increased
from 0.649 (0.645--0.654) to 0.915 (0.912--0.918) and 0.964
(0.962--0.965).  The measured input-domain support therefore changed from
strongly coordinate-aligned and local to increasingly distributed with depth;
the inverse did not remain a crop of the queried patch.

We independently zeroed every patch-token state at layers 2, 8, 11, and 14,
giving 25,088 exact interventions.  At layer 8, the score-top-minus-random
margin-drop contrast was 0.373 (95\% interval 0.217--0.544), the
score-counter-minus-random contrast was $-0.268$ ($-0.388$ to $-0.162$), and
the across-patch score--drop Spearman correlation was 0.354
(0.319--0.388).  At layer 14, every patch-token intervention was numerically
zero because the final classifier reads only the CLS token after the last
normalisation.  Thus the support matrices describe where an inverse is
expressed, while causal language is reserved for the independent internal
intervention.  Complete query atlases and intervention maps are provided in
\cref{app:vit_coordinate_queries}.

\paragraph{Comparison with iterative preimage search.}
We next compared the operational regimes on the same 16 ResNet18 images and
terminal target.
The optimisation reference used three predeclared initialisations and 2,000
Adam steps per image, with learning rate 0.05, total-variation weight 0.1,
$\ell_2$ weight 0.001, and four-pixel jitter
\citep{kingma2015adam,rudin1992nonlinear}.
It reached a target re-encoding cosine of 0.958 but had an input-instance pixel
cosine of 0.035; mean pairwise output cosine across initialisations was 0.101.
After calibration, the closed-form method reached an input-instance pixel cosine
of 0.939 on the same images without online optimisation.
These values should not be read as a strictly matched estimator comparison:
the optimiser searches for a target-consistent preimage from noise, whereas
our estimator additionally uses the target-generating sample's local network
geometry.
The result instead demonstrates that the two procedures instantiate different
inverse objects and information sets.
Table~\ref{tab:regime_comparison} reports the matched operational comparison.

\begin{table*}[t]
  \centering
  \caption{\textbf{Canonical feature-inversion regimes on a matched
  16-image ResNet18 target.}
  Pixel cosine is the mean descriptive input-instance alignment on the same
  fixed 16 images.  Online time uses sequential batch-one image queries: the
  optimisation value is the mean over eight queries, while ours is the median
  over three isolated repeats.  The optimisation reference uses 2,000 steps
  per image.}
  \label{tab:regime_comparison}
  \scriptsize
  \setlength{\tabcolsep}{4.5pt}
  \begin{tabular}{lcccc}
    \toprule
    Method & Source-local geometry & Per-query search
    & Online/query (s) & Pixel cosine \\
    \midrule
    Iterative preimage optimisation & No & 2,000 steps
      & 24.379 & 0.035 \\
    Ours, final JVP-FC & Yes & 0 steps
      & \textbf{0.139} & \textbf{0.939} \\
    \bottomrule
  \end{tabular}
\end{table*}

On sequential batch-one image queries, post-calibration application required
0.139\,s per query versus 24.379\,s for iterative optimisation, a
176$\times$ reduction in online time.
Across image, depth, channel, and mixed workloads, the closed-form path required
0.110--0.141\,s per query, whereas the 2,000-step reference required
21.728--23.578\,s.
These measurements isolate deployment-time scaling: calibration is performed
once and is characterised separately by sample demand, fitted-node count, map
storage, and peak memory.

\subsection{One Recipe Across Backbones and Tensor Components}
\label{sec:cross_architecture}

\begin{figure*}[t]
  \centering
  \includegraphics[width=\textwidth]{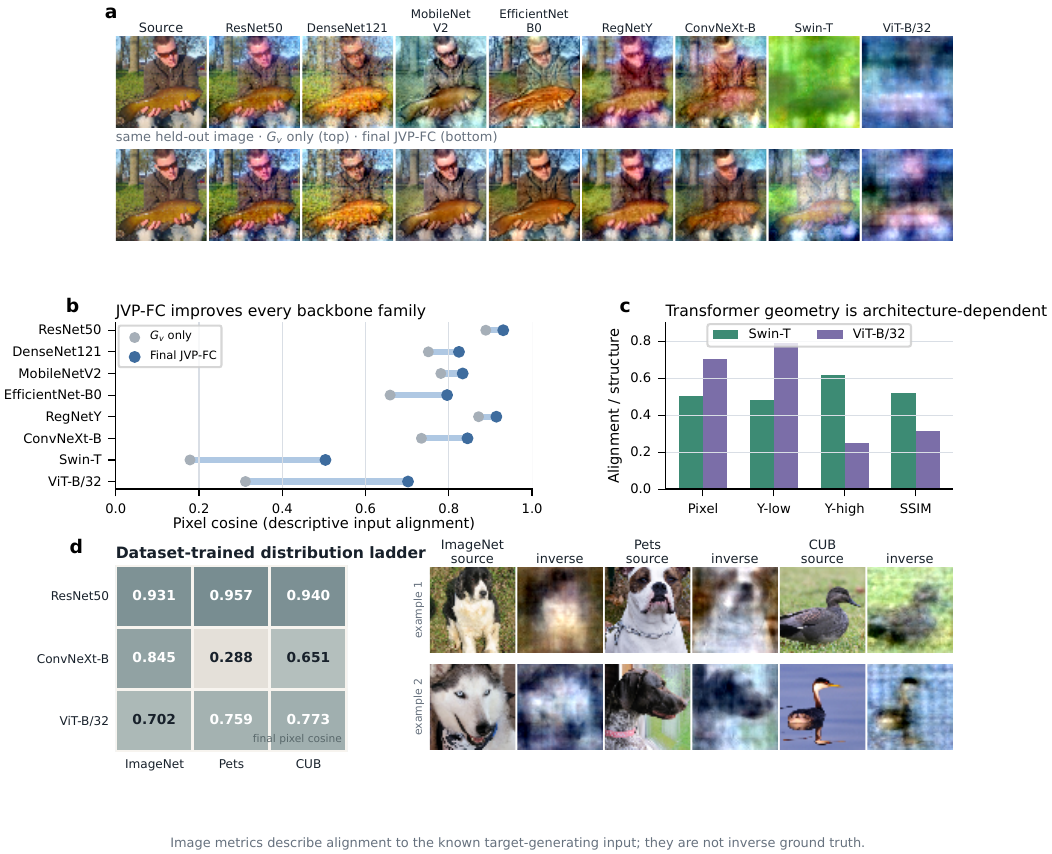}
  \caption{\textbf{Structural and distributional generality.}
  \textbf{a}, One fixed held-out image across eight backbone families under
  one recipe, with $G_v$-only outputs above the final JVP-FC outputs.
  \textbf{b}, Compact quantitative summary of input-instance alignment and
  second-stage gain.
  \textbf{c}, Swin-T and ViT-B/32 on identical images, showing
  architecture-dependent spatial and frequency structure.
  \textbf{d}, Representative ImageNet, Pets, and CUB results.
  Full per-image distributions and component-level results are reported in the
  supplement.}
  \label{fig:generality}
\end{figure*}

We evaluated the same tensor-axis rule and public inversion path on residual,
dense-connectivity, mobile, squeeze-and-excitation, modern convolutional, and
Transformer families
\citep{he2016deep,huang2017densely,sandler2018mobilenetv2,
tan2019efficientnet,radosavovic2020designing,liu2022convnet,
liu2021swin,dosovitskiy2021image}.
All nine ImageNet rows used the same 4,096/1,024 calibration/evaluation
protocol.
The final correction improved pixel alignment over $G_v$ alone for every
architecture, with gains from 0.037 on ResNet18 to 0.391 on ViT-B/32
(\cref{tab:cross_architecture}).
The effect was therefore not restricted to one CNN topology or one
normalisation regime.
Figure~\ref{fig:generality} combines the cross-architecture, Transformer, and
distribution evidence under the shared recipe.

\begin{table*}[t]
  \centering
  \caption{\textbf{One recipe across ImageNet backbone families.}
  All values are means over 1,024 validation images.
  $\Delta$ is the final-minus-$G_v$ pixel-cosine gain.
  LPIPS is lower for stronger perceptual alignment.}
  \label{tab:cross_architecture}
  \scriptsize
  \setlength{\tabcolsep}{4.1pt}
  \begin{tabular}{lccccc}
    \toprule
    Backbone & $G_v$ pixel & Final pixel & $\Delta$ & SSIM & LPIPS \\
    \midrule
    ResNet18 & 0.901 & \textbf{0.938} & 0.037 & 0.684 & 0.314 \\
    ResNet50 & 0.889 & \textbf{0.931} & 0.042 & 0.726 & 0.277 \\
    DenseNet121 & 0.751 & \textbf{0.824} & 0.074 & 0.588 & 0.419 \\
    MobileNetV2 & 0.781 & \textbf{0.833} & 0.052 & 0.710 & 0.260 \\
    EfficientNet-B0 & 0.659 & \textbf{0.796} & 0.137 & 0.722 & 0.272 \\
    RegNetY-1.6GF & 0.872 & \textbf{0.914} & 0.043 & 0.597 & 0.423 \\
    ConvNeXt-B & 0.734 & \textbf{0.845} & 0.111 & 0.617 & 0.399 \\
    Swin-T & 0.178 & \textbf{0.504} & 0.325 & 0.519 & 0.520 \\
    ViT-B/32 & 0.312 & \textbf{0.702} & 0.391 & 0.314 & 0.757 \\
    \bottomrule
  \end{tabular}
\end{table*}

For MobileNetV2 and EfficientNet-B0, residual blocks remain single units ending
at their residual adds, whereas each stage-entry non-residual block is
factorised at its architecture-native expansion and depthwise tensors before
the final projection.
MobileNetV2 uses pre-activation convolution tensors; EfficientNet-B0 uses its
post-activation spatial tensors while retaining squeeze-and-excitation gating
and multiplication inside the same local edge.
Both chains terminate at the final 320-channel feature block.
This architecture-aware boundary onboarding changes neither the estimator,
ridge, calibration size, nor online API, but raises final pixel cosine from the
incomplete block-only capture to 0.833 and 0.796, respectively.

The Swin--ViT contrast is particularly informative.
Swin-T had lower pixel cosine than ViT-B/32 (0.504 versus 0.702) but
substantially stronger SSIM and high-frequency alignment (0.519 and 0.614
versus 0.314 and 0.248).
ViT-B/32 retained stronger low-frequency alignment (0.789).
Strong spatial structure on Swin therefore rules out a blanket failure of the
closed-form method on Transformers.
The contrast instead shows that the revealed inverse geometry remains
architecture dependent.
It does not isolate patch size, hierarchy, or windowing as the sole cause.

\paragraph{Tensor-component coverage.}
The same public path was also evaluated on 11 component families, including
DenseNet concatenations, depthwise and MBConv/SE states, ConvNeXt
normalisation-adjacent tensors, Swin window and patch-merging states, ViT token
streams, and pooled vectors.
All 12 formal runs passed the component validator without a manual
component-specific inverse.
For the five ConvNeXt normalisation-adjacent targets, input-instance pixel cosine
ranged from 0.943 to 0.957 and target re-encoding cosine from 0.935 to 0.981.
The full target-node, shape, axis, and map-family audit is provided in
\cref{app:component_zoo}.

\subsection{Behaviour Across Visual Distributions}
\label{sec:distribution_behaviour}

We next separated recipe generality from fixed-map transfer.
Dataset-trained ResNet50, ConvNeXt-B, and ViT-B/32 checkpoints were calibrated
and evaluated independently on ImageNet, Pets, and CUB under the same final
recipe.
ResNet50 remained strongly input aligned across the three datasets, with final
pixel cosines of 0.931, 0.957, and 0.940.
ViT-B/32 increased from 0.702 on ImageNet to 0.759 on Pets and 0.773 on CUB.
ConvNeXt-B varied non-monotonically, from 0.845 on ImageNet to 0.288 on Pets
and 0.651 on CUB.
Although ConvNeXt-B has lower input-instance alignment on Pets and CUB, a
dedicated source-geometry intervention over 512 held-out images per dataset
showed that even its lowest-pixel-cosine quartile remained more consistent with
matched than source-shuffled Jacobian paths (path NMSE: 0.433 versus 0.719 on
CUB and 0.443 versus 0.636 on Pets; \cref{app:source_path_faithfulness}).
Thus low pixel cosine in this setting indicates weak pixel-space alignment
but does not by itself diagnose inversion failure; it can coexist with
measurable source-path faithfulness.
The data therefore support distribution- and model-instance-dependent
behaviour, not a universal scalar ordering of dataset difficulty.

The supplement further expands the distributional result in two ways.
First, an ImageNet-weight breadth study evaluates all eight families on Pets
and CUB without changing the public recipe.
Second, a $3\times3$ calibration-source transfer matrix measures distribution
transfer, while repeated calibration subsets quantify calibration-estimation
variation separately.
ResNet18 showed a small matched-versus-cross mean gap
(0.937 versus 0.932), whereas ViT-B/32 showed a larger gap
(0.721 versus 0.662).
These experiments establish architecture-dependent transfer sensitivity
without invoking a causal scalar dataset-complexity explanation.

\subsection{Topology-Respecting Two-Stage Repair Explains the Gain}
\label{sec:mechanism_experiments}

We used matched controls to isolate the method components.
A single global closed-form map did not replace local DAG composition:
pixel cosine was 0.072 for global $G$ and 0.067 after a global correction,
compared with 0.901 for local $G_v$ and 0.938 for the final local method.
Raw VJP was near zero (0.0002).
The input-instance alignment gain therefore came from topology-respecting local
composition, not merely from fitting one large linear map at the terminal
target.

\begin{figure*}[t]
  \centering
  \includegraphics[width=\textwidth]{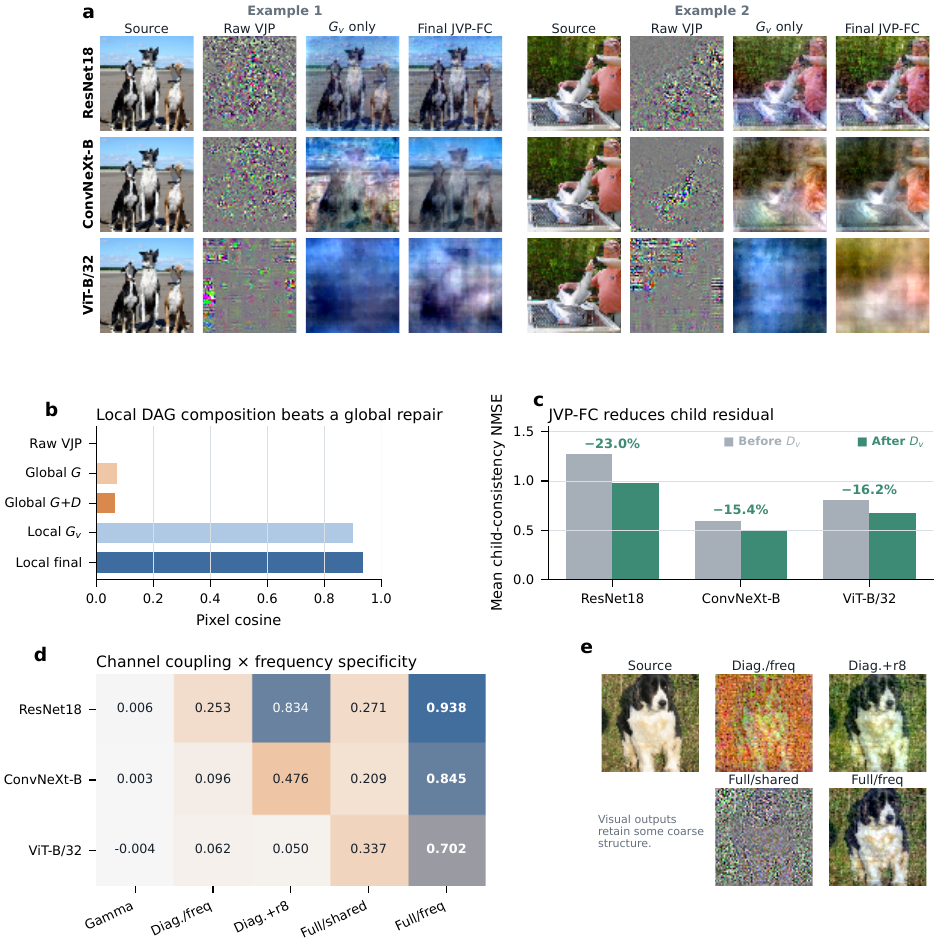}
  \caption{\textbf{Why topology-respecting two-stage repair works.}
  \textbf{a}, Raw VJP, distributional $G_v$-only repair, and final JVP-FC
  outputs on two predeclared examples per architecture.
  \textbf{b}, Global single-boundary repair versus local DAG composition.
  \textbf{c}, Child-consistency residuals before and after JVP-FC.
  \textbf{d}, Restricted operator families across three architectures.
  \textbf{e}, Matched visual examples showing why
  scalar image metrics must be interpreted together with spatial and spectral
  structure.}
  \label{fig:mechanism}
\end{figure*}

The second stage improved all three representative architectures.
On ResNet18, $G_v$ alone reached a pixel cosine of 0.901 and the second stage
added 0.037.  The corresponding gain increased to 0.111 on ConvNeXt-B and
0.391 on ViT-B/32.
Mean child-consistency NMSE fell by 23.0\%, 15.4\%, and 16.2\%,
respectively.
Intermediate-state relative $\ell_2$ decreased by 1.5\%, 0.4\%, and 5.9\%.
This architecture-level ordering is consistent with JVP-FC becoming more
important when normalisation and attention induce strongly sample-dependent
local operators.
Because architecture and component type were not independently controlled, the
comparison does not attribute the gain to any single module.
These measurements support JVP-FC as a residual repair stage, but they do not
imply monotonic improvement at every boundary.
Figure~\ref{fig:mechanism} aligns the quantitative ablations with fixed-image
outputs and boundary-level residual diagnostics.

\begin{table*}[t]
  \centering
  \caption{\textbf{Matched topology, stage, and operator ablations.}
  \textbf{a}, The topology control is a matched ResNet18 experiment.
  \textbf{b,c}, Stage and operator comparisons use matched c4096/e1024 runs
  across the three representative architectures.
  Values are pixel cosine unless marked otherwise.
  Restricted operators measure which input-aligned structures their operator
  class retains; low values do not define feature-inversion invalidity.}
  \label{tab:mechanism}
  \vspace{0.3em}
  \begin{minipage}[t]{0.43\textwidth}
    \centering
    \textbf{a\quad Topology control (ResNet18)}\\[0.25em]
    \scriptsize
    \setlength{\tabcolsep}{4.5pt}
    \begin{tabular}{lrl}
      \toprule
      Variant & Pixel & Computation \\
      \midrule
      Raw VJP & 0.0002 & adjoint transport \\
      Global $G$ & 0.0722 & one terminal map \\
      Global $G{+}D$ & 0.0671 & one terminal map \\
      Local $G_v$ & 0.9009 & boundary DAG \\
      Local final & \textbf{0.9380} & boundary DAG \\
      \bottomrule
    \end{tabular}
  \end{minipage}
  \hfill
  \begin{minipage}[t]{0.55\textwidth}
    \centering
    \textbf{b\quad Two-stage correction}\\[0.25em]
    \scriptsize
    \setlength{\tabcolsep}{4.2pt}
    \begin{tabular}{lrrrr}
      \toprule
      Model & $G_v$ & Final & $\Delta$ pixel & Child NMSE red. \\
      \midrule
      ResNet18 & 0.9009 & \textbf{0.9375} & 0.0367 & 23.0\% \\
      ConvNeXt-B & 0.7343 & \textbf{0.8450} & 0.1107 & 15.4\% \\
      ViT-B/32 & 0.3116 & \textbf{0.7021} & 0.3905 & 16.2\% \\
      \bottomrule
    \end{tabular}
  \end{minipage}

  \vspace{0.8em}
  \begin{minipage}{0.82\textwidth}
    \centering
    \textbf{c\quad Operator family}\\[0.25em]
    \scriptsize
    \setlength{\tabcolsep}{7pt}
    \begin{tabular}{lrrr}
      \toprule
      Operator & ResNet18 & ConvNeXt-B & ViT-B/32 \\
      \midrule
      Gamma-only & 0.0062 & 0.0034 & $-0.0044$ \\
      Diagonal/per-frequency & 0.2526 & 0.0961 & 0.0620 \\
      Diagonal+rank-8 & 0.8337 & 0.4763 & 0.0501 \\
      Full/shared-frequency & 0.2709 & 0.2090 & 0.3366 \\
      Full/per-frequency & \textbf{0.9375} & \textbf{0.8450} & \textbf{0.7021} \\
      \bottomrule
    \end{tabular}
  \end{minipage}
\end{table*}

The operator ablation was decisive.
Gamma-only maps had near-zero global input-instance alignment on all three
architectures.
Per-frequency diagonal maps reached 0.253, 0.096, and 0.062, while retaining
some visible low-frequency or spatial structure in selected cases.
A rank-8 channel correction was useful on ResNet18 (0.834) but not on ViT-B/32
(0.050).
Full channel rank with one shared-frequency map also remained far below the
final operator.
Thus cross-channel and frequency-specific second-order structure are both
substantive, while low-rank compression is architecture dependent.

\subsection{Target Features and Source-Local Geometry Jointly Determine the Inverse}
\label{sec:inverse_meaning}

\begin{figure*}[t]
  \centering
  \includegraphics[width=\textwidth]{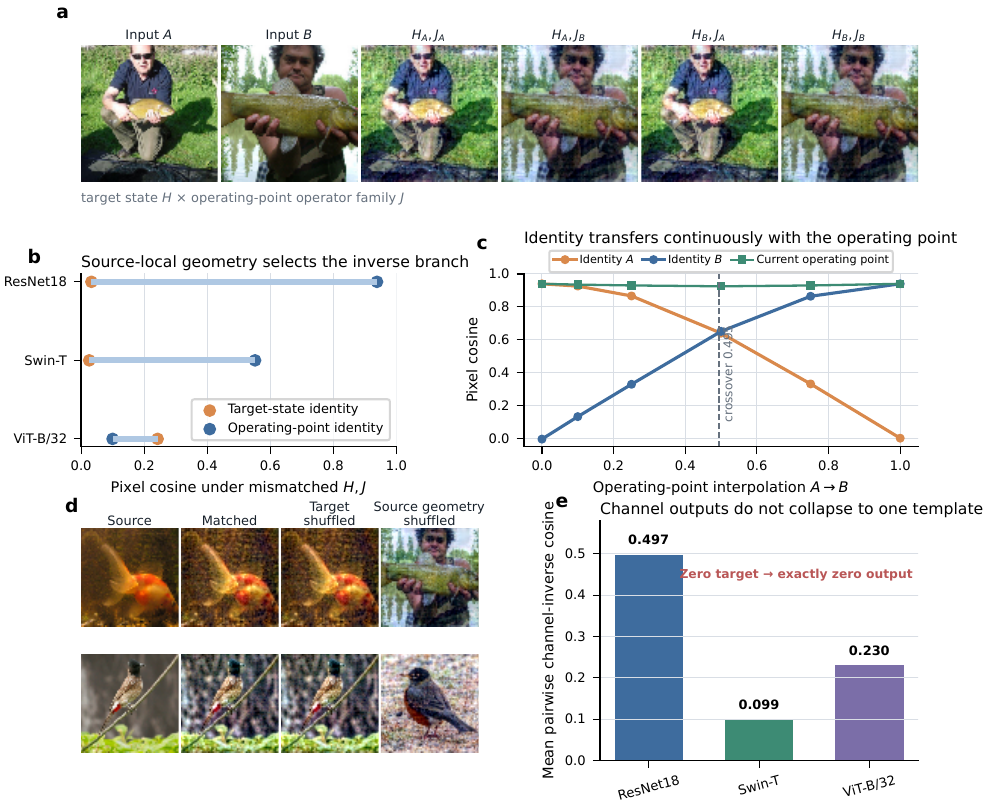}
  \caption{\textbf{What determines a feature inverse.}
  \textbf{a}, Target--operator swap for one fixed pair.
  \textbf{b}, Similarity to target-state and operating-point identities under
  mismatched conditions.
  \textbf{c}, Continuous operating-point interpolation with a fixed target.
  \textbf{d}, Matched, target-shuffled, and source-geometry-shuffled controls.
  \textbf{e}, Channel diversity and the exact zero-target result.
  The controls show that source-local geometry selects the inverse branch,
  while the terminal state modulates that branch in an
  architecture-dependent manner.}
  \label{fig:inverse_determinants}
\end{figure*}

The method is intentionally conditioned on the forward point of the input
being explained.
We therefore tested target identity and operating-point geometry separately.
For two images $A$ and $B$, we crossed terminal states $H_A,H_B$ with local
operator families $J_A,J_B$.
On ResNet18, both mismatched conditions followed the operating-point identity:
$H_A$ with $J_B$ had pixel cosine 0.939 to $B$ and 0.034 to $A$,
while $H_B$ with $J_A$ had cosine 0.939 to $A$ and 0.032 to $B$.
Swin-T showed the same direction with weaker invariance.
ViT-B/32 showed stronger terminal-state modulation, but source shuffling still
followed the new operating point (0.692) rather than the old input (0.016).
Figure~\ref{fig:inverse_determinants} visualises these matched, swapped,
interpolated, and zero-target conditions.

Several controls exclude simpler explanations.
Every zero target produced exactly zero output, as required by the
zero-intercept maps.
Changing the source geometry moved the inverse to the new operating point
rather than preserving a fixed image template.
Channel-specific outputs did not collapse to a single result: mean pairwise
pixel cosine was 0.497 on ResNet18, 0.230 on ViT-B/32, and 0.099 on Swin-T.
A continuous ResNet18 interpolation experiment gave a monotone identity
transfer with crossover at 0.495, whereas same-image noise up to 0.1 changed
input-instance pixel cosine by at most 0.00066.
Together, these results identify feature inversion as a target-modulated,
source-local branch computation.
They do not make similarity to the original image the ground truth of the
inverse.

\subsection{Prediction-Conditioned Atlases Link Feature Structure to Decisions}
\label{sec:decision_conditioned}

\begin{figure*}[t]
  \centering
  \includegraphics[width=\textwidth]{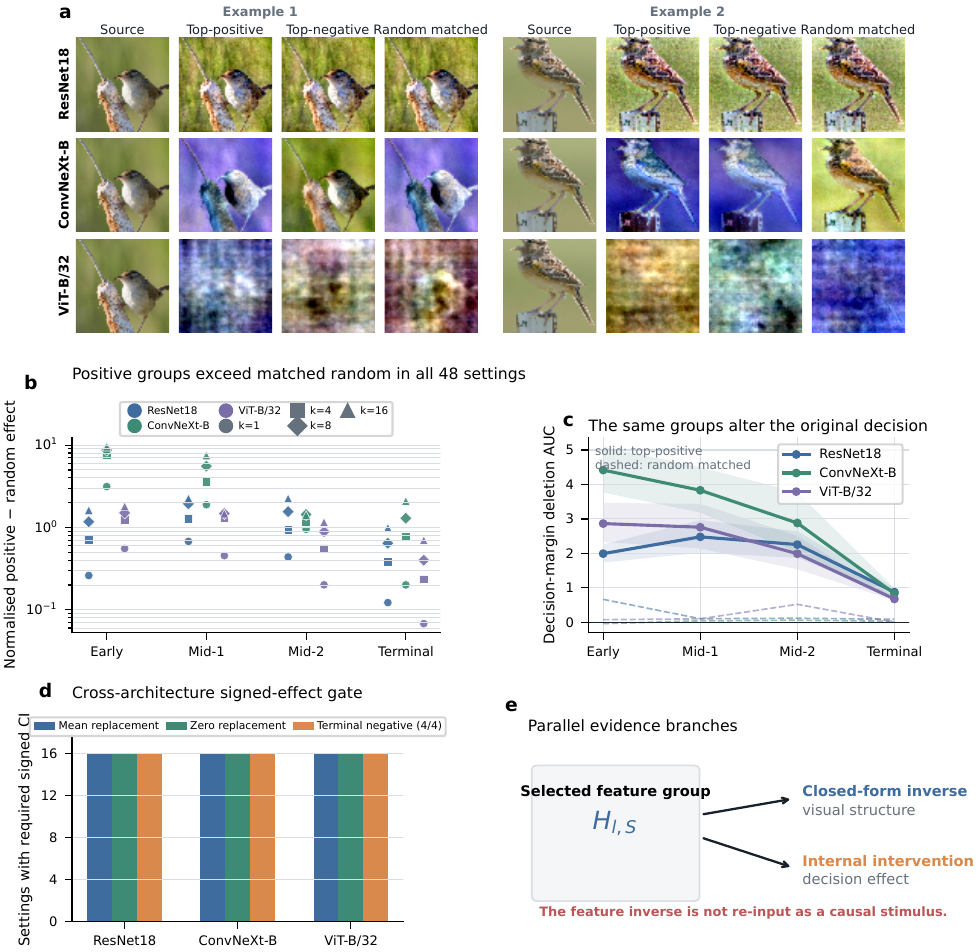}
  \caption{\textbf{Decision-grounded feature atlases across architectures.}
  \textbf{a}, Two predeclared source examples with top-positive, top-negative,
  and random-matched feature inverses for ResNet18, ConvNeXt-B, and ViT-B/32.
  \textbf{b}, Normalised positive-minus-random effects over all
  model--depth--group-size settings.
  \textbf{c}, Decision-margin deletion AUC across depth.
  \textbf{d}, Signed cross-architecture validation gate.
  \textbf{e}, Parallel inversion and internal-intervention evidence branches.
  The feature inverse is not re-input as a causal stimulus; causal evidence comes
  from intervention on the original internal representation.}
  \label{fig:decision_atlas}
\end{figure*}

The previous experiments establish a reusable representation-level interface.
We next tested whether the same visualised feature groups were related to the
current prediction.
For each of 16 fixed correctly classified CUB images per model, channels were
ranked by the activation--gradient contribution to the predicted margin
\citep{ancona2018towards}.
At four predeclared depths, we selected nested groups of
$k\in\{1,4,8,16\}$ top-positive, top-negative, activation-matched, and
20-repeat random-matched channels.
Each group was inverted with the same c4096 map family and no refitting.
The original internal representation was then intervened on independently,
using calibration-mean replacement as the primary protocol and zero
replacement as a robustness check.
Figure~\ref{fig:decision_atlas} presents the inversion and intervention branches
as parallel evidence rather than a single causal-image pipeline.

The strict cross-architecture audit covered 36,864 intervention rows,
3,072 group inversions, and 1,536 deletion curves.
Under calibration-mean replacement, the top-positive margin drop exceeded the
within-image mean of the random-matched controls in all 48
model--layer--$k$ settings, and every paired bootstrap interval excluded zero.
Zero replacement reproduced the same 48/48 direction.
All 12 terminal-layer top-negative controls had the expected opposite signed
effect.
The evidence chain was therefore consistent across residual CNN,
normalisation-heavy CNN, and global-token Transformer families, although their
visible feature-inverse geometries differed substantially.

This experiment supports a bounded decision-grounded interpretation:
the visualised feature groups have an independently measured effect on the
current decision.
It does not imply that the feature inverse itself causes the decision change, or
that every visible structure has a unique named semantic label.
Expanded image plates, deletion curves, and signed controls are reported in
\cref{app:decision_atlas}.

\subsection{Calibration and Deployment Define the Practical Boundary}
\label{sec:deployment_boundary}

Calibration demand was strongly architecture dependent.
For ViT-B/32, final pixel cosine increased from
0.129, 0.363, and 0.648 at 512, 1,024, and 2,048 calibration images to
0.702 at 4,096 and 0.724 at 8,192.
Swin-T followed a similar diminishing-return pattern:
0.047, 0.290, 0.457, 0.504, and 0.526.
Both vision token models showed their largest observed transition between
1,024 and 2,048 images, while doubling 4,096 to 8,192 added only about 0.022.
We therefore use 4,096 as a representative diminishing-return operating point
for these token architectures, not as a universal optimum.
Table~\ref{tab:deployment} reports post-calibration online application,
throughput, map storage, and evaluation memory for representative
architectures.

\begin{table}[!htbp]
  \centering
  \caption{\textbf{Representative post-calibration online deployment costs.}
  All online values are measured after calibration on the c4096/e1024 quality
  runs.  Online time is the measured 1,024-image application total divided by
  1,024 and is therefore not isolated batch-one latency.  Throughput is its
  reciprocal.  Map storage includes both $G_v$ and $D_v$; evaluation peak is
  the measured peak allocated memory.  These values describe the current
  public implementation rather than an optimised-kernel lower bound.}
  \label{tab:deployment}
  \scriptsize
  \setlength{\tabcolsep}{5.0pt}
  \begin{tabular}{lrrrrr}
    \toprule
    Model & Fitted nodes & Online (ms/image) & Throughput (images/s)
      & Map storage (GiB) & Eval peak (GiB) \\
    \midrule
    ResNet18 & 9 & 58.4 & 17.1 & 0.83 & 1.71 \\
    ConvNeXt-B & 37 & 369.7 & 2.70 & 15.51 & 16.78 \\
    Swin-T & 26 & 129.0 & 7.75 & 5.98 & 6.56 \\
    ViT-B/32 & 14 & 105.1 & 9.52 & 2.99 & 3.61 \\
    \bottomrule
  \end{tabular}
\end{table}

Post-calibration application ranged from 58.4\,ms/image for ResNet18 to
369.7\,ms/image for ConvNeXt-B in the current implementation.
The larger ConvNeXt-B deployment cost accompanies 37 fitted nodes and a
15.51-GiB map family, compared with 9 nodes and 0.83\,GiB for ResNet18.
This comparison therefore measures the deployed reverse pass actually used by
each architecture, while leaving hardware-specific calibration engineering
outside the cross-architecture timing claim.

The default boundary policy also reflects a quality--cost trade-off.
At matched c512/e1024 on ResNet18, a full-DAG fit increased pixel cosine from
0.929 to 0.963 and expanded the fitted nodes from 9 to 49.
The corresponding post-calibration online application increased from 53.9 to
74.7\,ms/image, exposing a direct quality--deployment trade-off for the chosen
boundary factorisation.
Ridge sensitivity was mild for ResNet18 over
$\rho\in\{0.003,0.01,0.03\}$, but much stronger for ViT-B/32 at low
calibration.
The complete ridge, boundary, storage, peak-memory, repeated-subset, and
true-child diagnostics are provided in the supplement.

These resource results define the practical boundary of the paradigm change.
The method removes query-specific iterative preimage optimisation and enables
repeated layer/channel queries after calibration, but it does not make
calibration or map storage negligible.
The main practical limits are high-dimensional channel covariance estimation,
large full-channel map storage, and the quality--cost trade-off of boundary
factorisation.
These measurements delimit the evaluated visual deployment regime.

\subsection{Beyond Vision: Source-Grounded Feature Inversion in a Canonical Language Transformer}
\label{sec:language_extension}

The vision experiments establish a reusable interface for asking what a frozen
feature identifies in the input domain, but they do not establish that the same
computation remains meaningful for a causal token sequence.
We therefore applied the same source-grounded closed-form construction to
GPT-2 small, a canonical decoder-only language Transformer
\citep{radford2019language}.
The purpose was not to reconstruct the original sentence or generate fluent
text.
It was to test whether a genuinely transformed internal token representation
could be mapped to the continuous input-embedding domain, and whether that
inverse exposed lexical, contextual, and syntactic evidence carried by the
representation.

We represented a length-$L=16$ input as a $B\times768\times L$ embedding
tensor, using hidden width as the channel axis and token position as the
coordinate axis.
The target was the layer-6 attention-module output token stream rather than the
residual stream or an attention-logit/weight matrix.
The formal query retained every hidden channel at the terminal position,
\begin{equation}
  S=\{1,\ldots,768\},
  \qquad
  Q=\{L-1\},
  \qquad
  y_T=P_{S,Q}H_T,
  \qquad
  s_T^{\mathrm{on}}=\frac{P_{S,Q}H_T}{768}.
  \label{eq:language_terminal}
\end{equation}
The reverse path used the frozen source-local graph from input embeddings
through the outputs of blocks 2 and 4 to the layer-6 attention output.
Zero-intercept full-channel $G_v$ and $D_v$ maps were fitted once on two
independently sampled sets of 4,096 WikiText-2 windows (seeds 123 and 321)
\citep{merity2016pointer}.
Both stages used all-shared spectral tying over the nine stored rFFT bins.
Evaluation used no per-example optimisation, map refitting, learned text
decoder, or autoregressive decoding.
Layer 6 was identified as the cue-minus-random peak in an exploratory six-layer
sweep rather than specified in advance.  The cluster-bootstrap intervals below
are conditional on this selected layer and quantify variation across words or
templates; they do not adjust for the layer search.

For an inverse embedding sequence $\widehat e_p(q)$ from terminal query $p$, we
defined the token-position evidence profile
\begin{equation}
  \mathcal E_{p\rightarrow q}
  =
  \frac{\|\widehat e_p(q)\|_2^2}
       {\sum_{r=1}^{L}\|\widehat e_p(r)\|_2^2}.
  \label{eq:language_inverse_energy}
\end{equation}
We call this quantity \emph{inverse energy}.
It is computed after the complete closed-form reverse path and is neither an
attention weight nor a decoded-token probability.

We first tested semantic routing in 72 controlled contexts covering 12
polysemous terminal words, two senses per word, and three contexts per sense.
Each context contained three to five predefined semantic cues, and 64 random
subsets matched both cue count and the empirical cue-position distribution.
After averaging the two calibration outputs, Raw VJP placed 0.511 of its
inverse energy on the terminal token and 0.259 on semantic cues.
Final JVP-FC reduced terminal-token mass to 0.169 and increased cue mass to
0.350, compared with matched-random mass 0.160.
At this exploratory layer-6 peak, the resulting Final cue-minus-random effect
was 0.190 (95\% word-cluster bootstrap interval 0.096--0.297, conditional on
the selected layer; 12 words and $n=72$ contexts).
Relative to Raw VJP, Final increased cue mass by 0.091 (0.024--0.158) and
reduced terminal-token mass by 0.342 (0.294--0.389).
The corresponding cue-minus-random change was 0.012 and was not resolved from
zero ($-0.064$--0.097).
The supported conclusion is therefore redistribution from terminal-token
identity toward a cue-selective contextual profile, rather than significant
superiority over Raw VJP on that scalar contrast.

\begin{figure*}[p]
  \centering
  \includegraphics[width=0.93\textwidth]{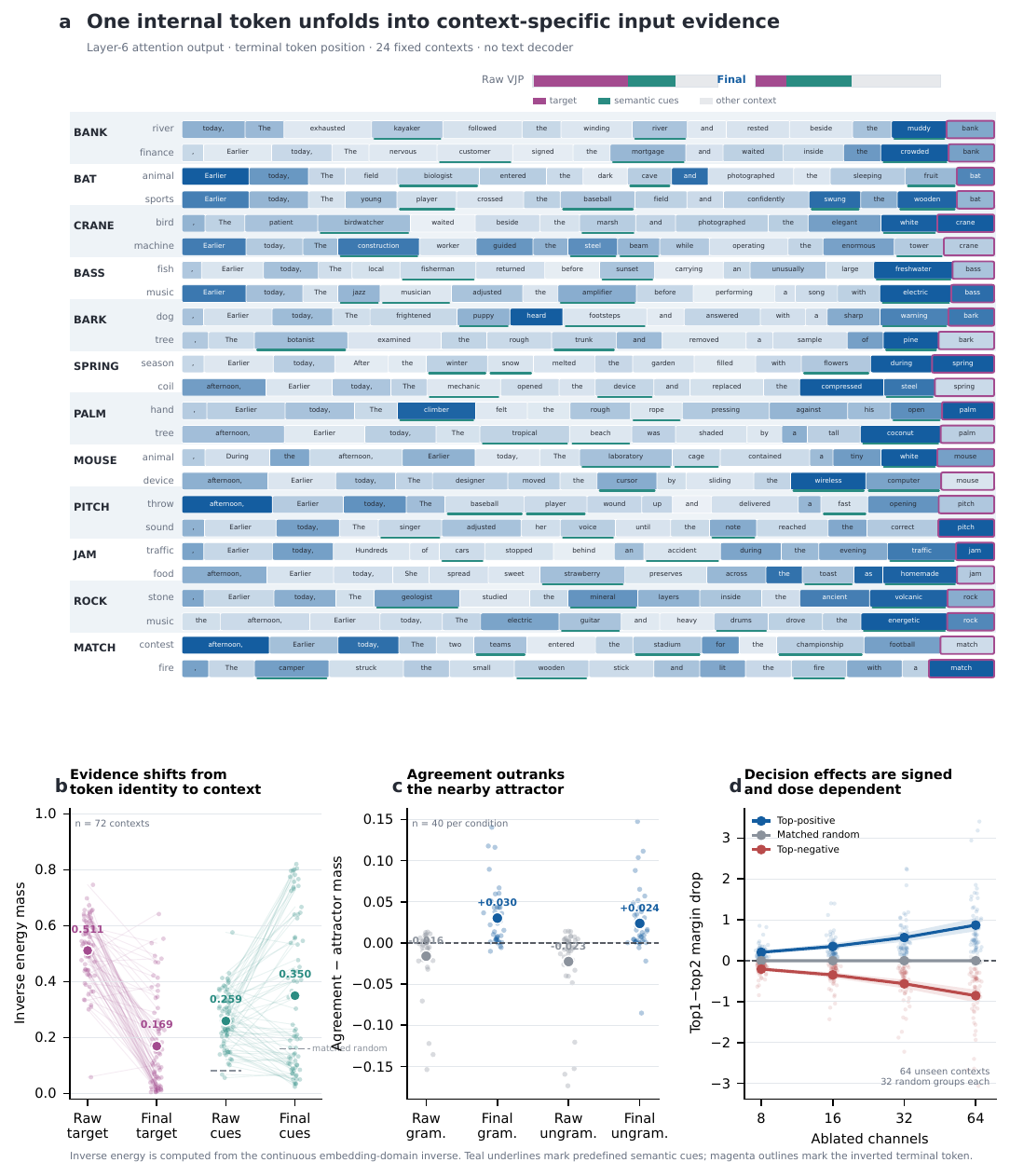}
  \caption{\textbf{Token-resolved source-grounded inversion in GPT-2.}
  The target is the complete layer-6 attention-output vector at the terminal
  token position.  Layer 6 was identified as the cue-minus-random peak in an
  exploratory six-layer sweep, so the reported benchmark intervals are
  conditional on this selected layer.  Panels a--c use continuous
  input-embedding inverses and no
  text decoder; panel d reports an independent intervention on the original
  representation.
  \textbf{a}, Rows show Final inverse energy for one fixed benchmark-order
  context from each sense of all 12 polysemous words; the stacked header
  compares mean Raw VJP and Final allocations.  Fill intensity is inverse
  energy averaged over two
  independently fitted c4096 map families; teal underlines mark predefined
  semantic cues and magenta outlines mark the inverted terminal token.
  \textbf{b}, Per-context target-token and semantic-cue mass for Raw VJP and
  Final JVP-FC ($n=72$); dashed grey marks position-matched random controls.
  \textbf{c}, Agreement-minus-attractor mass for 40 grammatical and 40
  ungrammatical number-agreement controls.
  \textbf{d}, Internal intervention on decision-conditioned channel groups over
  64 unseen WikiText contexts.  Lines show means and 95\% context-bootstrap
  intervals; faint points are individual contexts.  The intervention acts on
  the original internal representation, not on a rendered inverse.}
  \label{fig:language_main}
\end{figure*}

We next constructed 80 controlled number-agreement contexts from 20 lexical
templates.
Each template crossed singular and plural heads with grammatical and
ungrammatical auxiliaries while a nearer noun carried the opposite number.
Raw VJP assigned less energy to the agreement position than to the attractor,
with grammatical and ungrammatical agreement-minus-attractor effects of
$-0.016$ and $-0.023$.
Final reversed both signs to $+0.030$ (95\% template-cluster interval
0.017--0.045) and $+0.024$ (0.009--0.039).
Pooled across all four variants per template, Final improved on Raw VJP by
0.046 (0.027--0.066), but its effect was 0.031 lower than $G_v$ alone
(Final-minus-$G_v$ interval $-0.043$ to $-0.019$).
The two fitted stages therefore provided complementary repairs rather than a
guarantee that every language diagnostic improved monotonically.
Figure~\ref{fig:language_main} combines the fixed semantic atlas, per-context
redistribution, number-agreement controls, and the independent decision
intervention.

No reference simultaneously reproduced the cue-selective semantic profile and
agreement routing.
The direct target-only global linear reference and layer-6 attention gave
cue-minus-random effects of 0.004 and 0.006.
Gradient-times-input gave a larger contextual effect of 0.173 but favoured the
opposite-number attractor, with a grammatical agreement-to-attractor ratio of
0.559; integrated gradients gave 0.088 and 0.742, respectively.
Final jointly produced a 0.190 contextual effect and a descriptive agreement
ratio of 7.57, while $G_v$ alone gave a ratio of 13.42.
Attention and gradient methods are interpretability references rather than
feature-inversion baselines, and the direct global map does not use the
source-local DAG.

Source-local controls separated the operating-point geometry from terminal
content.
For 24 fixed cross-context pairs, the inverse of terminal state $A$ evaluated
on geometry $B$ was closer to the geometry-$B$ inverse than to the target-$A$
inverse (continuous cosine 0.496 versus 0.189; energy-profile cosine 0.623
versus 0.373).
Every zero terminal state produced exactly zero output, and replacing the
terminal query with a predefined cue-position query changed both the continuous
inverse and its profile (cosines 0.079 and 0.182).
The layer-6 result also reproduced across the independent calibrations:
continuous-inverse cosine was 0.912, profile cosine was 0.913, and top-five
evidence-position overlap was 82.4\% over all 152 controlled examples.

Finally, we connected decision-conditioned feature groups to GPT-2's current
prediction through a separate intervention branch.
For 64 unseen WikiText contexts, layer-6 attention-output channels were ranked
by their activation--gradient contribution to the top1--top2 logit margin.
At group sizes $k\in\{8,16,32,64\}$, top-positive, top-negative, and 32 random
groups were inverted with the same maps, while causal evidence was measured by
ablating the group inside the original representation.
At $k=32$, top-positive ablation reduced the margin by 0.567 (95\% interval
0.463--0.680), the mean of the per-context random medians was 0.0001
($-0.0033$--0.0038), and top-negative ablation had the opposite signed effect
($-0.564$).
The top-positive mean drop increased from 0.206 to 0.349, 0.567, and 0.871 over
the four group sizes, while random effects remained centred near zero.
The inverse was never re-input as a causal stimulus.

As a supplementary query control, we also held each decision-conditioned
32-channel set fixed while expanding the GPT-2 coordinate query from the
terminal token to the complete token stream.  The joint full-coordinate query
changed the inverse profile and the signed intervention effect without map
refitting; it is not an isolated nonterminal-position query
(\cref{fig:language_supp_full_coordinate}).

Together, these controlled results show that one calibrated reverse path can
turn a selected internal token representation into a token-resolved continuous
input-embedding inverse, reuse it across semantic and syntactic queries, and
pair it with an independent intervention on the corresponding feature group.
The main semantic and syntactic scope is GPT-2 small, 16-token causal contexts,
a terminal-token query, and the layer-6 attention output.  The supplementary
joint full-coordinate channel-subset query broadens the tested query family but
does not establish isolated nonterminal-position inversion, sentence
reconstruction, or universal validity across language-model architectures.
Within these boundaries, the experiment provides constructive evidence that
source-grounded feature inversion is not intrinsically tied to vision.

\clearpage

\section{Discussion and Limitations}
\label{sec:limitations}

The central implication of the target--operator controls across both modalities
is that source conditioning is part of the feature-inversion object rather than
an auxiliary input cue.  A target tensor produced by a non-injective nonlinear
network does not by itself identify the local branch from which it arose.  The
forward operating point supplies that missing condition through its
activations, nonlinear regimes, routing states, and local Jacobians.  Across
visual target--operator swaps and language cross-context geometry swaps,
changing the operating point redirected the inverse toward the new source-local
branch, whereas changing the terminal state, channel set, or coordinate query
modulated the content transported through that branch.  This behaviour is not
explained by a source-copy shortcut: zero targets give exactly zero outputs,
channel and coordinate selections remain distinct under the same operating
point, and the calibrated maps fail across mismatched checkpoints.  Taken
together, these controls support the interpretation of a feature inverse as a
target-modulated upstream-state estimate traced through the local computation
that generated it, rather than as an input-domain state determined by the
target tensor alone.

The architecture-dependent outputs are therefore an interpretability result,
not evidence of non-uniform reconstruction quality.  Convolutional models
typically expose coherent spatial and high-frequency structure, whereas
ViT-B/32 retains a stronger patch-grid, low-frequency, and globally mixed
inverse geometry.  Swin-T shows substantially stronger spatial and
high-frequency organisation than ViT-B/32 despite lower pixel cosine, ruling
out a blanket failure of the closed-form estimator on Transformers.  This
contrast is consistent with the different image biases introduced by global
patch-token mixing and hierarchical windowed computation
\citep{dosovitskiy2021image,liu2021swin}, although the current experiments do
not isolate patch size, hierarchy, or windowing as a unique cause.  More
generally, similarity to the target-generating image remains an
input-instance alignment diagnostic rather than the definition of correctness.
A lower ConvNeXt-B pixel cosine on Pets and CUB is therefore not by itself
evidence of failed inversion: the independently fine-tuned checkpoints remain
accurate, and even their lowest-alignment quartiles are more consistent with
matched than source-shuffled Jacobian paths.  This control identifies weak
pixel-space alignment without discarding the source-local computation captured
by the inverse.
A weak or visually sparse single-channel inverse may indicate that the selected
channel carries little independently visible structure, or that its evidence is
distributed across a channel combination; it does not by itself constitute an
inversion failure.

The ViT patch-query experiment makes the coordinate component of the terminal
query explicit.  At shallow boundaries, the patch-query input-domain support
matrix is strongly coordinate aligned; at deeper boundaries, the same
normalised inverse-magnitude support becomes increasingly distributed.  This
observation shows that a patch-selected inverse is not simply a crop operator.
The distributed support is consistent with contextual mixing, but is not by
itself an attention, attribution, semantic, or causal measure.  Decision
evidence instead comes from the independent internal intervention: layer-8 and
layer-11 patch-token states show signed margin effects, whereas layer-14
patch-token interventions are numerically null under the CLS-only readout.

The controlled GPT-2 experiment extends this interpretation to a causal token
sequence without making the deployed operator configurations identical across
modalities.  The core two-stage construction is unchanged, while the spectral
partition is instantiated according to the coordinate geometry: singleton
per-frequency maps in vision and all-shared tying in the reported GPT-2
setting.  Layer 6 was identified as the contextual peak in an exploratory depth
sweep, rather than specified in advance or established as universally optimal.
The formal language benchmarks use controlled, author-constructed semantic and
agreement contexts, complemented by held-out WikiText intervention contexts.
The semantic result is best interpreted as a redistribution from terminal-token
identity toward a cue-selective contextual profile, not as a resolved
Final-over-Raw improvement on cue-minus-random.  Likewise, Final improves pooled
agreement routing over Raw VJP but remains below $G_v$ alone on that diagnostic.
The two fitted stages therefore provide complementary repairs rather than a
monotonic ranking on every language measure.  Within these conditions, the
joint semantic and syntactic evidence, which is not simultaneously reproduced
by the direct global map or attribution references, provides constructive
evidence that source-grounded feature inversion is not intrinsically tied to
visual coordinates.

Across both modalities, prediction-conditioned atlases give the inverse a
specific role in model interpretation.  The inversion branch reveals the
input-domain structure selected by a layer, channel, or channel group, while
the independent intervention branch measures how the same internal feature
group affects the current prediction.  Visual top-positive groups exceed
matched random controls throughout the tested model--layer--group-size settings,
and the GPT-2 interventions separate top-positive, random, and top-negative
groups with the expected signed effects.  The two branches nevertheless answer
different questions.  The inverse is not re-input as a causal stimulus, so
causality is attributed only to the intervention on the original
representation.  Nor does the method imply that every revealed structure has a
unique human-semantic name.  It instead provides a representation-level view of
what the model extracts that can be paired with an independently grounded test
of how the corresponding internal feature group participates in the decision.

The main limitations arise from the deployed approximation to the ideal local
inverse.  The metric-aware operator $M_v^{-1}J^*M_u$ is sample and boundary
dependent, whereas the practical estimator replaces explicit online metric
estimation with fixed zero-intercept Wiener projections fitted over a
calibration distribution.  JVP-FC exposes and repairs predictable residuals at
the actual operating point, especially for sample-dependent components, but it
does not guarantee exact recovery of the online metrics or removal of every
local mismatch.  The larger online-child oracle gap observed for ViT-B/32
identifies efficient sample-specific metric or state estimation as a natural
next direction.  The repair maps are also frequency-diagonal over coordinate
axes and full-matrix over channels.  This makes calibration tractable while
excluding direct cross-frequency mixing inside a repair map, and it creates
architecture-dependent calibration, storage, and online-application costs.
Because the maps encode checkpoint-specific local geometry, a changed
checkpoint requires recalibration; the cross-checkpoint failures show that the
estimator does not behave as a checkpoint-agnostic target-only decoder.  More
efficient metric-aware updates and validated structured operator families could
reduce these costs without discarding source-local geometry.

The theoretical construction is defined on frozen finite differentiable
computational DAGs exposed by autograd, with declared tensor boundaries and
channel--coordinate axes, but the empirical evidence is narrower.
It covers the studied CNNs and vision Transformers, including all 49 non-CLS
patch-coordinate queries in ViT-B/32, together with GPT-2 small.  The formal
semantic and syntactic language-routing evidence uses terminal-token queries in
16-token author-constructed contexts at the exploratory layer-6 contextual
peak, complemented by held-out WikiText intervention contexts.  A supplementary
full-coordinate channel-subset control jointly selects $Q_{\mathrm{all}}$; it
does not validate isolated nonterminal-position queries.  For such isolated
queries, the deployed operator family can allocate substantial inverse energy
to future positions.
This diagnostic failure bounds the current spectral deployment surrogate rather
than the source-grounded DAG formulation itself.  Arbitrary nonterminal token
queries, internal attention-logit or attention-weight matrices, substantially
longer contexts, and transfer to other language-model architectures therefore
remain outside the present empirical validation.  The supplementary
full-sequence MLP-output experiment broadens continuous inversion feasibility,
but its optional renderer is only a display interface and does not establish
sentence reconstruction.  Causality- and position-structured deployed
operators, more efficient sample-specific metric estimation, and validation on
additional differentiable architectures are concrete next steps.

\section{Conclusion}
\label{sec:conclusion}

This work recasts feature inversion from target-matched preimage search as
source-grounded upstream-state estimation on the target-generating computational
DAG.  The selected feature and the target-generating sample's local network
geometry form a joint inversion object.  Along the actual DAG, a first
closed-form matrix Wiener map converts a mean-seed VJP adjoint into an
upstream-state estimate, and a second map corrects its JVP forward-consistency
residual.  Composing the repaired states in one finite reverse pass yields an
inverse interface that, after calibration, supports new inputs, depths,
and tested channel--coordinate queries without query-specific optimisation.

Across CNNs and vision Transformers, the method reveals architecture-dependent
spatial and frequency organisations over diverse tensor components and visual
distributions.  One ViT calibration supports all 49 non-CLS patch-coordinate
queries, whose input-domain support changes from locally aligned to distributed
with depth; independent interventions separately establish the decision effects
of the queried internal patch-token states.  In a controlled GPT-2 small
terminal-token setting, the same
core construction, instantiated with a modality-appropriate spectral partition,
produces token-resolved continuous input-embedding inverses without a text
decoder.  A supplementary joint full-coordinate channel-subset query broadens
the tested query family without establishing isolated nonterminal-position
inversion.
Target--operator, zero-target, and query controls show that the inverse is
jointly determined by the selected representation and source-local operators,
while independent interventions connect the corresponding internal feature
groups to model decisions without treating the inverse as a causal stimulus.
The language evidence provides constructive validation beyond vision but does
not establish isolated nonterminal-position inversion, sentence reconstruction,
or universal validity across language-model architectures.  Source-grounded
feature inversion therefore offers a reusable, modality-native interface for
revealing what selected internal representations identify in the input domain
and how the corresponding evidence shapes model decisions.  It does so by
tracing representations through the local computation that produced them,
rather than by searching for inputs that merely reproduce their target values.

\appendix

\section{Supplementary Method Details}
\label{app:method_details}

This appendix expands the method definition in \cref{sec:method} and fixes the
axis, seed, stored-frequency tying, topology, and closed-form solve conventions
needed for reproducibility.
It proceeds from notation and mean-seed construction to the two calibration
solves, the online recursion, topology and boundary instantiation, the reported
GPT-2 instance, restricted operator families, complexity, and the final
reproducibility checklist.

\subsection{Notation and Axis Convention}
\label{app:notation_axes}

For each selected tensor node $v$ in the computational DAG, we use the following
objects:
\begin{itemize}
  \item $\Hv_v(x)$: the forward activation at node $v$.
  \item $\mathcal C_v$: the non-overlapping selected child frontier of $v$ on paths to the target.
  \item $\Rsrc_v(x)$: the local mean-seed VJP source in \cref{eq:local_mean_raw}.
  \item $G_v$: the first-stage local matrix Wiener repair.
  \item $y_v^0$: the first-stage state estimate $G_v\Rsrc_v$.
  \item $C_v(x)$: the pulled-back JVP residual source.
  \item $D_v$: the JVP forward-consistency correction map.
  \item $\SHR_v(\omega)$ and $\SRR_v(\omega)$: first-stage cross- and auto-spectral matrices.
  \item $S_v^{\Delta C}(\omega)$ and $S_v^{CC}(\omega)$: correction cross- and auto-spectral matrices.
  \item $\Omega_v^{\mathrm{st}}$: the stored non-redundant rFFT-bin set at node $v$.
  \item $\Pi_v$: a declared partition of $\Omega_v^{\mathrm{st}}$ used for spectral tying.
  \item $P_{S,Q}$: a shape-preserving terminal channel-coordinate mask projector.
\end{itemize}

The implementation uses a deepest non-overlapping fitting frontier for
$\mathcal C_v$.
If two candidate selected children $u_1$ and $u_2$ satisfy
$u_1\leadsto u_2$ in the selected boundary DAG, then $u_1$ is removed from the
fitting frontier.
This prevents the same downstream computation from contributing through both a
branch-internal child and its later merge or output child.
The unpruned structural edges are retained for ancestor coverage and boundary
placement; only the local source frontier is pruned.

For an edge $v\to u$ in this fitting frontier, the local Jacobian is the partial
derivative of the intervening subgraph with respect to $H_v$ at the actual
forward point.
External inputs to that subgraph that are not descendants of $v$ are held fixed
at their forward values.
All directed paths from $v$ to $u$ remain active, so residual branches and
multi-use tensors contribute through native autograd semantics.

The scalar $C_u$ denotes the channel count at node $u$.
The calligraphic set $\mathcal C_v$ denotes child boundaries, whereas $C_v$
denotes the JVP-FC residual source.

The batch dimension is always dimension zero.
All reported models are frozen in evaluation mode and do not couple examples
across this dimension, so each calibration example contributes one channel vector per stored
frequency to the empirical statistics.
The channel axis is the feature axis over which the full matrix Wiener map is
applied.
For CNN features this is usually dimension one, while for token streams such as
$(B,N,D)$ it is the hidden dimension $D$.
The implementation may infer this axis from tensor shape and module type, and
the user may override it explicitly.
All remaining non-batch, non-channel axes are coordinate axes.
If there are no coordinate axes, then $k=0$ and $\calF_0$ is the identity
transform.
For global-pooled nodes, singleton coordinate axes that only record the pooled
spatial extent are trivial and may be collapsed before applying this rule.
Thus $(B,C,1,1)$ is algebraically equivalent to the coordinate-free vector
$(B,C)$.
For notation, tensors may be written as $(B,C,d_1,\ldots,d_k)$ after this axis
identification; this does not require the original model tensor to store the
channel axis in dimension one.

This convention is intentionally axis-based rather than module-name-based.
The same estimator applies to convolutional features, token features, vector
nodes, and attention tensors once architecture-aware boundary and axis metadata
are specified.
After the coordinate transform, each node also declares a stored-frequency
partition $\Pi_v$.  Frequencies in the same group share one complex
full-channel gain matrix.  This parameter tying remains frequency diagonal and
does not imply that the underlying network is translation equivariant or that a
causal support constraint has been imposed.

\subsection{Mean-Seed Local Source}
\label{app:mean_seed_source}

For a child boundary $u$ with channel count $C_u$, the seed is
\[
  s_u=\frac{\Hv_u}{C_u}.
\]
This is the differential of the channel-averaged child energy
\[
  \Psi_u(\Hv_u)=\frac{1}{2C_u}\|\Hv_u\|_F^2 .
\]
For a single child, the local source is $J_{u\leftarrow v}^*s_u$.
For multiple selected children, the source is the native multi-output VJP
\[
  \Rsrc_v
  =
  \sum_{u\in\mathcal C_v}
  J_{u\leftarrow v}^*
  \frac{\Hv_u}{C_u}.
\]
The division by $C_u$ is applied once, at the child seed.
Coordinate positions are not divided out, and there is no second division by
the number of children.
This convention preserves the standard autograd gauge: multi-output
backpropagation sums cotangents contributed by each output.

At online inversion time, let $s_u^{\mathrm{on}}$ denote the seed attached
to child state $y_u$.
For a nonterminal child, the same channel convention is used.
For the terminal child, choose a nonempty channel set
$S\subseteq\{1,\ldots,C_T\}$ and a nonempty coordinate set $Q$.
For a tensor with multiple coordinate axes, elements of $Q$ are coordinate
tuples; for a coordinate-free vector, $Q$ contains the single trivial
coordinate.  Define channel and coordinate masks that preserve the original
tensor shape and their orthogonal product projector
\[
  P_{S,Q}
  =P_S^{\mathrm{ch}}P_Q^{\mathrm{coord}},
  \qquad
  P_{S,Q}^*=P_{S,Q}=P_{S,Q}^2.
\]
Unselected entries are set to zero rather than removed from the tensor.
With $C_S=|S|$, the terminal state and seed are
\[
  y_T=P_{S,Q}\Hv_T,
  \qquad
  s_T^{\mathrm{on}}=\frac{P_{S,Q}\Hv_T}{C_S},
\]
which is the differential of
\[
  \Psi_T^{(S,Q)}(\Hv_T)
  =
  \frac{1}{2|S|}\left\|P_{S,Q}\Hv_T\right\|_F^2.
\]
The coordinate-set cardinality is therefore not part of the seed divisor.
For the complete online recursion,
\[
  s_u^{\mathrm{on}}
  =
  \begin{cases}
    y_T/C_S, & u=T,\\
    y_u/C_u, & u\neq T.
  \end{cases}
  \qquad
  \Rsrc_v^{\mathrm{on}}
  =
  \sum_{u\in\mathcal C_v}
  J_{u\leftarrow v}^*s_u^{\mathrm{on}}.
\]
The full target has $P_{S,Q}=I$ and $C_S=C_T$.
Thus channel normalisation is always part of the child seed, never part of the
fitted Wiener map.

\subsection{Full Matrix Wiener Derivation}
\label{app:wiener_derivation}

The repair family in \cref{eq:local_operator} is frequency diagonal along
coordinate axes.
For a fixed node $v$, let $\Hv'_v=\calF_k\Hv_v$ and
$\Rsrc'_v=\calF_k\Rsrc_v$.
We take $\calF_k$ to be the full unitary transform in the derivation.
Parseval's identity gives
\begin{equation}
  \E\|\Gv_v(\Rsrc_v)-\Hv_v\|^2
  =
  \sum_{\omega}
  \E\left\|
    \Gv_v(\omega)\Rsrc'_v(\omega)-\Hv'_v(\omega)
  \right\|^2 .
  \label{eq:app_parseval}
\end{equation}
Under the singleton partition, the unknown matrix at frequency $\omega$ appears
only in the $\omega$ term, so the objective decomposes across coordinate
frequencies.

For one frequency, write $R=\Rsrc'_v(\omega)\in\CC^C$,
$H=\Hv'_v(\omega)\in\CC^C$, and $G=\Gv_v(\omega)$, with all vectors treated as
columns.
The frequency-domain repair uses ordinary complex matrix multiplication $GR$;
it is linear, not conjugate-linear, in $R$.
The ridge objective is
\begin{equation}
  J(G)=\E\|GR-H\|_2^2+\lambda\|G\|_F^2 .
\end{equation}
The problem decomposes over output channels.
Let $w_i$ be the $i$-th row of $G$ and $H_i$ the $i$-th component of $H$.
For that row,
\[
  J_i(w_i)
  =
  \E |w_i R-H_i|^2+\lambda \|w_i\|_2^2 .
\]
The complex least-squares stationarity condition is
\[
  w_i\left(\E[RR^*]+\lambda I\right)=\E[H_iR^*].
\]
It follows by treating $w_i$ and its complex conjugate as independent variables
and setting the derivative with respect to the conjugate row to zero.
Stacking the $C$ row equations yields
\[
  G\left(\E[RR^*]+\lambda I\right)=\E[HR^*].
\]
With
\[
  \SRR_v(\omega)=\E[RR^*],
  \qquad
  \SHR_v(\omega)=\E[HR^*],
\]
we obtain the singleton-group case of the closed form in
\cref{eq:matrix_wiener}.
Since $\SRR_v(\omega)$ is Hermitian positive semidefinite,
$\SRR_v(\omega)+\lambda I$ is strictly positive definite for $\lambda>0$, so
the solution is unique.
If $\lambda=0$, uniqueness requires $\SRR_v(\omega)$ to be nonsingular.
In finite calibration, expectations are replaced by empirical averages.
This Parseval derivation directly establishes the singleton, per-frequency
case.  When parameters are tied across stored rFFT bins, the precise objective
is the declared stored-spectrum objective below.

For real tensors, the Fourier spectrum is conjugate symmetric.
Implementations may therefore use a real FFT and store only the non-redundant
spectrum.
This does not change the represented real operator family or the per-frequency
normal equations, provided the forward and inverse transforms use a consistent
convention and the inverse restores the corresponding real signal.
The relative-ridge average below is defined over the stored spectrum, so results
must report and retain that storage convention.
The evaluated implementation uses orthonormal real transforms
(\texttt{norm=ortho}).
It converts channel-first tensors to float32 and stores spectral statistics and
fitted gains in complex64.
A nonunitary FFT normalisation rescales both stored second moments; the
effective relative ridge then rescales by the same factor.

\subsection{Stored-Frequency Partitions and Grouped Closed Forms}
\label{app:stored_frequency_groups}

Let $\Omega_v^{\mathrm{st}}$ be the non-redundant frequency-bin set stored by
the real-transform implementation, and let $\Pi_v$ partition this set into
nonempty groups.  Frequencies in group $g\in\Pi_v$ share one matrix
$G_{v,g}$, so $G_v(\omega)=G_{v,g}$ for every $\omega\in g$.
To state the constrained problem independently of a particular storage
weighting, assign a positive weight $a_\omega$ to every stored bin and define
$a_g=\sum_{\omega\in g}a_\omega$.  The group objective is
\begin{equation}
  \mathcal L_{v,g}(A)
  =
  \frac{1}{a_g}
  \sum_{\omega\in g}a_\omega
  \E\left\|
    A\Rsrc'_v(\omega)-\Hv'_v(\omega)
  \right\|_2^2
  +\lambda\|A\|_F^2.
  \label{eq:app_group_objective}
\end{equation}
Define the weighted pooled moments
\begin{equation}
  \overline{\SHR}_{v,g}
  =
  \frac{1}{a_g}\sum_{\omega\in g}a_\omega\SHR_v(\omega),
  \qquad
  \overline{\SRR}_{v,g}
  =
  \frac{1}{a_g}\sum_{\omega\in g}a_\omega\SRR_v(\omega).
  \label{eq:app_group_moments}
\end{equation}
The row-wise stationarity argument above then gives the exact
equality-constrained closed form
\begin{equation}
  G_{v,g}^\star
  =
  \overline{\SHR}_{v,g}
  \left(\overline{\SRR}_{v,g}+\lambda I\right)^{-1},
  \qquad
  G_v^\star(\omega)=G_{v,g}^\star\quad(\omega\in g).
  \label{eq:app_group_wiener}
\end{equation}
The reported implementation uses $a_\omega=1$ for every stored
non-redundant rFFT bin.  Its grouped solve is therefore exact for the equally
weighted stored-spectrum objective in \cref{eq:wiener_objective}.  It does not
claim equivalence to a coordinate-domain MSE in which full-spectrum Parseval
multiplicities weight ordinary positive frequencies twice and DC and Nyquist
frequencies once.  Changing to those multiplicity weights would define a
different estimator and is not used for any reported map.

The singleton partition $\Pi_v=\{\{\omega\}:\omega\in
\Omega_v^{\mathrm{st}}\}$ recovers the per-frequency solve.  The all-shared
partition $\Pi_v=\{\Omega_v^{\mathrm{st}}\}$ fits one complex full-channel
gain and broadcasts it to every stored bin under the declared rFFT/irFFT
convention.  Spectral tying removes bin-specific gain variation; it does not
remove frequency content already present in the source-local VJP or JVP
observables and does not impose causal support.

The same partition and weights are used for JVP-FC.  With correction moments
$S_v^{\Delta C}(\omega)$ and $S_v^{CC}(\omega)$, their weighted group averages
replace $\overline{\SHR}_{v,g}$ and $\overline{\SRR}_{v,g}$ in
\cref{eq:app_group_wiener}, yielding $D_{v,g}^\star$ in
\cref{eq:jvp_fc_wiener}.

\subsection{Ridge and Zero-Intercept Solve}
\label{app:source_ridge}

Both $G_v$ and $D_v$ are solved as zero-intercept maps.
The zero-frequency coefficient is treated like every other coordinate
frequency: it has a full channel matrix and no additive term.
The ridge is implemented as a relative ridge with respect to the source
auto-spectral second moment of the current solve:
\[
  \lambda_{\mathrm{eff}}
  =
  \rho\,
  \max\!\left\{
    \mathrm{mean}_{\omega,c}S^{XX}_{v,cc}(\omega),
    \epsilon_{\mathrm{ridge}}
  \right\},
  \qquad
  \rho=0.01,
  \quad
  \epsilon_{\mathrm{ridge}}=10^{-30},
\]
where $X=\Rsrc$ for the first-stage map $G_v$ and $X=C$ for the correction map
$D_v$.
This keeps the same nominal relative coefficient $\rho$ meaningful across
nodes with different source scales.
In the closed-form solve, $\lambda_{\mathrm{eff}}$ is the value added to the
diagonal of the current source second moment.
Because $S^{XX}_v(\omega)$ is Hermitian positive semidefinite, its diagonal
entries are real and nonnegative; implementations use the real diagonal when
computing the relative ridge.
The average is taken over the same stored frequency convention used by the
solve, for example over the non-redundant spectrum when a real FFT
implementation is used.
The numerical floor guarantees a positive ridge for a degenerate zero source
when $\rho>0$; it has no practical effect at ordinary source scales.
With $N$ calibration examples, an empirical source second moment has rank at
most $\min(C,N)$ at each singleton frequency.  For a group $g$, the pooled
moment contains $N|g|$ spectral coefficient vectors and has algebraic rank at
most $\min(C,N|g|)$, although these coefficients arise from only $N$
independent calibration inputs.  A positive ridge guarantees an invertible
solve even when the empirical moment is rank deficient.

\subsection{Local Calibration Algorithm for \texorpdfstring{$G_v$}{Gv}}
\label{app:local_calibration_algorithm}

The first-stage map $G_v$ is calibrated from local source-target pairs
$(\Rsrc_v,\Hv_v)$.
It does not use repaired downstream maps as part of its source construction.
Given frozen model $f$, target $T$, selected boundary nodes $\calV_b$, and
calibration set $\calD_{\mathrm{cal}}$:
\begin{enumerate}
  \item For each selected node $v\neq T$ on a path to $T$:
  \begin{enumerate}
    \item find the deepest non-overlapping child frontier $\mathcal C_v$ on paths to $T$;
    \item initialise spectral accumulators $\SHR_v(\omega)$ and $\SRR_v(\omega)$;
    \item for each calibration batch:
    \begin{enumerate}
      \item run the forward pass and cache $\Hv_v$ and $\{\Hv_u:u\in\mathcal C_v\}$;
      \item form child seeds $\Hv_u/C_u$;
      \item compute $\Rsrc_v=\sum_{u\in\mathcal C_v}J_{u\leftarrow v}^*(\Hv_u/C_u)$ by autograd;
      \item transform $\Hv_v$ and $\Rsrc_v$ over coordinate axes;
      \item accumulate $\Hv'_v(\omega)\Rsrc'_v(\omega)^*$ and $\Rsrc'_v(\omega)\Rsrc'_v(\omega)^*$.
    \end{enumerate}
    \item average the per-bin accumulators over calibration inputs;
    \item pool the moments within each declared group $g\in\Pi_v$ using the
    stored-bin weights $a_\omega$ and solve \cref{eq:app_group_wiener};
    \item broadcast $G_{v,g}$ to the stored bins in $g$ for application under
    the declared inverse-transform convention.
  \end{enumerate}
\end{enumerate}
Since each $G_v$ uses only local child activations and local VJPs, the $G_v$
fits are independent once the boundary DAG is fixed.
The implementation may still traverse nodes in a particular order for memory
management, but the estimator itself is local.

\subsection{Calibration Algorithm for JVP-FC Correction}
\label{app:jvp_fc_calibration_algorithm}

After all first-stage maps $G_v$ are available, the correction map $D_v$ is
calibrated from residual pairs.
Each first-stage source already contains the local Jacobian of its calibration
sample, but the fitted $G_v$ is shared across the calibration distribution.
For fixed or weakly sample-dependent local operators, this first projection can
already account for most of the local correction.
JVP-FC is principally motivated by operators whose differential changes more
strongly with the source operating point, including normalisation, attention,
and data-dependent gating or routing.
It remains a general residual regression and can correct any first-stage error
that is statistically observable from its correction source.
For each selected node $v\neq T$:
\begin{enumerate}
  \item For each calibration batch, compute the local source $\Rsrc_v$ exactly as in the $G_v$ calibration.
  \item Apply the first-stage map $y_v^0=G_v\Rsrc_v$.
  In the following JVP, $y_v^0$ is supplied as a tangent vector at the cached
  forward point rather than as a replacement activation.
  \item For every child $u\in\mathcal C_v$, compute the local JVP
  \[
    \widehat \Hv_{u\mid v}=J_{u\leftarrow v}y_v^0 .
  \]
  \item Form child residuals $e_{u\mid v}=\Hv_u-\widehat \Hv_{u\mid v}$.
  This residual contains both the zero-intercept base term
  $\Hv_u-J_{u\leftarrow v}\Hv_v$ and the pushed-forward first-stage error
  $J_{u\leftarrow v}(\Hv_v-y_v^0)$; it is not assumed to vanish at
  $y_v^0=\Hv_v$.
  The base term vanishes for a positively one-homogeneous local map but is
  generally nonzero in the presence of offsets or non-homogeneous nonlinearities.
  \item Pull the residuals back:
  \[
    C_v=\sum_{u\in\mathcal C_v}J_{u\leftarrow v}^*e_{u\mid v}.
  \]
  This is the negative gradient at $y_v^0$ of
  $\frac12\sum_{u\in\mathcal C_v}\|\Hv_u-J_{u\leftarrow v}y\|_F^2$.
  The source therefore uses the Euclidean child-state residual gauge and is not
  divided by $C_u$.
  \item Accumulate $S_v^{\Delta C}(\omega)$ and $S_v^{CC}(\omega)$, pool them
  under the same partition $\Pi_v$ and stored-bin weights as $G_v$, and solve
  $D_{v,g}$ by \cref{eq:jvp_fc_wiener}.
\end{enumerate}
Thus $D_v$ uses the same axis rule and Fourier decomposition as $G_v$.
For $\Delta_v=\Hv_v-y_v^0$, its stored group matrices are
\[
  D_{v,g}^\star
  =
  \overline S_{v,g}^{\Delta C}
  \left(
    \overline S_{v,g}^{CC}+\lambda_D I
  \right)^{-1},
  \qquad
  D_v^\star(\omega)=D_{v,g}^\star\quad(\omega\in g).
\]
This is a second closed-form regression, not query-specific iterative preimage
optimisation.
The JVP and VJP are matrix-free applications of the frozen local subgraph
differential at the current forward point.

\subsection{Online Feature-Inversion Algorithm}
\label{app:online_inversion_algorithm}

At inversion time, the computation proceeds from the target toward the input.
Choose nonempty terminal channel and coordinate sets $(S,Q)$ and let
$C_S=|S|$.
Initialise the terminal state and seed as
\[
  y_T=P_{S,Q}\Hv_T,
  \qquad
  s_T^{\mathrm{on}}=\frac{y_T}{C_S}
  =\frac{P_{S,Q}\Hv_T}{C_S},
\]
where $P_{S,Q}$ is the shape-preserving projector defined in
\cref{eq:terminal_mask_projector}.
The full target has $P_{S,Q}=I$ and $C_S=C_T$.
A single-channel, all-coordinate query has $C_S=1$ and
$Q=Q_{\mathrm{all}}$, where $Q_{\mathrm{all}}$ is the complete coordinate set.
If $Q$ is restricted separately, only the selected coordinates are retained.
Then for each shallower selected node $v$ whose child states have already been
inverted to their boundaries:
\begin{enumerate}
  \item attach the seed
  \[
    s_u^{\mathrm{on}}
    =
    \begin{cases}
      y_T/C_S, & u=T,\\
      y_u/C_u, & u\neq T,
    \end{cases}
  \]
  to each child, and compute the online local source
  \[
    \Rsrc_v^{\mathrm{on}}
    =
    \sum_{u\in\mathcal C_v}
    J_{u\leftarrow v}^*
    s_u^{\mathrm{on}};
  \]
  \item apply the first-stage map $y_v^0=G_v\Rsrc_v^{\mathrm{on}}$;
  \item compute the online child residual source
  \[
    C_v^{\mathrm{on}}
    =
    \sum_{u\in\mathcal C_v}
    J_{u\leftarrow v}^*
    \left(y_u-J_{u\leftarrow v}y_v^0\right);
  \]
  \item return $y_v=y_v^0+D_vC_v^{\mathrm{on}}$.
\end{enumerate}
The input-node state is the reported input-domain feature inverse.
Only the terminal child uses $C_S$ in its seed.
Every inverted nonterminal child uses its physical channel count $C_u$.
The residual source uses the state $y_u$ itself, without seed normalisation.
The calibrated $G_v$ and $D_v$ maps are unchanged when $(S,Q)$ changes.

\subsection{DAG Topology: Branches and Multi-Use Nodes}
\label{app:dag_topology}

The method relies on autograd's VJP semantics for graph topology.
No special mathematical rule is introduced for residual connections,
concatenations, or branching.

\paragraph{Residual addition.}
For $z=a+b$, the VJP distributes the cotangent of $z$ to both $a$ and $b$.
Each branch receives its own cotangent and is repaired at its own tensor nodes.

\paragraph{Concatenation.}
For $z=\mathrm{concat}(a,b)$, the VJP slices the cotangent of $z$ back into the
coordinates or channels corresponding to $a$ and $b$.
Repairs are then applied on the branch tensors.

\paragraph{One-to-many use.}
If a tensor feeds multiple downstream operations, autograd sums the cotangents
contributed by each downstream user.
The node repair acts on this accumulated incoming cotangent.
If the capture granularity exposes separate edge tensors, the same local repair
rule can be applied at those edge nodes instead.

\subsection{Boundary Modes}
\label{app:boundary_modes}

The theoretical maximal form is Full-DAG: every captured differentiable floating
tensor node with a specified channel axis has a well-defined local matrix
Wiener repair.
Scalar nodes, non-floating tensors, and bookkeeping operations without a
meaningful channel axis are traversed by autograd but are not instantiated as
repair nodes.
The implementation exposes three modes:
\begin{itemize}
  \item \textbf{Full-DAG}: instantiate every captured differentiable floating tensor node with a specified channel axis.
  \item \textbf{Coarse/stage block}: instantiate the input, stem or patch embedding, and architecture block or stage edges; this is the default efficient cross-architecture mode.
  \item \textbf{Custom list}: instantiate user-specified nodes after full DAG capture.
\end{itemize}
Skipping nodes is an implementation choice, not a change in the local estimator.
The selected nodes are repair boundaries, not a replacement for the full
forward graph: every intervening computation remains inside the corresponding
frozen autograd subgraph.  The formulation requires matrix-free VJP and JVP
support for these subgraphs.  The current JVP implementation uses a
VJP-of-probes identity and therefore requires double-backward support; fused
attention kernels are replaced by a mathematically equivalent kernel with the
required derivative support when necessary.

\subsection{Architecture-Specific Instances}
\label{app:architecture_instances}

\paragraph{CNNs.}
Feature maps have shape $(B,C,H,W)$.
The channel axis is $C$ and the coordinate axes are $(H,W)$, giving a 2D
Fourier transform and one $C\times C$ matrix per 2D frequency.

\paragraph{DenseNet and ConvNeXt.}
The same tensor rule applies.
Coarse/stage block boundaries provide a compact cross-model instantiation,
while Full-DAG can be used when memory permits.

\paragraph{ViT token streams.}
For tokens with shape $(B,N,D)$, the hidden dimension $D$ is the channel axis
and token position $N$ is the coordinate axis.
The repair uses a 1D Fourier transform over tokens and a full $D\times D$
matrix per token frequency.
When a class token is present, it remains at its stored sequence position and is
included in this transform unless an experiment explicitly specifies a
different token convention.

\paragraph{Causal language token streams.}
For a language-token tensor shaped $(B,L,D)$, the hidden width $D$ is the
channel axis and token position $L$ is the coordinate axis.  Learned positional
embeddings and causal attention remain part of the frozen forward subgraphs and
their source-local VJP/JVP operators.  A stored-frequency partition controls
only how the repair gains are tied; it does not replace or certify those causal
dependencies.

\paragraph{Attention tensors.}
For attention logits or weights shaped $(B,h,N,N)$, the head dimension $h$ is
treated as channel and query-key axes as coordinates.
The repair acts on cotangents or reverse-state estimates rather than a
probability matrix that is reinserted into the forward model, so row-stochastic
softmax constraints are not explicitly enforced.  The estimator remains
algebraically well defined, but both this relaxation and the query-key
frequency-diagonal approximation remain empirical questions.

\paragraph{Vector, GAP, and MLP nodes.}
For $(B,D)$ tensors, $k=0$, so there is no Fourier transform and the repair is a
single $D\times D$ matrix.
Global-pooled tensors stored as $(B,D,1,1)$ are treated equivalently because the
singleton-coordinate Fourier transform is the identity.

\subsection{Reported GPT-2 Small Instantiation}
\label{app:gpt2_method_instance}

The reported language experiment instantiates the same estimator on pretrained
GPT-2 small.  A length-$L=16$ sequence enters the inversion API as continuous
input embeddings with shape $(B,768,16)$; inside GPT-2, token-stream boundaries
have shape $(B,16,768)$.  The target is the complete layer-6 attention-output
sequence before residual addition, rather than the layer-6 residual stream.
Both $G_v$ and $D_v$ are calibrated on these complete sequence tensors before
any terminal-position mask is applied.

The selected repair boundaries are the input embeddings, the outputs of the
second and fourth GPT-2 blocks, and the layer-6 attention output.  These four
boundaries define three fitted nonterminal nodes, each with one $G_v$ map and
one $D_v$ map.
They do not omit the intervening model computation: blocks and operations
between consecutive repair boundaries remain inside the actual frozen autograd
subgraphs used to evaluate their local VJPs and JVPs.

At every fitted boundary, the channel matrices are full and zero-intercept.
The length-16 real transform exposes nine stored rFFT bins, and the all-shared
partition ties one complex full-channel gain matrix across all nine bins for
each map.  The reported implementation gives every stored bin equal weight in
the pooled objective.  The relative ridge coefficient is $\rho=0.01$ for both
$G_v$ and $D_v$; the diagonal shift applied at each solve is its
source-moment-dependent $\lambda_{\mathrm{eff}}$, not an absolute ridge of
$0.01$.

Each formal map family is fitted from 4,096 fixed-seed WikiText-2 windows, and
the reproducibility analysis uses independently sampled calibration sets with
seeds 123 and 321.  At deployment, the reported token query uses
$S=\{1,\ldots,768\}$ and $Q=\{L-1\}$, yielding
\[
  y_T=P_{S,\{L-1\}}\Hv_T,
  \qquad
  s_T^{\mathrm{on}}
  =\frac{P_{S,\{L-1\}}\Hv_T}{768}.
\]
The same full-sequence maps are reused without per-query optimisation or
refitting, and the reported inverse remains in the continuous input-embedding
domain without a learned text decoder or autoregressive decoding.  The general
projector permits other coordinate sets algebraically.  The formal semantic and
syntactic claims use terminal-token queries; the supplementary joint
full-coordinate channel-subset control does not validate isolated
nonterminal-position queries.

\subsection{Restricted Operator Families}
\label{app:degenerate_forms}

The full-channel singleton per-frequency partition is the final vision
configuration.  The visual ablations impose additional restrictions in both
$G_v$ and $D_v$:
\begin{itemize}
  \item \textbf{Diagonal-channel}: retain a separate map at each frequency but force every channel matrix to be diagonal.
  \item \textbf{Shared-frequency}: retain full channel matrices but tie one
  gain across stored coordinate-frequency bins.  This removes
  frequency-specific gain variation from the repair maps; it does not remove
  frequency content already present in the source-local VJP and JVP
  observables.
  \item \textbf{Gamma-only}: force diagonal channel gains and share those gains across frequencies.
  \item \textbf{Diagonal plus low-rank}: use a diagonal channel map plus a low-rank channel correction at each frequency.
\end{itemize}
Diagonal-channel removes cross-channel second-order coupling.  Shared-frequency
is a capacity ablation relative to the vision singleton partition, while the
same equality-constrained closed-form family is the declared GPT-2
instantiation on a different coordinate domain; it is not treated as a GPT-2
ablation.
The matched quantitative and visual comparisons are reported in
\cref{tab:mechanism,tab:supp_operator_family,fig:mechanism}.
Those experiments evaluate not only global input-instance alignment but also
the spatial and spectral structures retained by each restricted family.

\subsection{Complexity and Memory}
\label{app:complexity}

Let $F$ be the number of stored coordinate frequencies.
Let $K_v=|\Pi_v|$ be the number of distinct stored-frequency groups at node
$v$.  For a node with $C$ channels, storing the unpooled full spectral
statistics requires $O(FC^2)$ memory.  Accumulating either source-target
statistic costs $O(NFC^2)$ arithmetic for $N$ calibration examples, excluding
VJP/JVP and FFT costs.  Pooling the stored moments costs $O(FC^2)$, and the
dense grouped solves cost $O(K_vC^3)$ in the worst case.

Each fitted map contains $K_v$ distinct full matrices, or $K_vC^2$ distinct
complex gain coefficients.  The two stages together contain
$2K_vC^2$ distinct complex coefficients per fitted boundary.  These are not
real-parameter counts.  The current implementation may broadcast each group
gain to its member bins for application, in which case physical map storage
remains $FC^2$ complex coefficients per map.  Complex64 storage uses eight
bytes per stored complex coefficient.

Applying one fitted spectral map costs $O(BFC^2)$ arithmetic, in addition to
coordinate-transform cost $O(BCP\log P)$ for $P$ coordinate positions.
At each frequency, each calibration example contributes one $C$-dimensional
Fourier coefficient.  A singleton group therefore has $N$ coefficient vectors.
A group $g$ pools $N|g|$ coefficient vectors but still contains only $N$
independent input sequences; its empirical auto-spectrum has algebraic rank at
most $\min(C,N|g|)$.
Vector nodes have $F=1$ and are therefore comparatively cheap.
ViT token nodes have $C=D$ and $F$ proportional to the token sequence frequency
count, so calibration size and statistic accumulation can be substantial.  In
the reported GPT-2 instance, $F=9$ and $K_v=1$ at every fitted boundary.

Engineering controls include frequency chunking, solve chunking, CPU/GPU
placement of maps, cached calibration tensors, and cotangent caching.
These optimisations do not change the estimator as long as they preserve the
local source-target pairs defined above.
JVP-FC additionally requires an autodiff path with second-order derivative
support because the matrix-free JVP is evaluated through a VJP-of-probes
identity.
For fused attention implementations, a mathematically equivalent kernel with
double-backward support must be enabled when the default kernel lacks it.

\subsection{Reproducibility Details}
\label{app:reproducibility}

Each experiment should report:
\begin{itemize}
  \item model architecture, frozen weights, and evaluation mode,
  \item data split for calibration and evaluation,
  \item sampling scheme and random seeds for both splits,
  \item input preprocessing, including tokenisation and sequence construction
  when applicable,
  \item calibration size and batch size,
  \item target node, nonempty channel and coordinate sets $(S,Q)$, and the
  shape-preserving terminal mask definition,
  \item boundary mode and the exact selected repair boundaries,
  \item non-overlapping child-frontier construction,
  \item calibration seed $s_u=\Hv_u/C_u$, nonterminal online seed
  $s_u^{\mathrm{on}}=y_u/C_u$, and terminal subset seed
  $s_T^{\mathrm{on}}=P_{S,Q}\Hv_T/|S|$,
  \item channel-axis convention or override,
  \item source-normalisation setting and zero-intercept linear convention,
  \item channel solve mode, stored-frequency partition $\Pi_v$, and group
  weights $a_\omega$,
  \item FFT normalisation, stored-spectrum convention, numerical precision,
  and whether group gains are broadcast for storage,
  \item first-stage and correction ridge values and whether relative ridge is
  used,
  \item nominal relative ridge coefficient $\rho$ and effective ridge definition,
  \item JVP-FC child-target mode, residual scale, and correction weight,
  \item map device and chunking settings,
  \item autodiff or attention-kernel settings needed for JVP double backward,
  \item post-calibration online runtime, hardware, software version, and memory constraints.
\end{itemize}
These conventions define the deployed estimator used by the supplementary
theory and by every headline experiment.

\section{Supplementary Theory: From the Statistical Ideal to DAG Stability}
\label{app:theory_details}

This appendix follows the derivation order in \cref{sec:theory}.
As in the main text, the estimated objects are upstream tensor states along the
source-local inverse branch, not generic reconstructions of the complete source
input from a compressed code.
It first proves the conditional-mean optimum, then connects local Wiener
inversion to a covariance-induced natural adjoint.
It next formalises the mean-seed observable and the two distributional Wiener
projections, before deriving channel--coordinate-conditioned finite-DAG
stability and the remaining operator boundaries.

\subsection{MSE-Optimal Conditional-Mean Inversion}
\label{app:optimal_inversion}

Let $H_v$ and $H_u$ be square-integrable random states on the data
distribution, with $H_v$ upstream of $H_u$.
Define
\[
  m(H_u)=\E[H_v\mid H_u].
\]
For any measurable estimator $\mathcal T(H_u)$, write
\[
  H_v-\mathcal T(H_u)
  =
  \bigl(H_v-m(H_u)\bigr)
  +
  \bigl(m(H_u)-\mathcal T(H_u)\bigr).
\]
Conditional-expectation orthogonality gives
\[
  \E\left[
    \left\langle
      H_v-m(H_u),
      m(H_u)-\mathcal T(H_u)
    \right\rangle
  \right]
  =0.
\]
Therefore
\begin{equation}
  \E\|H_v-\mathcal T(H_u)\|_2^2
  =
  \E\|H_v-m(H_u)\|_2^2
  +
  \E\|m(H_u)-\mathcal T(H_u)\|_2^2.
  \label{eq:app_conditional_mean_decomposition}
\end{equation}
The second term is nonnegative, so the conditional mean is the MSE-optimal
inverse, uniquely up to almost-sure equality.

This statistical ideal uses only the downstream observation.
The deployed method is input-conditioned in a different sense: the input being
explained is available to instantiate the frozen forward point and its local
Jacobians.
It is not included as an extra variable in the conditional mean, because for a
deterministic frozen model the full input already determines every forward
state.
The method does not estimate the full nonlinear conditional law in
\cref{eq:app_conditional_mean_decomposition}.
It uses the forward point to construct a sequence of local observable inverse
problems on the boundary DAG.

\subsection{Local Wiener Inversion and the Natural Adjoint}
\label{app:wiener_natural_adjoint}

Vectorise the local tensors and hold the Jacobian $J$ fixed at one forward point
for this unrestricted reference calculation.
Write $\widetilde H_v$ and $\widetilde H_u$ for centred parent and child
deviations, and assume
\begin{equation}
  \widetilde H_u=J\widetilde H_v+\eta,
  \qquad
  \E[\eta\widetilde H_v^*]=0.
  \label{eq:app_local_linear_model}
\end{equation}
Multiple child states can be stacked, with $J$ formed by vertically stacking
their local Jacobians.
Define the covariances
\[
  \Sigma_v=\E[\widetilde H_v\widetilde H_v^*],
  \qquad
  \Sigma_\eta=\E[\eta\eta^*],
  \qquad
  \Sigma_u=J\Sigma_vJ^*+\Sigma_\eta.
\]
For a linear estimator $A\widetilde H_u$, the normal equation for
$\min_A\E\|A\widetilde H_u-\widetilde H_v\|_2^2$ is
\[
  A\Sigma_u
  =
  \E[\widetilde H_v\widetilde H_u^*]
  =
  \Sigma_vJ^*.
\]
If $\Sigma_u$ is nonsingular, the unique best linear estimator is
\begin{equation}
  A^\star
  =
  \Sigma_vJ^*\Sigma_u^{-1}.
  \label{eq:app_local_wiener_inverse}
\end{equation}
Under joint Gaussianity, this linear estimator equals
$\E[\widetilde H_v\mid\widetilde H_u]$, the zero-mean specialisation of the
conditional-mean inverse.
Without Gaussianity, it remains the best zero-intercept linear estimator.

The same operator arises from weighted geometry when both covariances are
positive definite.
Use the precision metrics
\[
  M_v=\Sigma_v^{-1},
  \qquad
  M_u=\Sigma_u^{-1},
\]
and the inner products $\langle a,b\rangle_M=a^*Mb$.
The Riesz adjoint $J^\sharp$ is determined by
\[
  \langle J\delta_v,s_u\rangle_{M_u}
  =
  \langle\delta_v,J^\sharp s_u\rangle_{M_v}.
\]
Equality for all arguments gives
\begin{equation}
  J^\sharp
  =
  M_v^{-1}J^*M_u
  =
  \Sigma_vJ^*\Sigma_u^{-1}
  =
  A^\star.
  \label{eq:app_wiener_natural_equivalence}
\end{equation}
Thus, after the metric-induced identification of vectors and covectors, local
Wiener inversion and the covariance-induced natural adjoint are the same ideal
operator under the centred linear model.
If a covariance is singular, the LMMSE map can be written with a pseudoinverse
on its support.
A positive-definite metric on the full space instead requires regularisation,
for example replacing $\Sigma$ by $\Sigma+\tau I$ with $\tau>0$.
Centring is used only for this ideal identity.
The deployed estimator retains the uncentred, zero-intercept source-target
objectives defined in the Method and does not fit an additive bias.

This identity motivates the geometry that an inverse correction should supply.
It is not an algebraic characterisation of the deployed $G_v$ or $D_v$.
In a nonlinear network, the local Jacobian and appropriate covariance geometry
vary with the input and boundary.
Explicit deployment would require estimating and applying both metrics online.

\subsection{Variational Interpretation of the Mean-Seed Observable}
\label{app:mean_seed_theory}

For a child activation $H_u\in\R^{B\times C_u\times\cdots}$, define
\begin{equation}
  \Psi_u(H_u)
  =
  \frac{1}{2C_u}\|H_u\|_F^2.
  \label{eq:app_mean_energy}
\end{equation}
Its differential with respect to $H_u$ is $H_u/C_u$.
Therefore
\[
  J_{u\leftarrow v}^*\frac{H_u}{C_u}
\]
is the exact adjoint pullback of a scalar child-state observable.

It is also the arithmetic mean of the channel-energy pullbacks.
Writing $H_{u,c}$ for channel $c$ and
$\Psi_{u,c}=\|H_{u,c}\|_F^2/2$ gives
\[
  \frac{1}{C_u}
  \sum_{c=1}^{C_u}
  J_{u\leftarrow v}^*
  \frac{\partial\Psi_{u,c}}{\partial H_u}
  =
  J_{u\leftarrow v}^*
  \frac{H_u}{C_u}.
\]
This equality uses only VJP linearity.

For a node with multiple selected children, define
\[
  \Psi_{\mathcal C_v}
  =
  \sum_{u\in\mathcal C_v}
  \frac{1}{2C_u}\|H_u\|_F^2.
\]
Its differential with respect to $H_v$ is
\begin{equation}
  R_v
  =
  \sum_{u\in\mathcal C_v}
  J_{u\leftarrow v}^*
  \frac{H_u}{C_u}.
  \label{eq:app_mean_seed_source}
\end{equation}
Thus summing child VJPs is the differential of a well-defined local observable.
Dividing again by $|\mathcal C_v|$ would define a different observable and
change the child gauge.

The channel--coordinate terminal seed has the same variational interpretation.
For a nonempty channel set $S$ and nonempty coordinate set $Q$, let $P_{S,Q}$
be the shape-preserving orthogonal projector in
\cref{eq:terminal_mask_projector}, and let $C_S=|S|$.
Then
\begin{equation}
  \Psi_T^{(S,Q)}(H_T)
  =
  \frac{1}{2C_S}\|P_{S,Q}H_T\|_F^2,
  \qquad
  \nabla_{H_T}\Psi_T^{(S,Q)}
  =
  \frac{P_{S,Q}H_T}{C_S}.
  \label{eq:app_subset_mean_seed}
\end{equation}
The coordinate-set cardinality does not enter the divisor.
The full target selects every terminal channel and coordinate, so
$P_{S,Q}=I$ and $C_S=C_T$.

\subsection{Distributional Wiener Projection and the Proxy Bridge}
\label{app:jacobian_induced_stats}
\label{app:proxy_bridge_rigor}

The ideal operator in \cref{eq:app_wiener_natural_equivalence} applies a child
precision before the adjoint and a parent covariance after it.
The deployed source in \cref{eq:app_mean_seed_source} instead uses an observable
Euclidean VJP.
Calibration therefore estimates a ridge-regularised distributional map from
this source to the parent state.

Let $g\in\Pi_v$ be a stored-frequency group, assign each stored bin a positive
weight $a_\omega$, and define $a_g=\sum_{\omega\in g}a_\omega$.
The grouped ridge objective is
\[
  \mathcal L_{v,g}(G)
  =
  \frac{1}{a_g}\sum_{\omega\in g}a_\omega
  \E\left\|G R'_v(\omega)-H'_v(\omega)\right\|_2^2
  +\lambda\|G\|_F^2.
\]
Using the weighted pooled moments in \cref{eq:app_group_moments}, its normal
equation is
\[
  G\left(\overline{\SRR}_{v,g}+\lambda I\right)
  =
  \overline{\SHR}_{v,g}.
\]
For $\lambda>0$, the unique equality-constrained solution is
\begin{equation}
  G_{v,g}^\star
  =
  \overline{\SHR}_{v,g}
  \left(
    \overline{\SRR}_{v,g}+\lambda I
  \right)^{-1},
  \qquad
  G_v^\star(\omega)=G_{v,g}^\star
  \quad(\omega\in g).
  \label{eq:app_distributional_projection}
\end{equation}
The full complex least-squares derivation and Fourier decomposition are given
in \cref{app:wiener_derivation}.
No Gaussian assumption is required for optimality of this ridge objective
within the selected linear class.
The reported estimator sets $a_\omega=1$ for every stored non-redundant rFFT
bin.
The singleton and all-shared partitions are therefore two equality-constrained
instances of the same equally weighted stored-spectrum objective; neither is
claimed to equal a coordinate-domain Parseval objective with different bin
weights.

The statistics in \cref{eq:app_distributional_projection} encode the relation
induced jointly by the input-dependent Jacobian, the data distribution, and the
mean-seed observable.
They do not separately identify the Jacobian or either covariance metric.
The fitted $G_v$ is therefore a closed-form distributional projection surrogate
for the unavailable metric-aware inverse.
During calibration, $R_v(x)$ combines the forward child states with the
source-local Jacobians through its VJP.
During deployment, $R_v^{\mathrm{on}}(x)$ instead uses the supplied repaired
child states, while the fitted projection $G_v$ remains shared across samples.
This distinction is minor for fixed or weakly sample-dependent local operators,
where one distributional projection can capture the dominant correction.
For strongly sample-dependent local operators, the relation between $R_v(x)$
and $H_v(x)$ can retain an operating-point-specific mismatch.

\paragraph{Dependency support.}
By the chain rule, a child seed affects node $v$ only through computational paths
connecting $v$ to that child.
Entries outside the corresponding Jacobian sparsity pattern receive zero
cotangent.
This is a theorem-level dependency statement.

\paragraph{State-structure proxy.}
Dependency support does not imply that nonzero cotangent support equals forward
activation support.
The working assumption is weaker: the coordinate or semantic organisation of
$R_v$ remains predictive of the organisation carried by $H_v$.
This assumption is evaluated through state-inversion diagnostics,
modality-native input-domain structure, and transfer.

\paragraph{Local transfer.}
Different terminal targets change the seed and therefore the observable source.
Using one calibrated map family across compatible target depths and
channel--coordinate masks tests whether the dominant correction is node-local
rather than a stored query-specific reconstruction.

\subsection{JVP-FC as a Second Residual Regression}
\label{app:jvp_fc_theory}

Let $y_v^0=G_vR_v$ be the first-stage estimate.
JVP-FC evaluates this estimate under the local differential of the actual source
operating point and thereby constructs a second, operating-point-conditioned
observable.
This mechanism is especially relevant for normalisation, attention, and
data-dependent gating or routing, although the residual regression itself is
not restricted to those components.
In the JVP below, $y_v^0$ is interpreted as a tangent vector at the fixed
forward point, not as a replacement forward activation.
For each selected child, define the JVP residual
\begin{equation}
  e_{u\mid v}
  =
  H_u-J_{u\leftarrow v}y_v^0.
  \label{eq:app_jvp_residual}
\end{equation}
The exact decomposition is
\begin{equation}
  e_{u\mid v}
  =
  \bigl(H_u-J_{u\leftarrow v}H_v\bigr)
  +
  J_{u\leftarrow v}(H_v-y_v^0).
  \label{eq:app_jvp_residual_decomposition}
\end{equation}
The first term is the offset or non-homogeneous base component missing from the
zero-intercept differential.
The residual therefore need not vanish at the exact parent state.
For a positively one-homogeneous local map, Euler's identity makes this base
term zero at differentiable points.
The second term transports the first-stage state error through the current
sample's local differential and exposes its source-specific child response.

Pulling the residual back gives
\begin{equation}
  C_v
  =
  \sum_{u\in\mathcal C_v}
  J_{u\leftarrow v}^*e_{u\mid v}.
  \label{eq:app_jvp_correction_source}
\end{equation}
Define
\[
  \Phi_v(y)
  =
  \frac12\sum_{u\in\mathcal C_v}
  \|H_u-J_{u\leftarrow v}y\|_F^2.
\]
Direct differentiation gives $C_v=-\nabla\Phi_v(y_v^0)$.
The unnormalised child residual therefore corresponds to a specific Euclidean
child-consistency objective.
Let $\Delta_v=H_v-y_v^0$, $C'_v=\calF_kC_v$, and
$\Delta'_v=\calF_k\Delta_v$.
Using the same partition and stored-bin weights as $G_v$, the second map solves
\begin{equation}
  D_{v,g}^\star
  =
  \arg\min_D
  \frac{1}{a_g}\sum_{\omega\in g}a_\omega
  \E\left\|D C'_v(\omega)-\Delta'_v(\omega)\right\|_2^2
  +\lambda_D\|D\|_F^2.
  \label{eq:app_jvp_projection_objective}
\end{equation}
Let $\overline S_{v,g}^{CC}$ and
$\overline S_{v,g}^{\Delta C}$ denote the correction moments pooled with the
same weights as in \cref{eq:app_group_moments}.
Its normal equation and unique solution for $\lambda_D>0$ are
\[
  D\left(\overline S_{v,g}^{CC}+\lambda_D I\right)
  =\overline S_{v,g}^{\Delta C},
  \qquad
  D_{v,g}^\star
  =\overline S_{v,g}^{\Delta C}
  \left(\overline S_{v,g}^{CC}+\lambda_D I\right)^{-1}.
\]
Thus $D_v$ has the same grouped ridge-regularised Wiener form as $G_v$, with
$C_v$ as source and $\Delta_v$ as target.
This does not make every residual component observable.
Any child-residual component in the null space of the summed pullback is absent
from $C_v$, and the fitted map can estimate only parent-error components that
are statistically predictable from the resulting source.

The online version replaces each true child state by the already inverted
state:
\begin{equation}
  C_v^{\mathrm{on}}
  =
  \sum_{u\in\mathcal C_v}
  J_{u\leftarrow v}^*
  \left(y_u-J_{u\leftarrow v}y_v^0\right).
  \label{eq:app_online_jvp_source}
\end{equation}
This is deployable because $y_u$ is available during target-to-input recursion.
When the downstream state estimate is accurate, the online residual approaches
the calibration residual.
Otherwise it exposes the child state that is actually propagated.

The term $J_{u\leftarrow v}y_v^0$ is a JVP at the original forward point.
It is not a nonlinear evaluation of $f_{u\leftarrow v}$ on $y_v^0$.
The correction is therefore a residual regression in local differential
coordinates, not a claim that the nonlinear subgraph is homogeneous.

\subsection{Channel--Coordinate-Conditioned Finite-DAG Error Propagation and Stability}
\label{app:finite_dag_stability}

The local closed-form objectives do not by themselves control how errors
accumulate during the target-to-input recursion.
We first give an exact recursion relative to an arbitrary
channel--coordinate-conditioned reference family, then obtain the forward-state
theorem as the full-target special case.

Fix an input $x$, a nonempty terminal channel set $S$, and a nonempty terminal
coordinate set $Q$.
Let $C_S=|S|$.
Equip each selected node space with a fixed norm and each block between node
spaces with the corresponding induced operator norm.
For a child node $u$, define the online seed weight
\begin{equation}
  \alpha_u^{(S)}
  =
  \begin{cases}
    1/C_S, & u=T,\\
    1/C_u, & u\neq T.
  \end{cases}
  \label{eq:app_online_seed_weight}
\end{equation}
For a nonterminal boundary node $v$, stack arbitrary child states as
$z_{\mathcal C_v}=(z_u)_{u\in\mathcal C_v}$ and define
\begin{align}
  \mathcal P_v^{(S)}z_{\mathcal C_v}
  &=
  \sum_{u\in\mathcal C_v}
  J_{u\leftarrow v}^*\alpha_u^{(S)}z_u,
  &
  \mathcal Q_vz_{\mathcal C_v}
  &=
  \sum_{u\in\mathcal C_v}
  J_{u\leftarrow v}^*z_u,
  \\
  \mathcal L_v
  &=
  \sum_{u\in\mathcal C_v}
  J_{u\leftarrow v}^*J_{u\leftarrow v}.
  \label{eq:app_child_operators}
\end{align}
At the fixed forward point, the complete update driven by these child states is
\begin{align}
  y_v^0
  &=G_v\mathcal P_v^{(S)}z_{\mathcal C_v},\\
  C_v
  &=\mathcal Q_vz_{\mathcal C_v}
    -\mathcal L_vG_v\mathcal P_v^{(S)}z_{\mathcal C_v},\\
  y_v
  &=\mathcal K_v^{(S)}z_{\mathcal C_v},
  \qquad
  \mathcal K_v^{(S)}
  =
  G_v\mathcal P_v^{(S)}
  +D_v\left(
    \mathcal Q_v-\mathcal L_vG_v\mathcal P_v^{(S)}
  \right).
  \label{eq:app_effective_reverse_operator}
\end{align}
Thus the local update is linear in the supplied child states, although its
Jacobian-dependent coefficients remain input-conditioned.
The displayed operator uses the unit residual and correction weights evaluated
in the paper; any fixed scalar weights can be absorbed into the effective
correction operator in $\mathcal K_v^{(S)}$.
The superscript $S$ records dependence through the terminal normalisation
$C_S=|S|$; at a fixed operating point, the operator does not depend on the
identity of the selected channels or on $Q$.
The channel and coordinate identities enter through the terminal state
$P_{S,Q}H_T$.

Let $\{\bar H_v^{(S,Q)}\}$ be any channel--coordinate-conditioned reference
family satisfying
\begin{equation}
  \bar H_T^{(S,Q)}=P_{S,Q}H_T.
  \label{eq:app_reference_terminal}
\end{equation}
Define the local defect
\begin{equation}
  \epsilon_v^{(S,Q)}
  =
  \mathcal K_v^{(S)}
  \bar H_{\mathcal C_v}^{(S,Q)}
  -\bar H_v^{(S,Q)}.
  \label{eq:app_local_defect}
\end{equation}
For online errors
$\delta_v^{(S,Q)}=y_v-\bar H_v^{(S,Q)}$, linearity gives
\begin{equation}
  \boxed{
  \delta_v^{(S,Q)}
  =
  \epsilon_v^{(S,Q)}
  +
  \mathcal K_v^{(S)}
  \delta_{\mathcal C_v}^{(S,Q)}
  }.
  \label{eq:app_error_recursion}
\end{equation}
The terminal error is zero because the online and reference terminal states are
both $P_{S,Q}H_T$.

For full-target inversion,
$S=\{1,\ldots,C_T\}$ and $Q=\Omega_T^{\mathrm{coord}}$, the complete terminal
coordinate set, so $P_{S,Q}=I$ and $C_S=C_T$.
Choosing $\bar H_v^{(S,Q)}=H_v$ then recovers the forward-state errors and
local defects used in the main text.
For a strict channel or coordinate subset, \cref{eq:app_error_recursion}
remains valid relative to any chosen channel--coordinate-conditioned reference
family.
The theorem controls propagation around that reference but does not define a
unique semantic ground truth for channel- or coordinate-specific inversion.
The reported GPT-2 query selects every hidden channel but only the terminal
position, so it is a strict coordinate query rather than this full-target
special case.

Write $\mathcal K_{vu}^{(S)}$ for the block acting on child $u$ and
$\kappa_{vu}^{(S)}=\|\mathcal K_{vu}^{(S)}\|_{\mathrm{op}}$.
Then
\begin{equation}
  \|\delta_v^{(S,Q)}\|
  \le
  \|\epsilon_v^{(S,Q)}\|
  +
  \sum_{u\in\mathcal C_v}
  \kappa_{vu}^{(S)}\|\delta_u^{(S,Q)}\|.
  \label{eq:app_error_bound}
\end{equation}
Because the graph is acyclic and the terminal error is zero, this inequality
can be unrolled over all paths from $v$ toward the target.
If $w\succeq v$ denotes that $w$ is a selected descendant of $v$ and
$\mathrm{Paths}(v,w)$ is the set of selected-DAG paths from $v$ to $w$, then
\begin{equation}
  \|\delta_v^{(S,Q)}\|
  \le
  \sum_{w\succeq v,\,w\ne T}
  \left(
    \sum_{p\in\mathrm{Paths}(v,w)}
    \prod_{(a,b)\in p}
    \kappa_{ab}^{(S)}
  \right)
  \|\epsilon_w^{(S,Q)}\|,
  \label{eq:app_global_path_bound}
\end{equation}
where the empty path for $w=v$ has product one.

If every local defect is zero, reverse topological induction gives zero error
at every selected node.
If the local reverse gain satisfies
\begin{equation}
  \sup_{v\ne T}
  \sum_{u\in\mathcal C_v}
  \kappa_{vu}^{(S)}
  \le q<1,
  \label{eq:app_contraction_condition}
\end{equation}
then
\begin{equation}
  \max_v\|\delta_v^{(S,Q)}\|
  \le
  \frac{
    \max_{v\ne T}\|\epsilon_v^{(S,Q)}\|
  }{1-q}.
  \label{eq:app_contraction_bound}
\end{equation}
Indeed, if
$M=\max_v\|\delta_v^{(S,Q)}\|$ and
$E=\max_{v\ne T}\|\epsilon_v^{(S,Q)}\|$, then
\cref{eq:app_error_bound} gives $M\le E+qM$.
This condition is sufficient rather than necessary.
When it fails, \cref{eq:app_global_path_bound} still identifies the boundaries
and paths that can amplify local defects.
These are finite-pass stability statements, not iterative convergence claims.

\subsection{Zero Target-Independent Injection}
\label{app:zero_hallucination}

The two stages have the zero-intercept form
\[
  y_v^0=G_vR_v,
  \qquad
  y_v=y_v^0+D_vC_v.
\]
If the selected target state satisfies $P_{S,Q}H_T=0$, its terminal seed is
zero.
Induction through the DAG then gives zero local VJP sources, zero first-stage
states, zero JVP residual sources, and zero inverted states.
The calibrated matrices can encode a calibration-distribution-dependent linear
prior, but they
cannot add a fixed target-independent component.

This is narrower than support non-expansion.
A frequency-diagonal multiplier is convolutional in coordinate space and can
spread energy.
The guarantee is only that every inverted component is induced by the selected
target, the frozen model derivatives, and fixed zero-intercept maps.

\subsection{The Frequency-Diagonal Structural Boundary}
\label{app:frequency_diagonal}

The main structural restriction is coordinate-frequency diagonality.
At each frequency, channels may mix through a full matrix, but different
coordinate frequencies do not mix inside a repair map.
The stored-frequency partition $\Pi_v$ may additionally impose equality
constraints that tie the matrices assigned to selected bins.
For a linear operator on the selected discrete Fourier grid, this is equivalent
to a matrix-valued circular convolution under that grid's boundary convention.
The equivalence describes the repair operator, not the original network.

Without this restriction, a node with $C$ channels and $P$ coordinate positions
would require an unrestricted $(CP)\times(CP)$ linear map.
Frequency diagonality instead gives $F$ channel-matrix blocks, while a partition
with $K_v=|\Pi_v|$ groups reduces the number of distinct fitted matrices to
$K_v$.
This provides the tractability needed for full-channel calibration.

Whether the optimal repair is adequately approximated in this class is an
empirical question.
The evaluated CNN and vision-Transformer boundaries, including the strict
ViT-B/32 patch-coordinate queries, together with the controlled GPT-2 small
terminal-token setting and joint full-coordinate channel-subset control,
support this operator class in the studied configurations.
Internal attention-logit or attention-weight matrices, arbitrary nonterminal
language-token queries, and transfer to other language-model architectures
remain outside the present empirical validation.
The supplementary experiments next test each empirical bridge separately:
observable transfer, residual repair, finite-DAG diagnostics, information flow,
and the practical calibration and deployment boundaries.

\section{Supplementary Experiments}
\label{app:experiments}

This supplement follows the claim--evidence order of the main experiments.
It records the detailed evaluation protocol, expanded quantitative results,
negative controls, deployment audits, and the optional breadth studies.
Visual experiments use image-space metrics to describe alignment with the known
target-generating input, whereas the language experiment uses token-position
energy profiles and modality-specific routing controls.
Neither family is treated as a universal ground-truth correctness measure for a
feature inverse.
The progression is protocol and reuse, structural and distributional scope,
source-path faithfulness, mechanism and theory bridge, inverse meaning,
deployment audits, decision-grounded interpretation, optional visual
extensions, and finally language-Transformer validation.

\subsection{Reproducibility, Metric Definitions, and Final-Result Criteria}
\label{app:experiment_protocol}

\paragraph{Fixed protocol.}
Unless a row is explicitly marked as an ablation or diagnostic, all results use
the final public method: local mean seeds, zero-intercept full-channel maps for
both $G_v$ and $D_v$, repaired online child states, declared
architecture-aware boundaries, no activation or cotangent pre-normalisation,
and relative ridge $\rho=0.01$.
The default vision configuration uses the singleton per-frequency partition;
the reported GPT-2 configuration uses all-shared spectral tying over stored
token-frequency bins.
The calibration and evaluation sets are disjoint.
For the headline ImageNet experiments, 4,096 training images are sampled with
seed 123 and 1,024 validation images are sampled with seed 456.
Dataset-specific Pets experiments use all 3,680 available training images for
calibration.  Other dataset-specific calibrations use 4,096 images.
We use c$N$/e$M$ as shorthand for $N$ calibration images and $M$ evaluation
images.

The final vision runs used Python 3.10.4, PyTorch 2.5.1 with CUDA 12.1,
TorchVision 0.20.1, and timm 1.0.25.  Runtime measurements used one NVIDIA
GeForce RTX 3090 GPU with 24~GiB of memory on a host with two Intel Xeon Gold
6426Y processors.  Each run was bound to one GPU.  The fitted-map placement
followed the reported deployment configuration: ResNet18 maps remained on the
GPU, while the larger headline map families were applied from CPU memory.

Each vision run records the model and weight identifier, dataset root, preprocessing
transform, calibration and evaluation indices, target node, fitted boundary
DAG, tensor shape, channel axis, coordinate axes, frequency shape, ridge,
source-code hashes, map metadata, and per-image metrics.
The formal component validator additionally requires the same public code path,
no component-specific inverse, finite outputs, and a complete target-axis
record.  Nondeployable oracle conditions are labelled diagnostic-only and are
never used as headline method rows.

\paragraph{Tensor and Fourier convention.}
A floating tensor is represented as batch, channel, and zero or more coordinate
axes as defined in \cref{app:notation_axes}.  Real-valued coordinate axes use an
orthonormal real Fourier transform.  The calibrated map is full over channels
and diagonal over coordinate frequency, while the declared partition $\Pi_v$
may tie selected stored-frequency matrices.
Vector states therefore have one frequency bin, while token streams use the
token coordinate as their Fourier axis.
The partition changes operator capacity but not the source-local VJP/JVP
observables or the boundary DAG.
The forward model fixes the target node and boundary topology before
calibration; no inverse component is manually inserted at evaluation time.

\paragraph{Input-instance alignment.}
For flattened prediction $\widehat x$ and known target-generating input $x$,
pixel cosine is
\begin{equation}
  \operatorname{cos}_{\mathrm{pix}}(\widehat x,x)
  =
  \frac{\langle \widehat x,x\rangle}
       {\|\widehat x\|_2\|x\|_2}.
\end{equation}
The same cosine is computed after separating luminance low-pass,
luminance high-pass, and chroma components.  SSIM and LPIPS provide
complementary structural and perceptual descriptions
\citep{wang2004image,zhang2018unreasonable}.  Relative $\ell_2$ is
$\|\widehat x-x\|_2/\|x\|_2$.  These metrics quantify input-instance alignment
for a known source reference.  A low value for a channel or channel-subset
target does not by itself invalidate the inverse, because the selected target
is $P_{S,Q}H_T$ rather than the complete input.

\paragraph{Representation and mechanism diagnostics.}
Re-encoding cosine compares the representation of the inverse with the target
representation and is reported only as representation consistency.  Child
NMSE measures the local JVP prediction error before and after $D_v$.
Intermediate-state relative $\ell_2$ measures the corresponding state-space
error.  Runtime is reported for post-calibration online application.
Calibration complexity is described separately by calibration-set size,
fitted-node count, map storage, and peak memory rather than by
cross-architecture fit wall time.  Map storage includes both $G_v$ and $D_v$;
peak memory is measured separately during calibration and evaluation.

\paragraph{Uncertainty and visual selection.}
Headline ImageNet means use all 1,024 fixed evaluation images
\citep{deng2009imagenet,russakovsky2015imagenet}.
Calibration-estimation variation is assessed with independently sampled
calibration subsets and is kept separate from evaluation-image uncertainty.
Operating-point curves use deterministic 10,000-resample bootstrap intervals.
The decision-conditioned study aggregates random controls within each image
before deterministic 20,000-resample image-cluster bootstrap intervals, which
prevents random-control repeats from being treated as independent images.
The language benchmarks average the two calibration-map outputs before 10,000
percentile cluster-bootstrap resamples over polysemous words or lexical
templates; decision-intervention intervals resample held-out contexts.
Visual sample indices, ordering, targets, and display transformations are fixed
before figure assembly.  Inverse tensors are scaled only for display after all
quantitative measurements.

\subsection{Large-Scale Atlas, No-Refit Reuse, Target Depth, and Optimisation Details}
\label{app:large_scale_atlas}

\paragraph{One fit, many queries.}
The canonical ResNet50 atlas uses one c4096 fitted map family.
It serves 1,024 evaluation images, four full-state depths, four terminal
channels, and one eight-channel terminal subset without refitting any map or
changing the ridge.  The four full-state target keys are \texttt{full\_n1},
\texttt{full\_n5}, \texttt{full\_n9}, and \texttt{full\_n17}.
Their pixel cosines are 0.9448, 0.9282, 0.9339, and 0.9333, respectively.
The four individual terminal channels have pixel cosines 0.8254, 0.5979,
0.9243, and 0.9199; the fixed eight-channel subset reaches 0.8932.
The channel results demonstrate public-API reuse and output diversity.  They
are not evaluated as attempts to reproduce the complete input from one channel.

\paragraph{Canonical iterative preimage optimisation.}
The external numerical baseline starts from noise and minimises the terminal
feature discrepancy for 2,000 Adam steps per image \citep{kingma2015adam}.
The learning rate is 0.05,
the total-variation and $\ell_2$ weights are 0.1 and 0.001, and random spatial
jitter is bounded by four pixels \citep{rudin1992nonlinear}.  A fixed pilot
sweep from 1,000 to 3,000
steps selected this setting because additional steps produced only a small
alignment change while increasing cost.  The target layer, preprocessing,
evaluation images, and feature objective are matched to the closed-form run.
Three predeclared initialisation seeds are used.  No learned decoder or saliency
method is included because the intended numerical comparison is between
canonical feature-preimage search and the proposed source-grounded closed-form
regime.

Across the matched 16-image set, iterative optimisation obtains mean target
re-encoding cosine 0.9575 and feature NMSE 0.0794, but pixel cosine is 0.0351.
Mean pairwise output cosine across the three initialisations is 0.1010,
demonstrating that the converged preimages remain seed dependent.
After calibration, the final closed-form method obtains pixel cosine 0.9388
and requires 0.9339 seconds for all 16 online queries, or 0.0584 seconds per
query in the measured batched application run.
Raw VJP has pixel cosine $2.5\times10^{-5}$ on the same set.
The optimisation result is not called an incorrect preimage.  It demonstrates
that target-only preimage search and source-local feature inversion use
different information and can select different solutions of an underdetermined
feature constraint.

\paragraph{Initialisation sensitivity and post-calibration latency.}
The three optimisation seeds reach similar target objectives, while their
outputs remain seed dependent.  We therefore report the predeclared seed mean
as the baseline result and retain the median only as a diagnostic.
In an isolated three-repeat benchmark, closed-form batch-one online latency is
0.1385 seconds when images vary,
0.1104 seconds when depths vary, 0.1408 seconds when channels vary, and
0.1274 seconds for a mixed workload.
The corresponding 2,000-step iterative-optimisation references require
23.5421, 21.7281, 23.5781, and 23.0087 seconds per query.
All values in this comparison begin after calibration so that they measure the
deployment-time scaling of each query type.

\begin{figure*}[t]
  \centering
  \includegraphics[width=\textwidth]{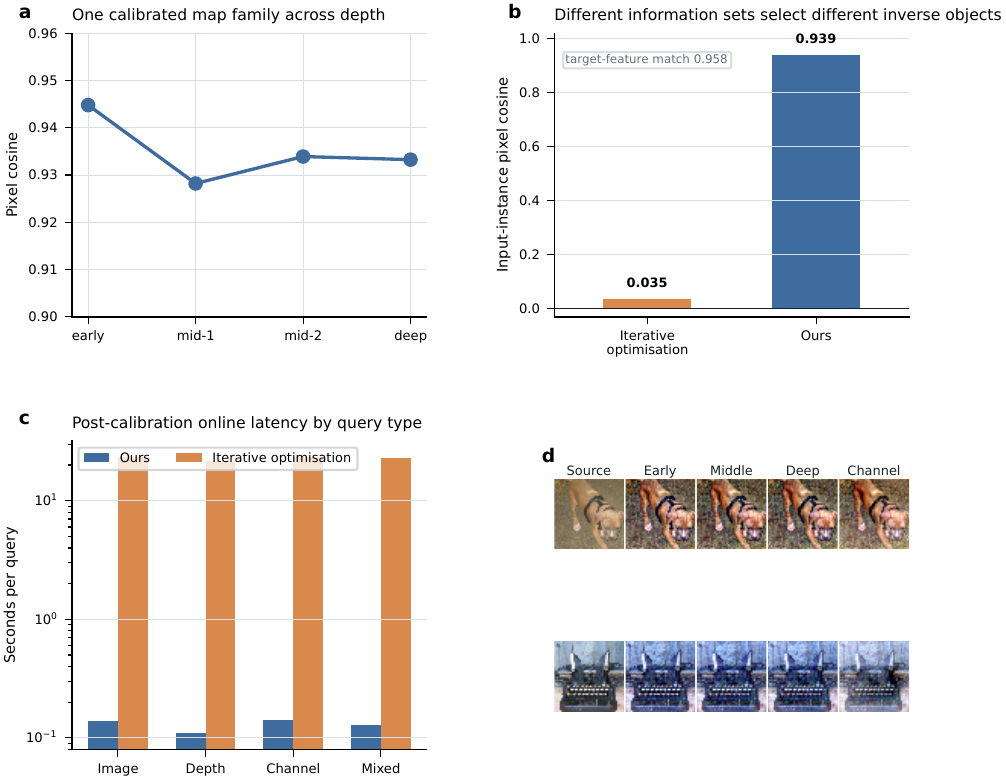}
  \caption{\textbf{Expanded no-refit reuse and optimisation-regime evidence.}
  \textbf{a}, One ResNet50 calibration across four full-state depths.
  \textbf{b}, Descriptive input-instance alignment for the matched iterative
  and closed-form rows; the iterative target-feature match is annotated
  separately because it is a representation-consistency diagnostic.
  \textbf{c}, Post-calibration online latency across image, depth, channel, and
  mixed queries.
  \textbf{d}, Fixed depth and channel queries under one fitted map family.}
  \label{fig:supp_reuse_baseline}
\end{figure*}

Figure~\ref{fig:supp_reuse_baseline} combines the no-refit atlas and fixed
depth/channel queries with the matched regime and online-latency comparisons.

\FloatBarrier
\subsection{Channel--Coordinate Query Generalisation in ViT}
\label{app:vit_coordinate_queries}

\paragraph{Protocol and support definition.}
The formal patch-coordinate experiment used the ImageNet-pretrained
torchvision ViT-B/32 checkpoint and a target token stream of shape
$B\times50\times768$, comprising one CLS token and a $7\times7$ patch grid.
The hidden width remained the channel axis and the token index remained the
coordinate and Fourier axis.  Each query retained all hidden channels and one
non-CLS patch coordinate,
\[
  S=\{1,\ldots,768\},
  \qquad
  Q=\{q\},
  \qquad
  q\in\{1,\ldots,49\}.
\]
One isolated c4096 map family was reused for all queries, images, and target
depths without patch-specific refitting.  Evaluation used 128 fixed, correctly
classified ImageNet validation images and layers 2, 8, and 14, giving
$128\times49\times3=18{,}816$ inversions.

For input patch $r$, the unnormalised support of inverse $\widehat x$ is
\[
  M_r(\widehat x)
  =
  \sum_{(h,w)\in r}\|\widehat x_{:,h,w}\|_2.
\]
The patch-query input-domain support matrix is then
\[
  A_l(q,r)
  =
  \frac{M_r(\widehat x_{l,q})}
       {\sum_{r'}M_{r'}(\widehat x_{l,q})}.
\]
Thus the reported normalised inverse-magnitude support mass first computes a
channel $L_2$ magnitude at every pixel, sums it inside each $32\times32$ input
patch, and normalises over the 49 input patches.  It is not squared inverse
energy.  The matrix reports where a patch-selected inverse is expressed in
input coordinates; it is not attention, attribution, or causal importance.

\paragraph{All 49 patch queries.}
The average layer-2 support matrix was strongly diagonal and locally banded.
With depth, self-patch and local-neighbour mass decreased while profile entropy
and expected support distance increased
(\cref{tab:vit_patch_query_support,fig:channel_coordinate_queries}).
The direct observation is a transition from local, coordinate-aligned support
to distributed input-domain support.  Contextual mixing is a compatible
interpretation of the deeper pattern, but semantic content is not inferred
from these scalar summaries alone.

\begin{table*}[h!]
  \centering
  \caption{\textbf{ViT patch-query input-domain support from one c4096 calibration.}
  Entries are means with 95\% image-bootstrap intervals over 128 fixed
  validation images.  Support radius is the expected patch-grid distance
  normalised by the maximum grid distance.}
  \label{tab:vit_patch_query_support}
  \scriptsize
  \setlength{\tabcolsep}{7.0pt}
  \begin{tabular}{lcccc}
    \toprule
    Layer & Self-patch mass & Profile entropy & Support radius & Local $3\times3$ mass \\
    \midrule
    2  & 0.417 (0.413--0.421) & 0.649 (0.645--0.654)
       & 0.183 (0.181--0.184) & 0.720 (0.716--0.725) \\
    8  & 0.139 (0.136--0.143) & 0.915 (0.912--0.918)
       & 0.324 (0.321--0.327) & 0.396 (0.389--0.402) \\
    14 & 0.065 (0.063--0.067) & 0.964 (0.962--0.965)
       & 0.376 (0.374--0.378) & 0.273 (0.268--0.278) \\
    \bottomrule
  \end{tabular}
\end{table*}

\begin{figure*}[t]
  \centering
  \includegraphics[width=\textwidth]{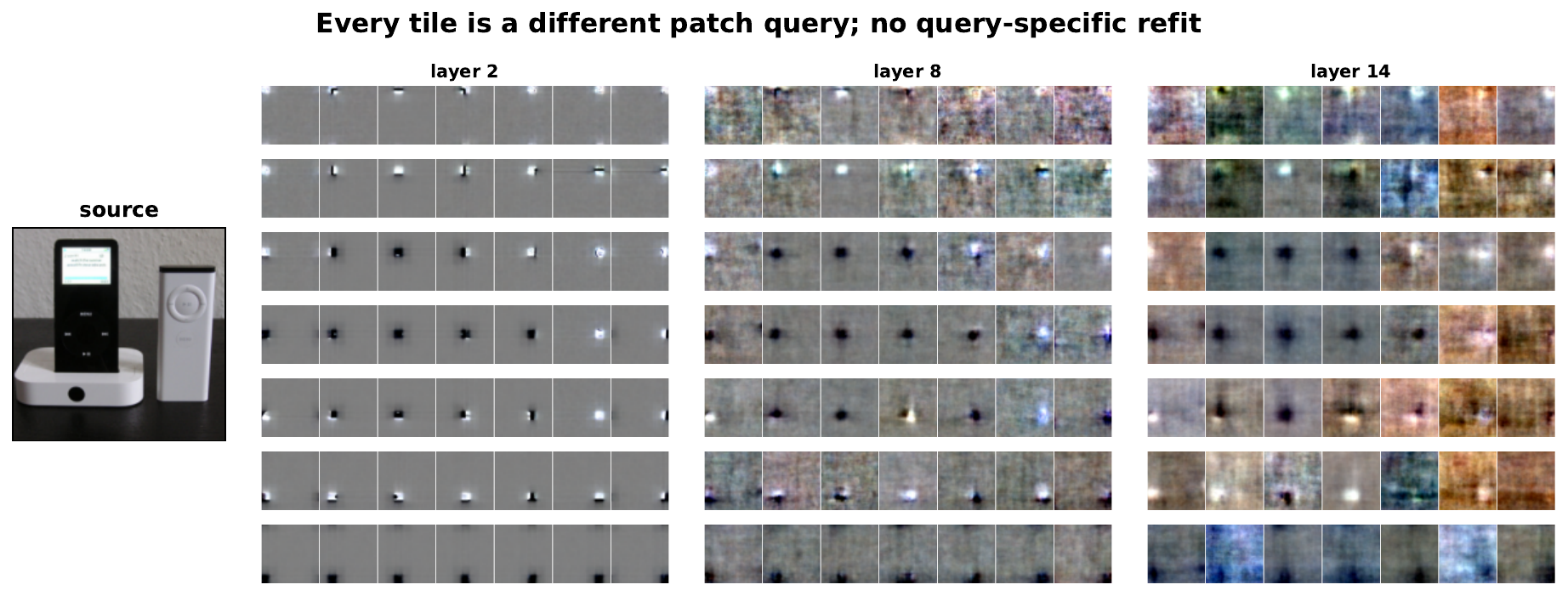}
  \caption{\textbf{Complete 49-query ViT inverse atlas for one fixed image.}
  The source is followed by every non-CLS patch query at layers 2, 8, and 14,
  ordered on the $7\times7$ patch grid.  All 147 inverses use the same c4096
  map family with no query-specific refit.  Each inverse is displayed with the
  same signed visualisation rule and its own fixed 99th-percentile magnitude
  scale; quantitative support is measured before display scaling.}
  \label{fig:vit_patch_query_atlas}
\end{figure*}

\paragraph{Independent internal intervention.}
For the same 128 images, every one of the 49 patch-token vectors was
independently zeroed at layers 2, 8, 11, and 14 and the fixed top-1--top-2
class margin was recomputed, giving 25,088 interventions.  Gradient times
activation was computed on the original representation and used only to
identify the score-top and score-counter patch in each image; exact margin drop
was measured by a separate forward evaluation.  A deterministic random patch
provided the paired reference.

\begin{table*}[h!]
  \centering
  \caption{\textbf{Exhaustive ViT patch-token intervention.}
  Entries are means with 95\% image-bootstrap intervals over 128 images.
  Spearman correlation compares the gradient-times-activation ranking with all
  49 exact intervention effects within each image.  It is undefined at layer
  14 because every patch-token intervention is numerically zero under the
  CLS-only readout.}
  \label{tab:vit_patch_intervention}
  \scriptsize
  \setlength{\tabcolsep}{9.0pt}
  \begin{tabular}{lccc}
    \toprule
    Layer & Score-top $-$ random & Score-counter $-$ random & Score/drop Spearman \\
    \midrule
    2  & 0.220 (0.031--0.420) & 0.029 ($-0.067$--0.130)
       & 0.134 (0.104--0.164) \\
    8  & 0.373 (0.217--0.544) & $-0.268$ ($-0.388$--$-0.162$)
       & 0.354 (0.319--0.388) \\
    11 & 0.234 (0.141--0.323) & $-0.148$ ($-0.243$--$-0.077$)
       & 0.372 (0.324--0.419) \\
    14 & 0.000 (0.000--0.000) & 0.000 (0.000--0.000) & --- \\
    \bottomrule
  \end{tabular}
\end{table*}

Layer 8 showed the clearest signed separation: score-top patch removal reduced
the class margin relative to random removal, whereas score-counter removal
shifted it in the opposite direction.  Layer 11 retained both effects.
Layer-14 patch-token interventions were null to numerical precision
(mean scale approximately $3\times10^{-8}$), as expected because the final
torchvision classifier reads only the CLS token after the last normalisation.
This is a patch-token null control under a CLS-only readout; no CLS ablation is
included in the formal c4096/e128 experiment.

\begin{figure*}[t]
  \centering
  \includegraphics[width=\textwidth]{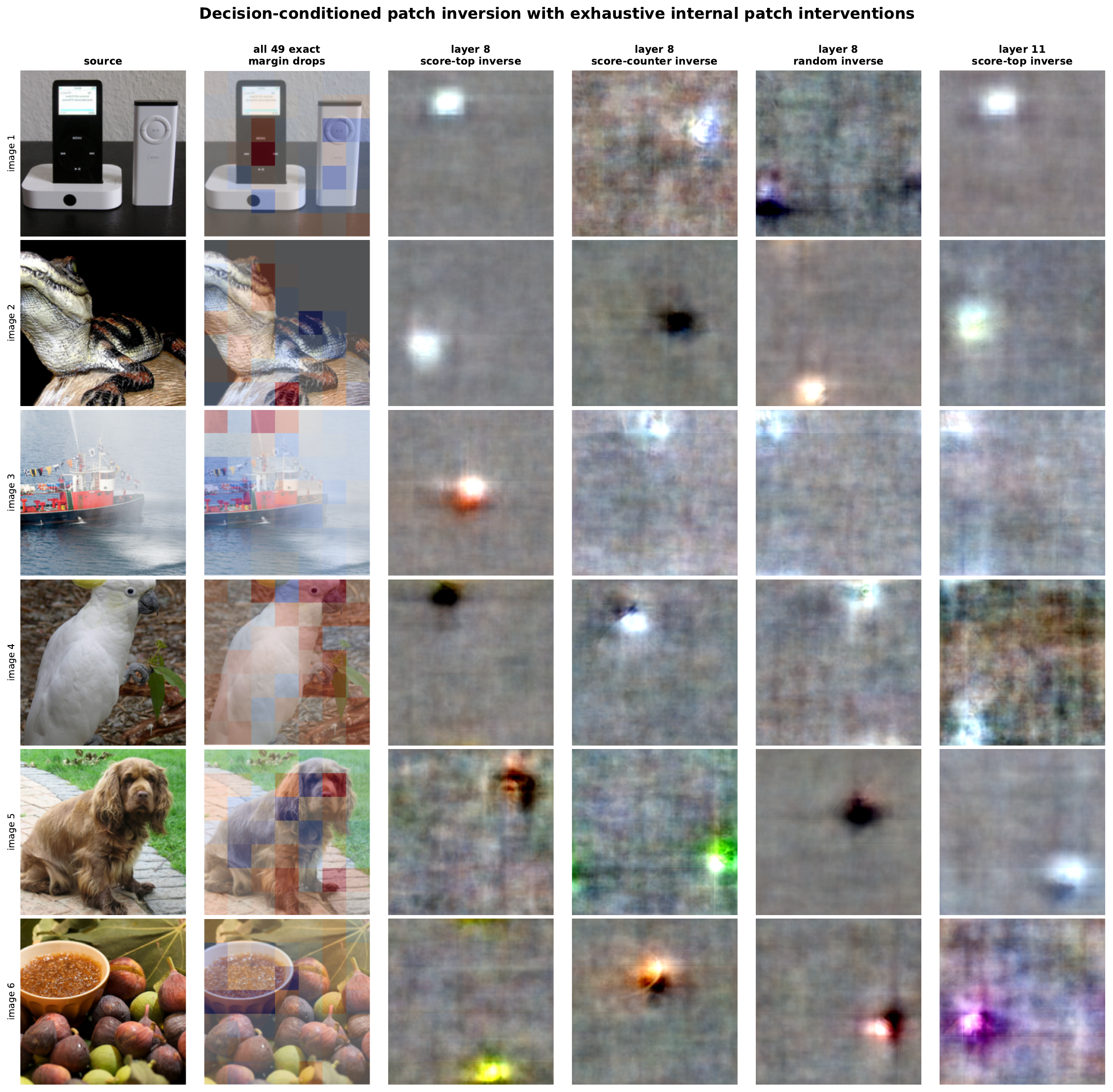}
  \caption{\textbf{Patch-query inverses paired with exhaustive internal interventions.}
  Rows are the first six fixed evaluation images.  The second column overlays
  all 49 exact layer-8 patch-token margin drops on the source.  Subsequent
  columns show the layer-8 score-top, score-counter, and random-query inverses,
  followed by the layer-11 score-top inverse.  The inverse panels are
  descriptive input-domain views; causal interpretation is restricted to the
  separately evaluated internal patch-token interventions.}
  \label{fig:vit_patch_decision_atlas}
\end{figure*}

\FloatBarrier
\subsection{Expanded Architecture and Tensor-Component Results}
\label{app:component_zoo}

Table~\ref{tab:supp_headline_setups} records the exact terminal tensor and axis
convention for every headline ImageNet backbone
\citep{he2016deep,huang2017densely,sandler2018mobilenetv2,
tan2019efficientnet,radosavovic2020designing,liu2022convnet,
liu2021swin,dosovitskiy2021image}.

\begin{table*}[t]
  \centering
  \caption{\textbf{Headline ImageNet terminal and boundary setups.}
  All rows use full-state terminal targets, c4096/e1024, and registered
  model-native boundary DAGs.  MobileNetV2 and EfficientNet-B0 factorise each
  stage-entry non-residual block at expansion/depthwise tensors, retain residual
  blocks at their add endpoints, and terminate at the final 320-channel feature
  block.  Tensor shapes include the batch dimension;
  channel and coordinate axes use zero-based indexing.}
  \label{tab:supp_headline_setups}
  \scriptsize
  \setlength{\tabcolsep}{3.5pt}
  \begin{tabular}{lllllr}
    \toprule
    Model & Captured terminal tensor & Shape & Channel axis & Coordinate axes & Fitted nodes \\
    \midrule
    ResNet18 & \texttt{op.add\_7.add:out} & $1\times512\times7\times7$ & 1 & 2,3 & 9 \\
    ResNet50 & \texttt{op.add\_15.add:out} & $1\times2048\times7\times7$ & 1 & 2,3 & 17 \\
    DenseNet121 & \texttt{op.batch\_norm\_120.batch\_norm:out} & $1\times1024\times7\times7$ & 1 & 2,3 & 9 \\
    MobileNetV2 & \texttt{features.17.conv.3:out} & $1\times320\times7\times7$ & 1 & 2,3 & 31 \\
    EfficientNet-B0 & \texttt{features.7.0.block.3.1:out} & $1\times320\times7\times7$ & 1 & 2,3 & 30 \\
    RegNetY-1.6GF & \texttt{op.add\_26.add:out} & $1\times888\times7\times7$ & 1 & 2,3 & 28 \\
    ConvNeXt-B & \texttt{op.add\_35.add:out} & $1\times1024\times7\times7$ & 1 & 2,3 & 37 \\
    Swin-T & \texttt{norm:out} & $1\times7\times7\times768$ & 3 & 1,2 & 26 \\
    ViT-B/32 & \texttt{encoder.ln:out} & $1\times50\times768$ & 2 & 1 & 14 \\
    \bottomrule
  \end{tabular}
\end{table*}

The main paper reports nine ImageNet rows spanning eight backbone families.
All rows pass the final-code validator and use c4096/e1024 with the same
public recipe.  The full results are reproduced in
\cref{tab:supp_architecture_metrics} to expose architecture-dependent
structure rather than reducing generality to one scalar ranking.

\begin{table*}[t]
  \centering
  \caption{\textbf{Expanded ImageNet architecture results.}
  Y-low and Y-high are luminance low- and high-frequency cosines.
  All metrics are descriptive alignment over 1,024 images.}
  \label{tab:supp_architecture_metrics}
  \scriptsize
  \setlength{\tabcolsep}{3.6pt}
  \begin{tabular}{lrrrrrrr}
    \toprule
    Model & $G_v$ & Final & Y-low & Y-high & Chroma & SSIM & LPIPS \\
    \midrule
    ResNet18 & 0.9009 & 0.9375 & 0.9486 & 0.8624 & 0.9041 & 0.6837 & 0.3137 \\
    ResNet50 & 0.8889 & 0.9308 & 0.9178 & 0.8900 & 0.9260 & 0.7261 & 0.2772 \\
    DenseNet121 & 0.7510 & 0.8245 & 0.8335 & 0.7243 & 0.8593 & 0.5878 & 0.4188 \\
    MobileNetV2 & 0.7810 & 0.8333 & 0.8247 & 0.8484 & 0.8244 & 0.7103 & 0.2597 \\
    EfficientNet-B0 & 0.6591 & 0.7960 & 0.7398 & 0.8581 & 0.7729 & 0.7223 & 0.2720 \\
    RegNetY-1.6GF & 0.8716 & 0.9143 & 0.9500 & 0.6553 & 0.8947 & 0.5966 & 0.4231 \\
    ConvNeXt-B & 0.7343 & 0.8450 & 0.8626 & 0.7088 & 0.7724 & 0.6166 & 0.3985 \\
    Swin-T & 0.1783 & 0.5037 & 0.4844 & 0.6144 & 0.4514 & 0.5187 & 0.5198 \\
    ViT-B/32 & 0.3116 & 0.7021 & 0.7891 & 0.2480 & 0.6044 & 0.3144 & 0.7568 \\
    \bottomrule
  \end{tabular}
\end{table*}

A matched boundary audit showed that block-output-only capture was still too
coarse for stage-entry MobileNetV2 and EfficientNet-B0 blocks.
Stem-only changes and indiscriminate full-DAG refinement did not improve the
matched c512 diagnostic.
The decisive change was to expose the expansion/depthwise tensors of each
non-residual stage-entry block while leaving residual blocks at their add
endpoints.  For EfficientNet-B0, making the squeeze-and-excitation
multiplication itself a boundary caused sample-wise sign instability, so the
gate and multiplication remain inside one local edge.
The final c4096/e1024 rows reach pixel cosines 0.8333 and 0.7960 at the same
last 320-channel targets; no shallower target was substituted.

Swin-T and ViT-B/32 expose different Transformer feature geometry.
Swin has stronger SSIM and high-frequency structure, whereas ViT has stronger
low-frequency and global pixel alignment.  The result shows that the method
operates on Transformer computation graphs without asserting that hierarchy,
windowing, or patch size is the unique causal explanation for the difference.

\paragraph{Tensor-component zoo.}
The formal zoo contains 12 validated runs and 11 component families.
It includes DenseNet concatenations, MobileNet depthwise and inverted-residual
states, EfficientNet MBConv and squeeze-and-excitation states, ConvNeXt block
and normalisation-adjacent tensors, Swin window-stage and patch-merging states,
ViT token streams, and pooled or vector states.  Representative results are
shown in \cref{tab:supp_component_zoo}.  Every row uses the same tensor-axis
interface and records \texttt{manual\_component\_inverse=False}.

\begin{table*}[t]
  \centering
  \caption{\textbf{Representative component-zoo targets.}
  Shapes include the batch dimension.  Re-encoding is a representation
  diagnostic and is not required to track pixel alignment monotonically.}
  \label{tab:supp_component_zoo}
  \scriptsize
  \setlength{\tabcolsep}{3.5pt}
  \begin{tabular}{llllrr}
    \toprule
    Model & Component & Target shape & Target example & Pixel & Re-enc. \\
    \midrule
    DenseNet121 & concatenation & $1\!\times\!256\!\times\!56\!\times\!56$ & dense cat & 0.8338 & 0.8353 \\
    MobileNetV2 & inverted residual & $1\!\times\!160\!\times\!7\!\times\!7$ & residual add & 0.4674 & 0.1961 \\
    MobileNetV2 & depthwise tensor & $1\!\times\!32\!\times\!112\!\times\!112$ & depthwise conv & 0.9861 & 0.9745 \\
    EfficientNet-B0 & MBConv/SE & $1\!\times\!192\!\times\!7\!\times\!7$ & MBConv add & 0.3080 & 0.0248 \\
    ConvNeXt-B & normalisation-adjacent & $1\!\times\!56\!\times\!56\!\times\!128$ & layer norm & 0.9487 & 0.9445 \\
    ConvNeXt-B & MLP state & $1\!\times\!56\!\times\!56\!\times\!512$ & MLP fc1 & 0.9475 & 0.9814 \\
    Swin-T & window stage & $1\!\times\!14\!\times\!14\!\times\!384$ & residual add & 0.7516 & 0.6485 \\
    Swin-T & patch merging & $1\!\times\!28\!\times\!28\!\times\!384$ & merge pad & 0.8924 & 0.9049 \\
    EfficientNet-B0 & pooled vector & $1\!\times\!1280\!\times\!1\!\times\!1$ & average pool & 0.2622 & $-0.1710$ \\
    EfficientNet-B0 & logit vector & $1\!\times\!1000$ & classifier & 0.2640 & $-0.0050$ \\
    \bottomrule
  \end{tabular}
\end{table*}

The component table establishes coverage, not uniform visual alignment.
In particular, a pooled or logit vector contains a different selected feature
object from a spatial tensor.  Its inverse should therefore not be judged by
whether it reproduces the complete input.

\subsection{Distribution Behaviour and Calibration Transfer}
\label{app:distribution_transfer}

\paragraph{Dataset-trained representative ladder.}
The representative study recalibrates a dataset-trained model on each dataset
while keeping the method recipe fixed.  Final pixel cosines for
ImageNet/Pets/CUB are 0.9308/0.9566/0.9398 for ResNet50,
0.8450/0.2876/0.6507 for ConvNeXt-B, and
0.7021/0.7590/0.7728 for ViT-B/32.
The non-monotonic ConvNeXt result prevents a scalar ``dataset complexity''
interpretation.  The supported conclusion is that feature-inverse behaviour
depends jointly on the data distribution, checkpoint, and calibrated local
statistics.

\paragraph{Eight-family breadth with ImageNet weights.}
The separate breadth experiment applies ImageNet-weight models to Pets and CUB
and recalibrates their maps on the corresponding images.  All 16 rows pass the
strict artifact and protocol audit.  Final pixel cosine ranges from 0.3864 to
0.9451 on Pets and from 0.3684 to 0.9182 on CUB.  Swin reaches 0.6757 on Pets
and 0.6098 on CUB, while ViT reaches 0.7808 and 0.7193.  These rows broaden the
visual-distribution evidence but remain distinct from dataset-trained model
comparisons.

\begin{figure*}[t]
  \centering
  \includegraphics[width=\textwidth]{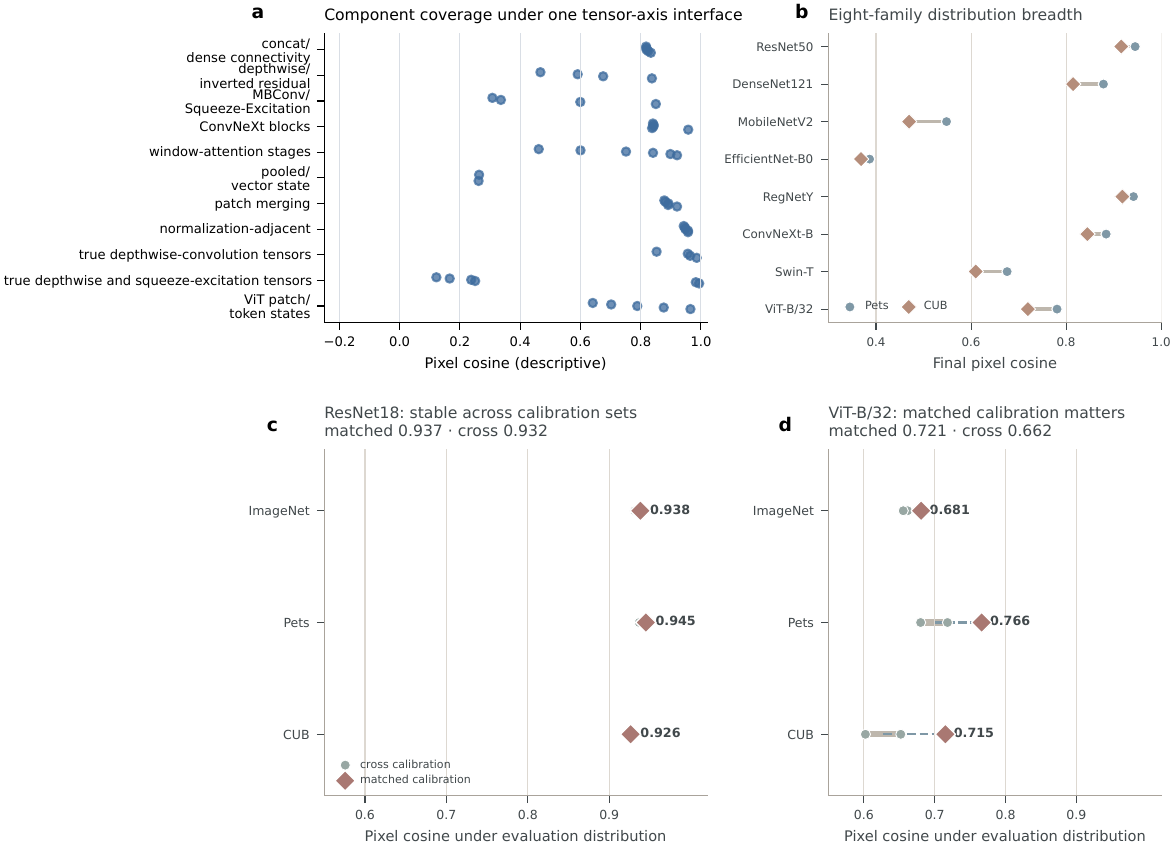}
  \caption{\textbf{Expanded component and distribution evidence.}
  \textbf{a}, All validated component-zoo targets grouped by tensor class.
  \textbf{b}, Paired Pets and CUB results for eight ImageNet-weight backbone
  families under dataset-specific recalibration.
  \textbf{c,d}, Calibration-transfer profiles for ResNet18 and ViT-B/32.  Each
  row retains the two cross-calibration values and the matched-calibration
  value for one evaluation distribution.}
  \label{fig:supp_arch_distribution}
\end{figure*}

\paragraph{Calibration-source transfer.}
We fitted maps on ImageNet, Pets, or CUB and evaluated each map family on all
three evaluation distributions.  The model checkpoint and method remained
fixed within each $3\times3$ matrix.
The complete transfer matrices are shown directly in
\cref{fig:supp_arch_distribution} rather than repeated in a separate table.

The matched-minus-cross average gap is small for ResNet18
(approximately 0.937 versus 0.932) and larger for ViT-B/32
(approximately 0.721 versus 0.662).  Thus fixed-map transfer is more sensitive
for the token model.  Recipe generality across datasets does not imply that one
map family is distribution invariant.

Figure~\ref{fig:supp_arch_distribution} collects the component-zoo breadth,
the ImageNet-weight distribution study, and both calibration-transfer matrices.

\subsection{Source-Path Faithfulness Under Low Pixel Alignment}
\label{app:source_path_faithfulness}

\paragraph{Question and protocol.}
Pixel cosine describes alignment between an inverse and the known
target-generating image, but does not directly test whether the reconstructed
states follow the local computation instantiated by that image.  We therefore
performed a dedicated source-geometry intervention on independently fine-tuned
ConvNeXt-B checkpoints for CUB-200-2011 and Oxford-IIIT Pets.  The checkpoints
reached 85.73\% and 95.42\% top-1 accuracy, respectively, excluding a
degenerate classifier as the explanation for low pixel alignment.  CUB used
4,096 calibration images and Pets used all 3,680 training images; both studies
evaluated 512 held-out images with the final full-channel, per-frequency
$G_v+D_v$ estimator, relative ridge 0.01, the 38-node ConvNeXt DAG, and terminal
node 37.  Seven preselected states spanning the path were probed at nodes
$\{1,4,7,20,34,35,36\}$.

\paragraph{Source-geometry intervention.}
For an evaluation pair $(A,B)$, the matched inverse used terminal target $H_A$
and source-local geometry $J_A$, whereas the geometry-swapped inverse retained
$H_A$ but replaced the operating-point geometry by $J_B$.  The primary path
test held the reconstructed states from $(H_A,J_A)$ fixed and evaluated their
JVP forward consistency on either the matched graph $J_A$ or the shuffled graph
$J_B$.  At every probed node, path NMSE is the normalised squared mismatch
between the JVP-predicted child state and the reconstructed child state, averaged
over the relevant children; lower values indicate greater compatibility with
that graph's local differential path.  This intervention does not require the
geometry-swapped output to become the image or feature identity of $B$:
$J_B$ specifies the complete local input-to-feature geometry, while the fixed
terminal state $H_A$ continues to modulate the selected inverse branch.

\begin{table*}[t]
  \centering
  \caption{\textbf{Source-path faithfulness in the lowest pixel-cosine
  quartile of ConvNeXt-B inverses.}
  Each row contains 128 of 512 held-out images.  Matched and shuffled path NMSE
  evaluate the same matched reconstructed states under $J_A$ and $J_B$,
  respectively.  Gap is shuffled minus matched NMSE, with a deterministic
  10,000-resample 95\% bootstrap interval.  Node win is the fraction of the
  seven probed nodes at which the shuffled-path NMSE exceeds the matched-path
  NMSE.  State margin is the cosine of a matched reconstructed state to the
  corresponding state of $A$ minus its cosine to the paired state of $B$.
  Branch cosine compares the matched and geometry-swapped input-domain outputs.}
  \label{tab:supp_source_path_faithfulness}
  \scriptsize
  \setlength{\tabcolsep}{4.0pt}
  \begin{tabular}{lrrrrrrr}
    \toprule
    Dataset & Pixel cosine & Matched NMSE $\downarrow$ & Shuffled NMSE
      & Gap [95\% CI] & Node win & State margin & Branch cosine \\
    \midrule
    CUB-200-2011 & $-0.135$ & 0.433 & 0.719
      & 0.286 [0.239, 0.336] & 0.800 & 0.343 & 0.334 \\
    Oxford-IIIT Pets & $-0.037$ & 0.443 & 0.636
      & 0.194 [0.155, 0.235] & 0.706 & 0.292 & 0.216 \\
    \bottomrule
  \end{tabular}
\end{table*}

\paragraph{Results.}
The lowest pixel-cosine quartile had mean pixel cosine $-0.135$ on CUB and
$-0.037$ on Pets, yet matched-path NMSE remained substantially below
shuffled-path NMSE on both datasets
(\cref{tab:supp_source_path_faithfulness}).  The 95\% bootstrap intervals
for the paired mean gaps excluded zero, and the matched source-state margins
remained positive.  The effect was strongest at deeper states: matched versus
shuffled path NMSE was 0.370 versus 0.986 on CUB and 0.411 versus 0.874 on
Pets.  Changing only the source-local geometry also selected substantially
different input-domain branches: in the lowest quartile, matched-versus-swapped
output cosine was 0.334 on CUB and 0.216 on Pets, with relative $\ell_2$
distances of 0.980 and 1.120.

Pixel cosine was nearly uncorrelated with the primary path diagnostics over all
512 images.  Its Spearman correlation with matched state margin was 0.012 on
CUB and 0.051 on Pets; with the log shuffled-to-matched path-NMSE ratio it was
$-0.049$ and $-0.035$; and with node-win fraction it was $-0.042$ and 0.034.
Thus poor input-instance alignment did not identify samples with weak
source-path consistency.

\paragraph{Boundary of the claim.}
The comparison supports path-level faithfulness, not a claim that every local
boundary must favour the matched graph.  Six of the seven probed nodes had
lower mean NMSE under $J_A$ on both datasets, whereas the early node 4 showed a
systematic reversal.  Pets therefore had a small shallow-state reversal
(matched/shuffled NMSE 0.466/0.458) despite a large deep-state advantage.  In
addition, geometry-swapped deep states remained modulated by the fixed terminal
target rather than globally changing into the states of $B$.  These observations
are consistent with the intended object: the source-local geometry selects an
inverse branch, while the terminal feature modulates the state transported
through that branch.  The experiment therefore shows that low pixel cosine
alone is not a sufficient failure diagnostic.  The low-alignment ConvNeXt
outputs retain measurable fidelity to the target-generating source path.  This
result does not assert exact recovery of a unique input or uniformly lower
error at every boundary.

\subsection{Topology, Two-Stage Mechanism, and Operator Ablations}
\label{app:mechanism_ablations}

\paragraph{Local topology versus a global repair.}
On the matched ResNet18 terminal target, raw VJP has pixel cosine 0.00019.
A single global $G$ reaches 0.07221, and a global $G+D$ reaches 0.06714.
Local $G_v$ reaches 0.90090, while local $G_v+D_v$ reaches 0.93795.
The comparison holds operator calibration and target data fixed while changing
whether repair is factorised along the fitted DAG.  It therefore isolates local
topological composition from the mere existence of a closed-form map.

\paragraph{Second-stage residual repair.}
On ResNet18, $G_v$ alone reaches a pixel cosine of 0.9009 and the second stage
adds 0.0367.  The corresponding gain increases to 0.1107 on ConvNeXt-B and
0.3905 on ViT-B/32.  Mean child NMSE decreases by 23.0\%,
15.4\%, and 16.2\%, while intermediate-state relative $\ell_2$ decreases by
1.52\%, 0.45\%, and 5.92\%.  This architecture-level ordering is consistent
with a larger role for JVP-FC when normalisation and attention induce strongly
sample-dependent local operators.  It does not
isolate any individual module as the cause because architecture and component
type are not independently controlled.  JVP-FC therefore improves the
aggregate local residual and end-to-end result across all three representative
architectures.
Boundary-level correlations are not uniformly positive.  Across 86 pooled
boundaries, the Spearman correlation between the pre-correction residual and
state-error reduction is 0.276 with bootstrap interval
$[0.068,0.462]$; its correlation with residual reduction is $-0.235$ with
interval $[-0.462,0.004]$.  These results support JVP-FC as an effective second
projection but not as a monotonic per-boundary causal law.

\paragraph{Operator family.}
The complete matched operator comparison is reported in
\cref{tab:supp_operator_family}.  Full channel mixing and frequency specificity
are separately required for strong input-instance alignment at the
representative calibration budget.

\begin{table*}[t]
  \centering
  \caption{\textbf{Restricted operator families.}
  Entries are final pixel cosine at c4096/e1024.  Restricted rows remain valid
  feature-inverse computations within their operator class; the values measure
  the input-aligned structures retained by that class.}
  \label{tab:supp_operator_family}
  \small
  \begin{tabular}{lrrr}
    \toprule
    Operator & ResNet18 & ConvNeXt-B & ViT-B/32 \\
    \midrule
    Gamma-only & 0.0062 & 0.0034 & $-0.0044$ \\
    Diagonal, per-frequency & 0.2526 & 0.0961 & 0.0620 \\
    Diagonal + rank-8, per-frequency & 0.8337 & 0.4763 & 0.0501 \\
    Full-channel, shared-frequency & 0.2709 & 0.2090 & 0.3366 \\
    Full-channel, per-frequency & \textbf{0.9375} & \textbf{0.8450} & \textbf{0.7021} \\
    \bottomrule
  \end{tabular}
\end{table*}

Rank-8 recovers much of the ResNet result but not the ViT result.
This architecture dependence becomes stronger at small calibration budgets.
Repeated ResNet18 seeds reproduce the Gamma-only result near 0.006, excluding
an accidental calibration subset as the explanation.  The visual outputs are
retained alongside these scalar values because restricted operators can retain
coarse spatial structure while differing in scale, colour, or frequency.

\paragraph{Frequency alignment control.}
Replacing the per-frequency maps with one coordinate-collapsed map reduces
pixel cosine from 0.9375 to 0.2709 on ResNet18 and from 0.7021 to 0.3366 on
ViT-B/32.  Permuting otherwise fitted frequency maps reduces the values further
to 0.00025 and 0.0288.  The result shows that map capacity alone is insufficient;
the learned second-order operator must remain aligned with the coordinate
frequency on which it was calibrated.

\paragraph{Mean-seed gauge.}
The seed-source ablation includes heterogeneous-channel, multi-child nodes.
On ResNet18, the full-state mean seed and an unweighted constant seed give
numerically identical final pixel cosine, 0.9740235 in both cases.
This is expected because a constant full-state rescaling is absorbed when
separate zero-intercept relative-ridge maps are refitted.
The mean gauge is retained because it defines the reusable terminal-subset API
$s_T=P_{S,Q_{\mathrm{all}}}H_T/|S|$ across channel masks, where
$Q_{\mathrm{all}}$ denotes the complete coordinate set.  It is not claimed to
improve full-state quality over an arbitrary constant gauge.  Unit-energy
seeding is nonfinite on the tested ResNet configuration, while a fixed uniform
seed produces zero output, confirming that the source must remain target
dependent.

\paragraph{Nested-target composition.}
Restarting the same calibrated inverse at an intermediate repaired state and
continuing to the deeper output is numerically identical to the original
continuation: output cosine is at least 0.999999996 on ResNet18 and ViT-B/32.
Directly beginning at the shallower target gives 0.9955 output cosine to the
deep continuation on ResNet18 and 0.8689 on ViT-B/32.  Thus the implementation
is compositionally consistent when the same repaired state is reused, while
changing the declared target changes the feature-inverse object.

\subsection{Metric-Aware Theory Bridge and Finite-DAG Stability Diagnostics}
\label{app:theory_bridge_experiments}

\paragraph{Metric-aware local oracle.}
We evaluated the natural-adjoint construction on held-out local perturbations
at 16 operating points.  In a tractable projected problem, the local LMMSE
operator and the metric-aware natural adjoint agree to numerical precision.
Both reach mean NMSE 0.0291, cosine 0.9858, and relative $\ell_2$ 0.1590.
The raw Euclidean adjoint reaches NMSE 0.9999 and relative $\ell_2$ 1.0000.
A calibrated scalar rescaling improves these values to 0.5226 and 0.7191 but
does not close the metric gap.  The deployed observable-plus-$G$ surrogate
reaches NMSE 0.4416, cosine 0.7682, and relative $\ell_2$ 0.6572.

This experiment has a deliberately limited role.  It verifies the algebraic
bridge between a covariance-induced natural adjoint and a local LMMSE inverse
in a tractable projected setting.  It does not claim that the deployed
distributional surrogate estimates $M_u$ and $M_v$ online or equals the local
oracle at every operating point.

\paragraph{Finite-DAG diagnostics.}
For each fitted boundary we recorded path depth, a conservative product-of-gain
proxy, local condition statistics, child NMSE, intermediate-state error, and
final pixel error.  The proxy spans several orders of magnitude and describes
the scale of the fitted reverse gains.  However, it does not reliably
rank final error across all nodes or architectures.  The pooled Spearman
correlation between pre-correction residual and state-error reduction is 0.276,
with a 95\% bootstrap interval of $[0.068,0.462]$.  Consequently, the empirical
path-gain study is reported as a stability diagnostic, not as a contraction
certificate or proof of global convergence.  The formal finite-DAG result in
\cref{app:finite_dag_stability} remains a conditional error-propagation bound.

\begin{figure*}[t]
  \centering
  \includegraphics[width=\textwidth]{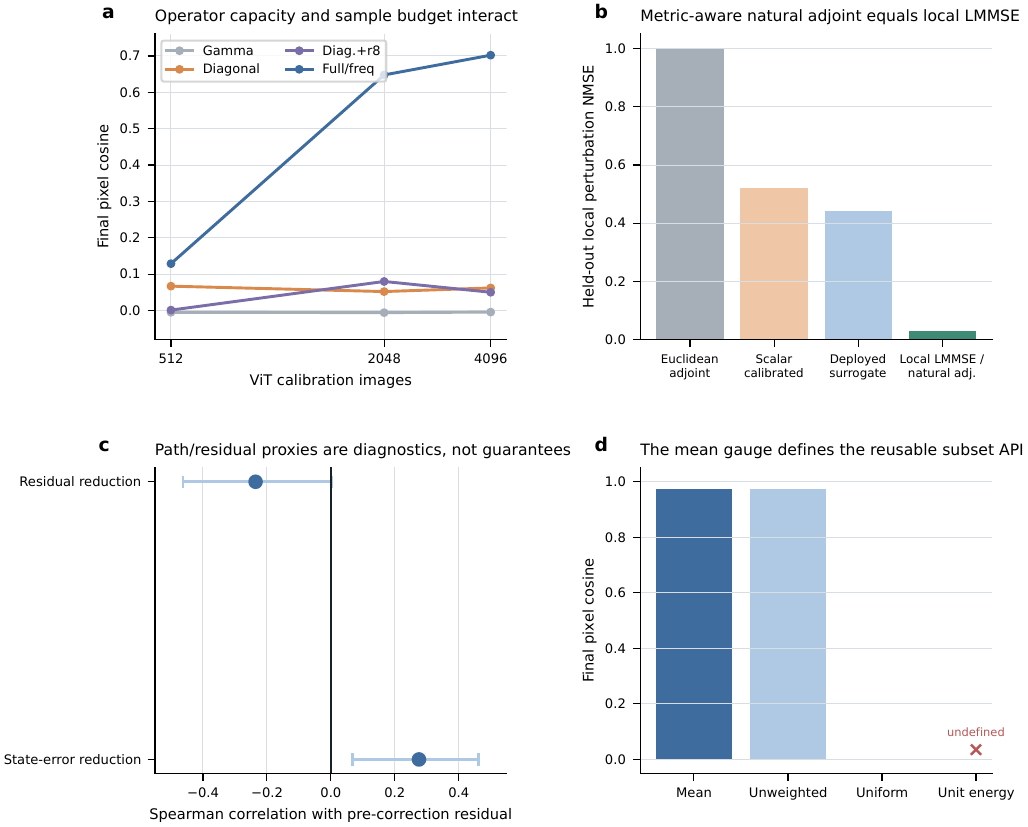}
  \caption{\textbf{Operator capacity, theory bridge, and stability diagnostics.}
  \textbf{a}, ViT operator-capacity by calibration-budget frontier.
  \textbf{b}, Held-out local perturbation NMSE for Euclidean, deployed, and
  metric-aware oracle operators.
  \textbf{c}, Pooled boundary correlations with 95\% bootstrap intervals.
  \textbf{d}, Multi-child seed-source ablation.}
  \label{fig:supp_mechanism_theory}
\end{figure*}

Figure~\ref{fig:supp_mechanism_theory} places the operator-capacity frontier,
metric-aware oracle, boundary correlations, and seed-source controls side by side.

\subsection{Input Alignment, Representation Consistency, and Information-Flow Controls}
\label{app:inverse_controls}

\paragraph{Zero and shuffle controls.}
The complete controls are summarised in \cref{tab:supp_information_flow}.
All zero targets produce exactly zero output, so the zero-intercept maps cannot
inject a target-independent additive image template.  Source shuffling moves
the output toward the new operating point.  Terminal-state shuffling produces
architecture-dependent modulation, which separates the role of the target
state from that of the source-local operators.

\begin{table*}[t]
  \centering
  \caption{\textbf{Information-flow controls.}
  All rows use the matched c4096/e512 target/operator-control protocol.
  ``Same'' is the matched target and operating point.  Terminal shuffle reports
  alignment to the fixed operating-point input and to the shuffled target
  image.  Source shuffle reports alignment to the old input and new source.}
  \label{tab:supp_information_flow}
  \scriptsize
  \begin{tabular}{lrrrrrr}
    \toprule
    Model & Same & Zero norm ratio & Term. $\to$ input & Term. $\to$ target
      & Source $\to$ old & Source $\to$ new \\
    \midrule
    ResNet18 & 0.9380 & 0 & 0.9379 & 0.0225 & 0.0186 & 0.9380 \\
    Swin-T & 0.3890 & 0 & 0.3492 & 0.0312 & 0.0096 & 0.5056 \\
    ViT-B/32 & 0.6922 & 0 & 0.0995 & 0.2494 & 0.0158 & 0.6922 \\
    \bottomrule
  \end{tabular}
\end{table*}

\paragraph{Target--operator swap.}
For paired images $A$ and $B$, we cross $H_A,H_B$ with operator families
$J_A,J_B$.  ResNet18 conditions $H_A,J_B$ and $H_B,J_A$ align with the
operator-point images at 0.9386 and 0.9386, while alignment with the target-state
images is 0.0340 and 0.0316.  Swin shows the same direction at 0.5509 and 0.5526.
ViT has stronger target modulation: the mismatched conditions have paired
identity values 0.2495/0.0959 and 0.1040/0.2358.
These results show that source-local geometry selects
the inverse branch while the selected terminal state modulates that branch.

\paragraph{Continuous operating-point locality.}
On 128 fixed ResNet18 image pairs, the operating point is interpolated from
$A$ to $B$ while the terminal target remains $H_A$.
Alignment to $A$ decreases monotonically from 0.9402 to 0.0021, while alignment
to $B$ increases from $-0.0036$ to 0.9402.  The two curves cross at interpolation
coefficient 0.495.  At every point, alignment to the current operating-point
image remains between 0.9251 and 0.9402.  By contrast, same-image noise with
relative magnitude up to 0.1 changes alignment to the clean source by at most
0.00066.  The curve therefore characterises the local branch selected by the
operating point.  Its child residual does not increase monotonically with
mismatch and must not be used as a source-match detector.

\paragraph{Channel diversity and representation checks.}
Mean pairwise pixel cosine among three fixed channel inverses is 0.4969 for
ResNet18, 0.2304 for ViT-B/32, and 0.0993 for Swin-T.  The outputs therefore do
not collapse to one map-dependent template.  Re-encoding is reported against
architecture-specific random-pair nulls and remains separate from direct input
alignment.  The ROC AUC distinguishing matched from random-pair re-encoding is
1.000, 0.996, and 0.947 for ResNet18, Swin-T, and ViT-B/32, respectively.
A high re-encoding value establishes representation consistency, while a low
value does not prove that the selected feature inverse is false.

\paragraph{Failure taxonomy.}
The fixed 1,024-image evaluation sets are stratified by low-frequency
structure, weak high-frequency alignment, and chroma drift.
ConvNeXt and Swin low-alignment cases are often concentrated in colour or fine
texture, whereas deep ViT outputs are dominated more strongly by low-frequency
and patch-grid structure.  These categories describe architecture-dependent
feature-inverse geometry.  Mismatch with the complete input is not, by itself,
treated as feature-inversion failure.

\begin{figure*}[t]
  \centering
  \includegraphics[width=\textwidth]{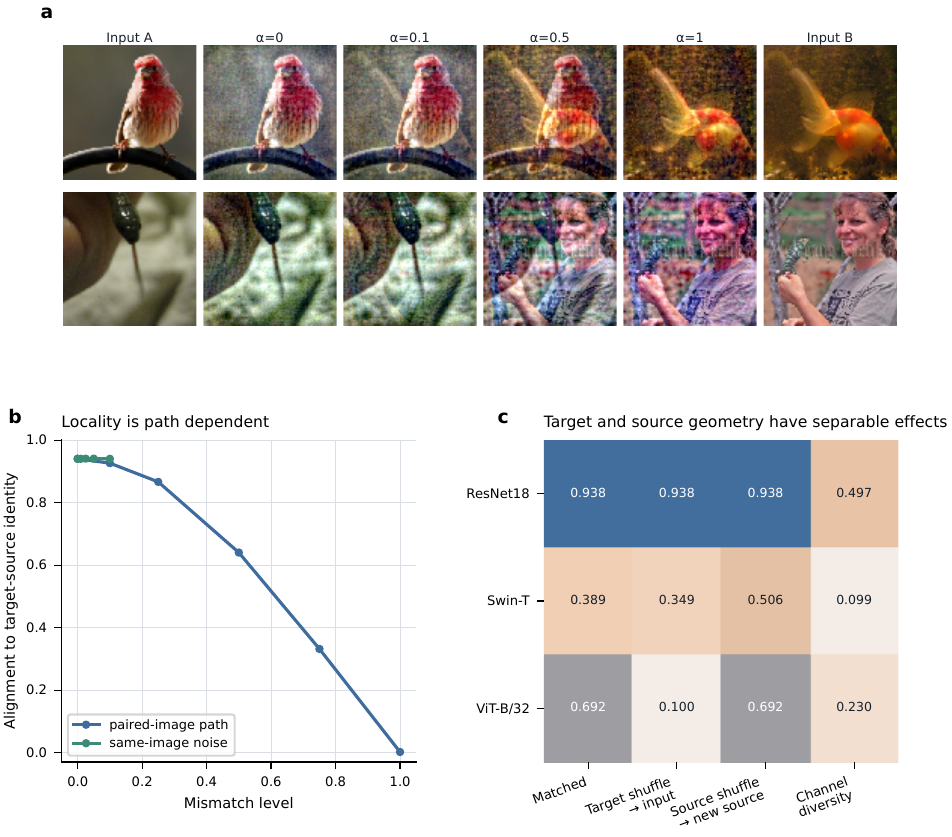}
  \caption{\textbf{Expanded operating-point locality and information-flow controls.}
  \textbf{a}, Fixed-target inverses along two paired-image operating-point paths.
  \textbf{b}, Paired-image interpolation versus same-image noise.
  \textbf{c}, Matched, shuffle, and channel-diversity summaries across three
  architectures.}
  \label{fig:supp_controls_locality}
\end{figure*}

Figure~\ref{fig:supp_controls_locality} expands the paired-image paths,
same-image noise control, shuffle conditions, and channel-diversity summaries.

\subsection{Calibration, Boundary, Ridge, Online Runtime, and Memory}
\label{app:resource_scaling}

\paragraph{Calibration size.}
The paired vision-Transformer sweep uses calibration sizes
$\{512,1024,2048,4096,8192\}$.  ViT-B/32 final pixel cosine is
0.1287, 0.3626, 0.6477, 0.7021, and 0.7243.  Swin-T reaches
0.0467, 0.2901, 0.4575, 0.5037, and 0.5259.
The largest increase occurs between 1,024 and 2,048 images, while the
4,096-to-8,192 gain is approximately 0.022 for both models.
This supports c4096 as a diminishing-return operating point, not as a universal
sample optimum.

Independent c4096 calibration subsets yield pixel-cosine standard deviations
of 0.00030 for ResNet18, 0.00362 for ConvNeXt-B, and 0.00464 for ViT-B/32.
The corresponding SSIM standard deviations are 0.00178, 0.00060, and 0.00122.
Thus the reported architecture differences are larger than the observed
three-subset calibration variation.

\paragraph{Boundary density.}
At c512/e1024, the ResNet18 coarse boundary uses 9 fitted nodes and reaches
pixel cosine 0.9293.  The full-DAG mode uses 49 fitted nodes and reaches 0.9631.
SSIM increases from 0.6747 to 0.7913 and LPIPS decreases from 0.3255 to
0.1664, while post-calibration online application increases from 53.9 to
74.7\,ms/image.
This diagnostic shows that finer boundaries can improve the inverse while
raising online application, map, and topology-onboarding demands.  The coarse
policy is therefore a practical default, not a theoretical claim that finer
factorisations cannot help.

\paragraph{Ridge sensitivity.}
At c512, ResNet18 is stable over $\rho\in\{0.003,0.01,0.03\}$, with final pixel
cosines 0.9247, 0.9278, and 0.9293.  At c1024, ConvNeXt-B reaches 0.7233,
0.7584, and 0.7674.  ViT-B/32 is more sensitive at the same small calibration
budget, increasing from 0.0726 to 0.3124 and 0.5258.  This interaction explains
why low-capacity or low-calibration token-model results should not be used to
claim universal operator compression.

\paragraph{Online runtime, storage, and memory.}
The expanded resource audit is shown in \cref{tab:supp_resources}.
Online times correspond to post-calibration application in c4096/e1024 quality
runs.  Storage includes both stages.

\begin{table*}[t]
  \centering
  \caption{\textbf{Expanded resource audit.}
  Online time is the measured 1,024-image application total divided by 1,024
  after calibration; it is not isolated batch-one latency.  Storage includes
  both $G_v$ and $D_v$.  Peak memory is GiB.}
  \label{tab:supp_resources}
  \scriptsize
  \begin{tabular}{lrrrrr}
    \toprule
    Model & Nodes & Online (ms/image) & Storage (GiB)
      & Calibration peak (GiB) & Eval peak (GiB) \\
    \midrule
    ResNet18 & 9 & 58.4 & 0.83 & 1.24 & 1.71 \\
    ResNet50 & 17 & 223.7 & 25.42 & 19.44 & 17.09 \\
    DenseNet121 & 9 & 104.1 & 5.84 & 8.79 & 7.42 \\
    MobileNetV2 & 31 & 155.0 & 4.35 & 6.46 & 5.39 \\
    EfficientNet-B0 & 30 & 164.0 & 5.43 & 7.63 & 6.79 \\
    RegNetY-1.6GF & 28 & 114.2 & 4.29 & 4.79 & 5.21 \\
    ConvNeXt-B & 37 & 369.7 & 15.51 & 16.69 & 16.78 \\
    Swin-T & 26 & 129.0 & 5.98 & 6.57 & 6.56 \\
    ViT-B/32 & 14 & 105.1 & 2.99 & 3.79 & 3.61 \\
    \bottomrule
  \end{tabular}
\end{table*}

Full-channel maps make calibration demand, storage, and online application
architecture dependent.  The practical benefit is therefore a change from
repeated per-target search to direct queries after calibration, not zero total
computation or memory.

\subsection{Deployment Validity, Oracle Gaps, and Implementation Quality Control}
\label{app:deployment_qc}

\paragraph{Public-path equivalence.}
Across all nine architecture implementations, the public inversion call and
the directly instrumented reference call agree to pixel cosine
$0.99999997$ or higher.  Maximum absolute differences are zero or below
$6.2\times10^{-7}$.  This audit verifies that the reported output is produced
by the public API rather than a private visualisation branch.

\paragraph{Legal online child state.}
At deployment, each $D_v$ source uses the repaired online child state, not a
cached true forward activation.  The same-map true-child condition is retained
only as a nondeployable c4096/e512 diagnostic upper reference.  On ResNet18, replacing
the online child with the true child changes pixel cosine from 0.9382 to 0.9391
and target re-encoding from 0.6522 to 0.7570.  On ViT-B/32, the same diagnostic
changes pixel cosine from 0.7005 to 0.9410 and re-encoding from 0.2481 to 0.7289.
The larger ViT gap identifies online child estimation as an important future
improvement direction.  It does not alter the legality of the deployed result.

\paragraph{Zero intercept, scaling, and superposition.}
Zero targets produce zero outputs in every information-flow audit.
For fixed operating-point operators, scaling the target by
$\alpha\in\{0,0.25,0.5,1,2\}$ scales the output with cosine at numerical unity
and relative errors at or below $1.7\times10^{-7}$ for ResNet18; ViT scaling is
exact to recorded precision.  Cross-target superposition has output cosine
$0.999999996$ for both models.  Weighted recomposition of a complete target
from channel partitions gives cosine 0.99999 for ResNet18 and 0.99950 for ViT.
After weighting for the public subset mean gauge, the remaining recomposition
residual reflects finite-precision application, not an additive image template.

\paragraph{Normalisation and residual gauge checks.}
The final path applies no source, activation, or cotangent pre-normalisation.
The $D_v$ source is the pulled-back local child residual in the same gauge used
by the online JVP, while its regression target is $H_v-y_v^0$.
Static checks validate the JVP direction, residual sign, stored
map layout, and zero-intercept application.  Runs that use true children,
alternative correction weights, or nondefault internal boundaries are labelled
diagnostic-only in their artifacts and are excluded from headline tables.

With the legal online path and its practical headroom fixed, the next section
expands the decision-grounded interpretation study without changing the fitted
maps or the inversion API.

\begin{figure*}[t]
  \centering
  \includegraphics[width=\textwidth]{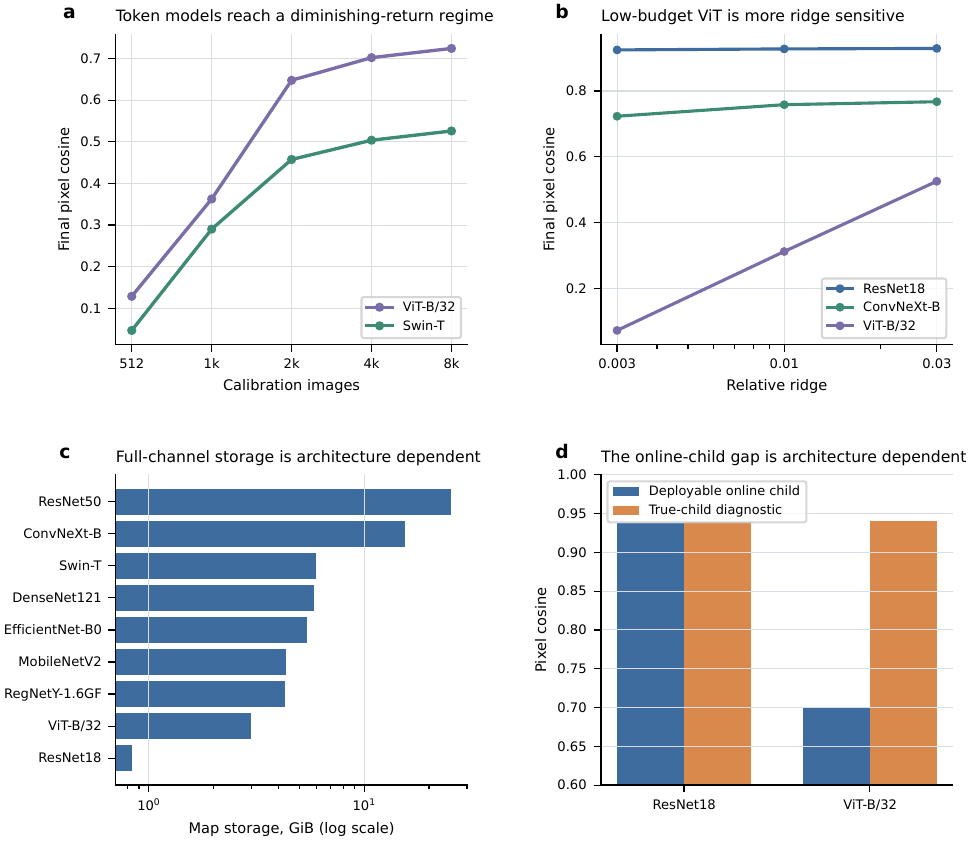}
  \caption{\textbf{Calibration and deployment diagnostics.}
  \textbf{a}, Paired Swin-T and ViT-B/32 calibration curves.
  \textbf{b}, Ridge sensitivity at the stated small-calibration budgets.
  \textbf{c}, Full $G_v+D_v$ storage across architecture families.
  \textbf{d}, Deployable online-child versus nondeployable true-child
  diagnostic.}
  \label{fig:supp_calibration_deployment}
\end{figure*}

Figure~\ref{fig:supp_calibration_deployment} summarises the token calibration
curves, ridge sensitivity, storage demand, and online-child oracle gap.

\subsection{Expanded Prediction-Conditioned Feature Atlas}
\label{app:decision_atlas}

\paragraph{Protocol.}
The decision-conditioned experiment uses CUB-trained ResNet18, ConvNeXt-B, and
ViT-B/32 checkpoints with top-1 accuracies 70.59\%, 85.81\%, and 70.16\%.
For each model, 16 correctly classified test images are fixed before feature
selection.  One c4096 map family is reused across four depths, all images, and
all selected channel groups.  The depths are 3/5/7/9 for ResNet18,
4/7/34/37 for ConvNeXt-B, and 2/6/10/14 for ViT-B/32.

For predicted class $\hat y$ and runner-up class $y_2$, channels are ranked by
their activation--gradient contribution to the decision margin
\citep{ancona2018towards},
\begin{equation}
  s_c
  =
  \left\langle H_{l,c},
  \frac{\partial(z_{\hat y}-z_{y_2})}{\partial H_{l,c}}
  \right\rangle .
\end{equation}
Nested groups of $k\in\{1,4,8,16\}$ top-positive, top-negative,
activation-matched, and 20-repeat random-matched channels are selected.
Each group is inverted through the legal subset gauge
$P_{S,Q_{\mathrm{all}}}H_l/|S|$.
In a parallel branch, the same channels are replaced in the original internal
representation.  Calibration-mean replacement is primary; zero replacement is
a robustness protocol.  The feature inverse is never re-input as a causal
stimulus.

\paragraph{Cross-architecture gate.}
The strict audit covers 36,864 intervention rows, 3,072 group inversions, and
1,536 deletion curves.  Under calibration-mean replacement, the top-positive
margin drop exceeds the within-image random-matched mean in all 48
model--layer--$k$ settings.  Every image-cluster bootstrap interval is above
zero.  Zero replacement gives the same 48/48 result.  All 12 terminal
top-negative controls have intervals below zero, demonstrating the expected
opposite signed effect.
Table~\ref{tab:supp_decision_gate} summarises the signed gate for each
architecture.

\begin{table*}[t]
  \centering
  \caption{\textbf{Decision-conditioned cross-architecture gate.}
  Counts give the number of settings whose bootstrap interval has the required
  sign.  The final column is the minimum--maximum prediction-flip fraction
  across the four tested depths at $k=16$ under calibration-mean replacement.}
  \label{tab:supp_decision_gate}
  \small
  \begin{tabular}{lrrrr}
    \toprule
    Model & Primary $>0$ & Zero $>0$ & Terminal negative $<0$
      & $k=16$ flip range \\
    \midrule
    ResNet18 & 16/16 & 16/16 & 4/4 & 0.313--0.813 \\
    ConvNeXt-B & 16/16 & 16/16 & 4/4 & 0.125--0.750 \\
    ViT-B/32 & 16/16 & 16/16 & 4/4 & 0.188--1.000 \\
    \bottomrule
  \end{tabular}
\end{table*}

Normalised positive-minus-random effects are positive at every tested depth and
group size.  They range from 0.122 to 2.253 on ResNet18, 0.201 to 9.559 on
ConvNeXt-B, and 0.068 to 1.805 on ViT-B/32.  The effect scale differs across
architectures, so the claim is based on within-image matched contrasts rather
than direct comparison of raw logit units.

\paragraph{Architecture-dependent inverse geometry.}
ResNet18 and ConvNeXt-B develop stronger deep common-mode inverse structure,
while positive and negative groups have opposing decision effects.
ViT-B/32 retains a patch-grid, low-frequency, globally mixed geometry.
The shared result across architectures is the
selection--inversion--intervention chain, not identical visible structure.
CUB segmentation and visible-part overlap remain secondary diagnostics and are
not used to support the cross-architecture gate.  They do not establish a
unique semantic name for every channel group.  Likewise, the CNN common-mode
observation is consistent with prior holographic-representation evidence
\citep{shu2026adjoint}, but does not establish destructive interference without
a dedicated signed cancellation experiment.

\subsection{Optional Generalisation and Application Extensions}
\label{app:optional_extensions}

\begin{figure*}[t]
  \centering
  \includegraphics[width=0.96\textwidth]{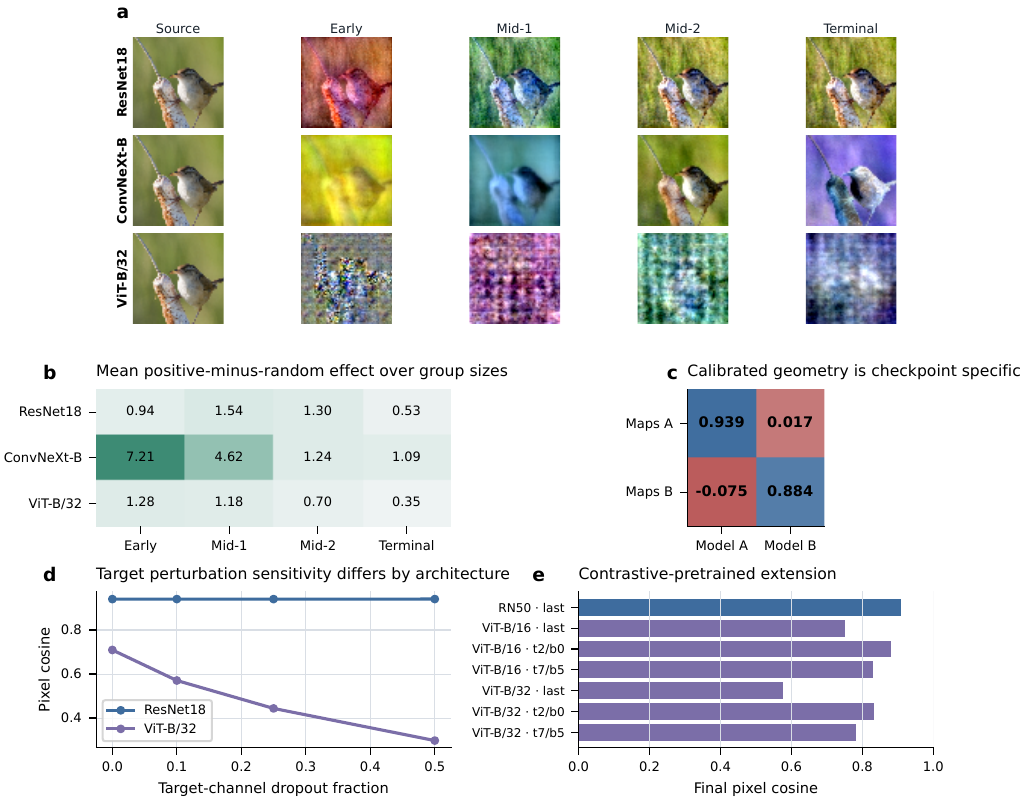}
  \caption{\textbf{Expanded decision-conditioned atlas and optional extensions.}
  \textbf{a}, One fixed CUB image across four depths and three architectures.
  \textbf{b}, Mean positive-minus-random intervention effect over group sizes.
  \textbf{c}, Same-architecture checkpoint/map mismatch.
  \textbf{d}, Channel-dropout sensitivity.
  \textbf{e}, Contrastive-pretrained CLIP extensions across architecture and
  depth.}
  \label{fig:supp_decision_extensions}
\end{figure*}

These experiments test additional boundaries after the core claim is already
established.  They are not required for the main paradigm comparison;
Figure~\ref{fig:supp_decision_extensions} summarises the decision-conditioned,
checkpoint, perturbation, and CLIP evidence.

\paragraph{Target perturbations.}
ResNet18 is nearly invariant to relative Gaussian target noise up to 0.1 and
to 8- or 16-bit quantisation.  Its inverse remains at pixel cosine 0.9406, and
output cosine to the clean inverse remains above 0.99999998 for Gaussian noise.
ViT-B/32 remains close under Gaussian noise and 8-bit quantisation, but is more
sensitive to channel dropout.  Dropping 50\% of target channels reduces pixel
cosine from 0.7094 to 0.2977 and output cosine to the clean inverse to 0.4083.
This contrast is consistent with the stronger channel-coupling demand observed
in the ViT operator ablation.

\paragraph{Checkpoint specificity.}
Maps fitted on one ResNet18 checkpoint transfer poorly to a different
same-architecture checkpoint.  Matched map/model conditions reach pixel
cosines 0.9388 and 0.8842, whereas mismatched conditions reach 0.0165 and
$-0.0754$.  The result confirms that calibrated maps encode checkpoint-specific
local geometry rather than an architecture-level image decoder.

\paragraph{Blind architecture onboarding.}
Automatic MLP-Mixer onboarding \citep{tolstikhin2021mlp} captures only the stem
projection rather than
the intended final Mixer norm.  The captured stem target itself is inverted
successfully, with final pixel cosine 0.9470, SSIM 0.8299, and LPIPS 0.0991.
The experiment is therefore a verified boundary-discovery failure, not a
closed-form inversion failure.  It defines a concrete engineering limitation:
new architectures require the boundary capture policy to expose the intended
target and reverse DAG.

\paragraph{Contrastive-pretrained models.}
The same estimator and tensor-axis rule extend to CLIP-RN50 and CLIP ViTs
\citep{radford2021learning}.
CLIP-RN50 reaches final pixel cosine 0.9102 at its last target.
CLIP-ViT-B/16 reaches 0.7519 at the last target and 0.8799 at an early block.
For CLIP-ViT-B/32, the best last-target result is 0.5774 with a finer
attention/MLP boundary, while early and middle targets reach 0.8338 and 0.7829.
Overly fine token-internal splits collapse, which reinforces the practical
importance of boundary selection.  These results broaden the architecture and
pretraining evidence but remain supplement-only because they do not change the
main claim.

\subsection{Language-Transformer Protocol and Full Evidence}
\label{app:language_extension}

\paragraph{Controlled benchmark.}
The formal language experiment uses pretrained GPT-2 small and the exact
instantiation in \cref{app:gpt2_method_instance}.
Each map family was calibrated on 4,096 fixed-seed WikiText-2 windows
\citep{merity2016pointer}, and the formal evaluation averaged independently
fitted seed-123 and seed-321 outputs before statistical resampling.
The semantic benchmark contained 72 contexts: 12 polysemous words, two senses
per word, and three contexts per sense.
Each context was deterministically padded and cropped to 16 GPT-2 tokens while
keeping the target word at the terminal position.
Semantic cues were specified before evaluation.
For every context, 64 random subsets matched both cue count and the empirical
cue-position distribution while excluding cue and target positions.

The syntactic benchmark contained 80 contexts from 20 lexical templates.
Within each template, singular/plural heads and grammatical/ungrammatical
auxiliaries were crossed while the nearer attractor noun carried the opposite
number.
All four variants shared the same token-position roles: head, attractor,
auxiliary, and terminal predicate.
The analysis asks whether the inverse routes through the agreement site rather
than the nearer attractor; it does not interpret the energy difference as a
probability of grammaticality.

\paragraph{Statistical resampling.}
The benchmark intervals used 10,000 percentile cluster-bootstrap resamples
after averaging the two calibration-map outputs.
Polysemy analyses resampled the 12 target-word clusters and retained all six
contexts for each sampled word.
Syntax analyses resampled the 20 lexical-template clusters and retained all
four number and grammaticality variants for each sampled template.
These intervals therefore quantify variation across words or templates rather
than treating contexts derived from one benchmark item as independent.  Layer
6 was selected as the cue-minus-random peak in the exploratory sweep below, so
the reported layer-6 intervals are conditional on that selected layer and do
not include uncertainty from the six-layer search.
Decision-intervention intervals separately resampled the 64 held-out WikiText
contexts.

\begin{figure*}[t]
  \centering
  \includegraphics[width=\textwidth]{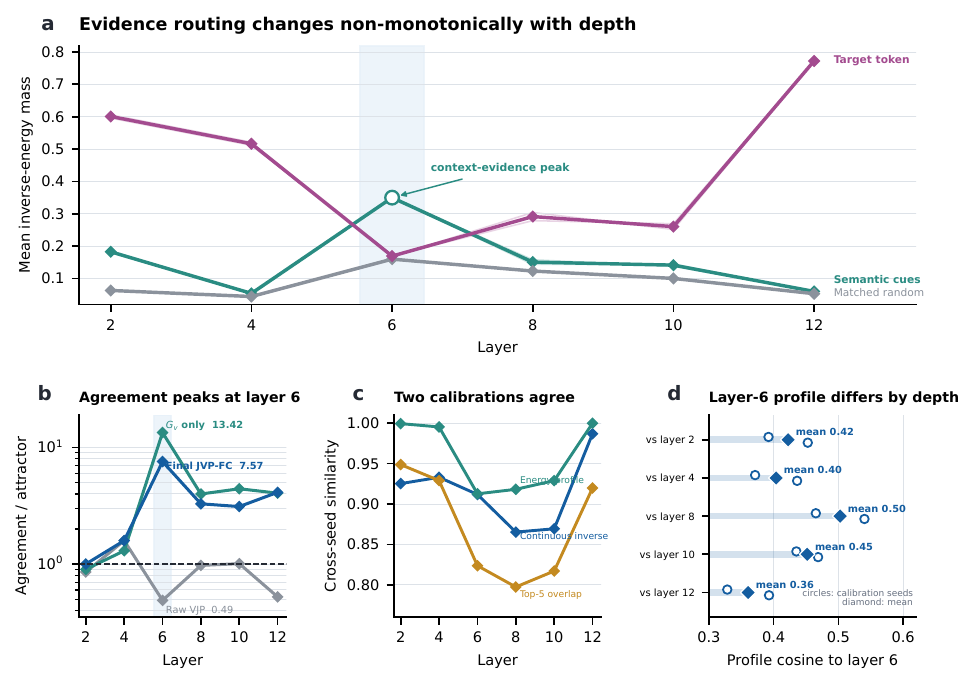}
  \caption{\textbf{Depth and calibration reproducibility of token-position inversion.}
  \textbf{a}, Target-token, semantic-cue, and position-matched-random mass for
  Final JVP-FC across layers 2, 4, 6, 8, 10, and 12.  Thin lines are independent
  c4096 calibrations and thick lines are their means.
  \textbf{b}, Grammatical agreement-to-attractor ratio for Raw VJP, $G_v$ only,
  and Final JVP-FC.
  \textbf{c}, Per-layer cross-seed continuous-inverse cosine,
  energy-profile cosine, and top-five evidence-position overlap.
  \textbf{d}, Layer-6 profile cosine relative to the other target depths.
  Circles are independent c4096 calibrations and diamonds are their means.}
  \label{fig:language_supp_layers}
\end{figure*}

\paragraph{Depth and calibration reproducibility.}
An exploratory sweep evaluated layer-2, 4, 6, 8, 10, and 12 attention outputs
under both independently fitted c4096 map families.
Both calibrations selected layer 6 as the cue-minus-random peak.
Across layers, mean continuous-inverse cosine was 0.915, energy-profile cosine
was 0.959, top-five evidence-position overlap was 87.3\%, and the
cue-minus-random depth-curve cosine was 0.999.
At layer 6, the corresponding cross-seed values were 0.912, 0.913, and 82.4\%.
The sweep did not support a monotonic lexical-to-semantic hierarchy: layer 2
remained target-token dominated, later layers reintegrated terminal evidence,
and layer-12 target mass reached 0.772 under the seed-123 sweep.
Figure~\ref{fig:language_supp_layers} reports the complete depth and
cross-calibration profiles.

\paragraph{Reference methods.}
Raw VJP, $G_v$ only, and Final JVP-FC share the same target, operating point,
boundaries, and terminal seed.
The direct global linear reference was fitted on 4,096 WikiText windows and
maps the complete layer-6 terminal feature directly to all 16 input embeddings;
it does not use the source-local forward graph.
Attention weights were averaged over heads at layer 6.
Gradient-times-input and integrated gradients used the squared norm of the
selected layer-6 attention feature as their scalar objective; integrated
gradients used 16 steps from a zero-embedding baseline
\citep{ancona2018towards,sundararajan2017axiomatic}.
These three quantities are attribution references, not feature-inversion
baselines.

\begin{table*}[t]
  \centering
  \caption{\textbf{Controlled token-routing summary averaged over two c4096 calibrations.}
  Cue-minus-random uses all 72 semantic contexts.  Agreement-to-attractor uses
  40 grammatical and 40 ungrammatical contexts.  Ratios are descriptive;
  inference on syntax uses the cluster-bootstrap agreement-minus-attractor
  differences reported in the text.}
  \label{tab:language_references}
  \scriptsize
  \setlength{\tabcolsep}{4.0pt}
  \begin{tabular}{lrrrr}
    \toprule
    Method & Cue $-$ random & Target mass & Gram. agr./attr. & Ungram. agr./attr. \\
    \midrule
    Raw VJP & 0.178 & 0.511 & 0.488 & 0.410 \\
    $G_v$ only & 0.161 & 0.147 & \textbf{13.42} & \textbf{9.55} \\
    Final JVP-FC & \textbf{0.190} & 0.169 & 7.57 & 4.82 \\
    Direct global linear & 0.004 & 0.053 & 0.870 & 0.869 \\
    Attention & 0.006 & 0.039 & 1.79 & 1.66 \\
    Gradient $\times$ input & 0.173 & 0.267 & 0.559 & 0.517 \\
    Integrated gradients & 0.088 & 0.216 & 0.742 & 0.729 \\
    \bottomrule
  \end{tabular}
\end{table*}

\paragraph{Paired stage comparisons.}
The scalar summaries in \cref{tab:language_references} should not be read as a
monotonic ranking of the two fitted stages.
Relative to Raw VJP, Final changed cue-minus-random by $+0.012$ (95\%
word-cluster bootstrap interval $-0.064$--0.097), increased cue mass by 0.091
(0.024--0.158), and reduced terminal-token mass by 0.342 (0.294--0.389).
Final also improved the pooled agreement-minus-attractor effect over Raw VJP by
0.046 (95\% template-cluster bootstrap interval 0.027--0.066).
However, the same effect was 0.031 lower than $G_v$ alone
(Final-minus-$G_v$ interval $-0.043$ to $-0.019$).
These comparisons support an interpretation in which $G_v$ and JVP-FC alter
different aspects of the inverse.

\begin{figure*}[t]
  \centering
  \includegraphics[width=\textwidth]{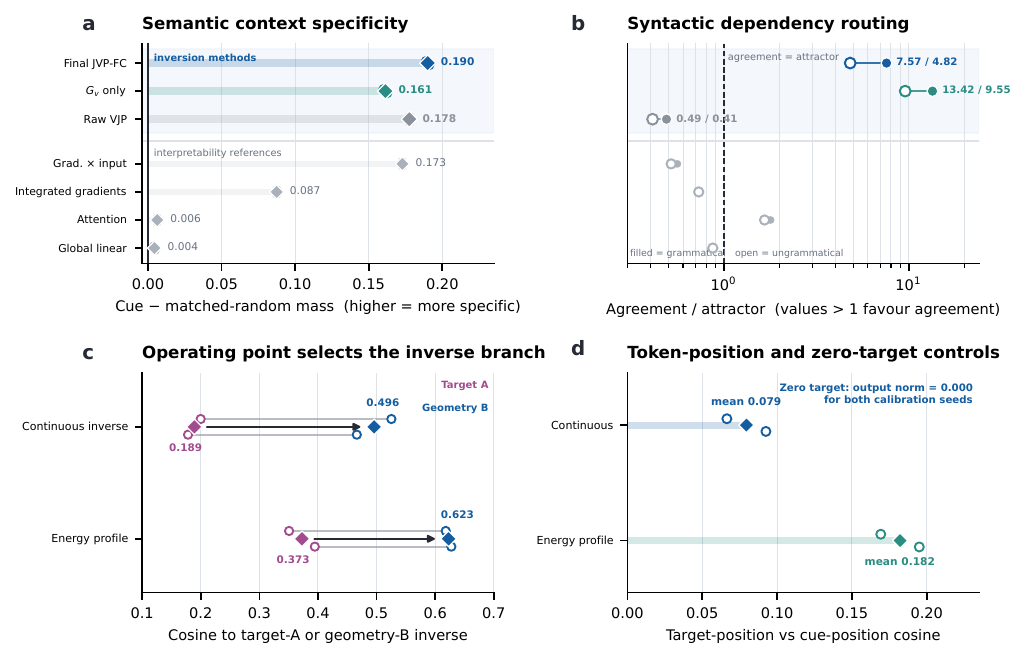}
  \caption{\textbf{Reference methods and source-local controls.}
  \textbf{a}, Semantic-cue mass above position-matched random controls for each
  inversion method and interpretability reference.  Small circles are the two
  c4096 calibrations and diamonds are their means.
  \textbf{b}, Grammatical and ungrammatical agreement-to-attractor ratios in
  the same method order as panel a.  Filled and open markers denote grammatical
  and ungrammatical contexts; the dashed line marks equal agreement and
  attractor mass.
  \textbf{c}, Terminal-state-$A$ inverses evaluated on source-local geometry
  $B$.  Paired seed points and mean arrows compare similarity to target-$A$ and
  geometry-$B$ inverses.
  \textbf{d}, Similarity between inverses of the terminal target position and a
  predefined cue position.  An exactly zero terminal state gives exactly zero
  output under both calibrated map families.}
  \label{fig:language_supp_references}
\end{figure*}

\begin{figure*}[t]
  \centering
  \includegraphics[width=\textwidth]{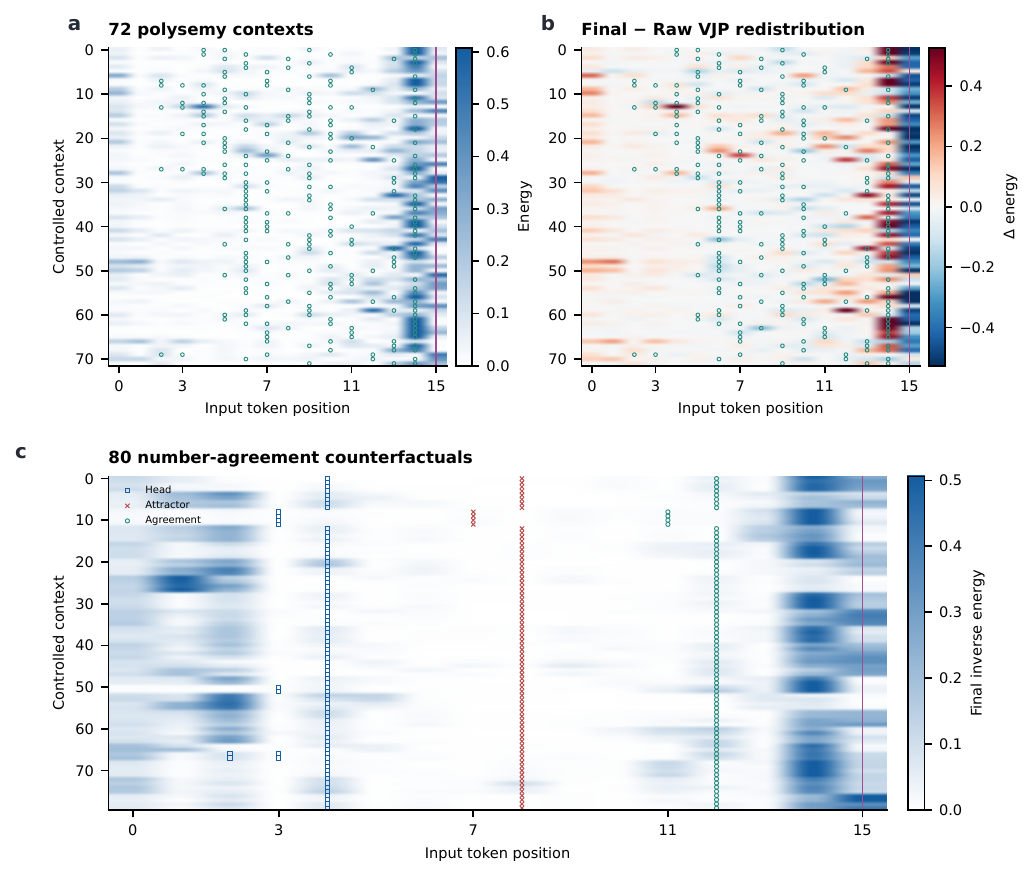}
  \caption{\textbf{Complete semantic and syntactic control matrices.}
  \textbf{a}, Final inverse energy for all 72 polysemy contexts.  Teal circles
  mark predefined cues and the magenta line marks the terminal target.
  \textbf{b}, Final-minus-Raw-VJP energy redistribution for the same contexts.
  \textbf{c}, Final inverse energy for all 80 number-agreement counterfactuals;
  blue squares, red crosses, and teal circles mark heads, attractors, and
  agreement positions, respectively.}
  \label{fig:language_supp_controls}
\end{figure*}

\paragraph{Source-local and query controls.}
For 24 fixed cross-context pairs, terminal state $A$ was evaluated on
operating-point geometry $B$.
Averaged over the two calibrations, the swapped continuous inverse was closer
to the geometry-$B$ inverse than to the target-$A$ inverse (0.496 versus
0.189); the corresponding profile cosines were 0.623 and 0.373.
Every zero terminal state produced an exactly zero output.
Replacing the terminal query with a predefined cue-position query also changed
the inverse substantially: target-versus-cue cosine was 0.079 for the
continuous inverse and 0.182 for the energy profile.
These controls separate source-local geometry, selected terminal content, and
selected token position.
Figure~\ref{fig:language_supp_references} places the reference-method and
source-local controls side by side, while
\cref{fig:language_supp_controls} reports every semantic and syntactic example.

\begin{figure*}[t]
  \centering
  \includegraphics[width=\textwidth]{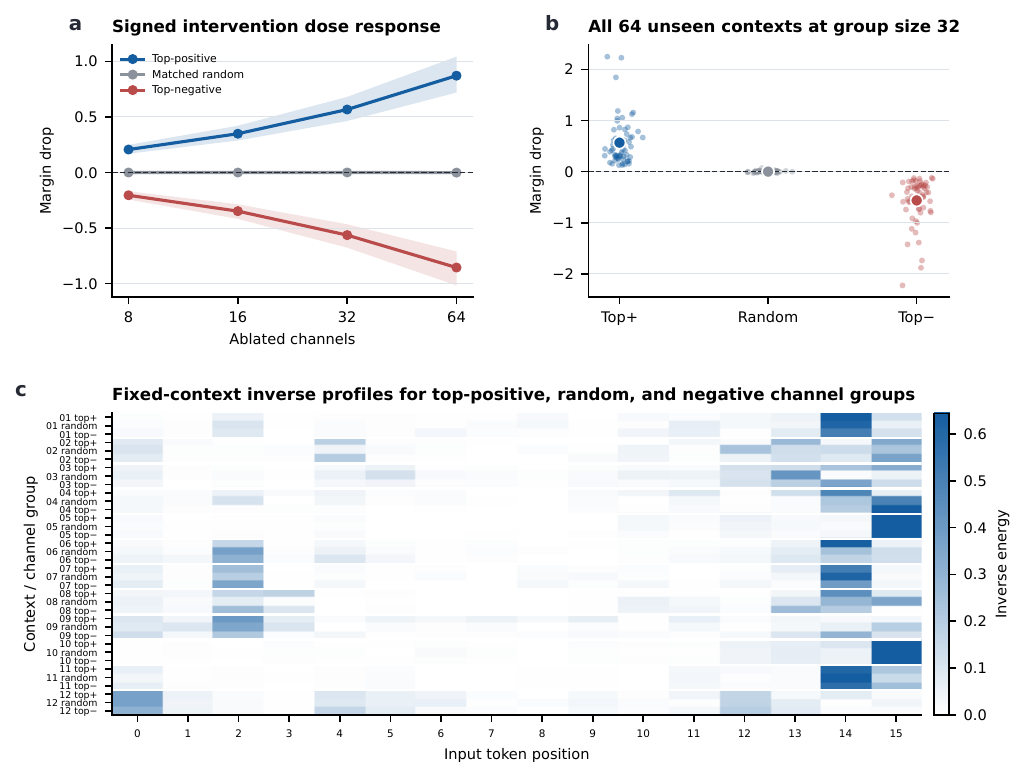}
  \caption{\textbf{Expanded decision-conditioned language analysis.}
  \textbf{a}, Mean signed intervention effect with 95\% context-bootstrap
  intervals over group sizes 8, 16, 32, and 64.
  \textbf{b}, All 64 context-level effects at group size 32.
  \textbf{c}, Fixed-context inverse-energy profiles for top-positive, median
  random, and top-negative channel groups.  Contexts are selected by fixed
  corpus order rather than effect size.}
  \label{fig:language_supp_decision}
\end{figure*}

\paragraph{Decision-conditioned protocol.}
Each decision context was a held-out 16-token WikiText prefix.
We computed the top1--top2 logit margin and ranked the 768 channels of the
layer-6 terminal attention output by activation times margin gradient.
Top-positive and top-negative groups contained the $k$ most positive and most
negative channels; each of 32 random controls sampled $k$ channels uniformly
without replacement.
Ablation set the selected attention-output channels to zero at the terminal
position.
The margin experiment therefore evaluates the feature group inside the original
model.
The corresponding inverse uses the fitted closed-form maps but is not fed back
into GPT-2.
At $k=32$, the top-positive mean margin drop was 0.567 (0.463--0.680), the
mean of the per-context random medians was 0.0001 ($-0.0033$--0.0038), and the
top-negative mean was $-0.564$.
Figure~\ref{fig:language_supp_decision} reports the complete group-size and
context-level analysis together with fixed-order inverse profiles.

\paragraph{Joint full-coordinate channel-subset query.}
\label{app:language_full_coordinate_query}
We next held a selected channel set fixed while changing its coordinate set.
The target remained the GPT-2 small layer-6 attention output with shape
$B\times16\times768$.  For each of 128 fixed WikiText-2 contexts, channels
were ranked by
\[
  \sum_{p=1}^{16}
  H_T[p,c]\,
  \frac{\partial(\ell_{\mathrm{top1}}-\ell_{\mathrm{top2}})}
       {\partial H_T[p,c]}.
\]
The 32 most positive and most negative channels formed the two
decision-conditioned groups.  Sixteen deterministic random groups excluded
these channels; the group with the median all-position margin drop was retained
as the representative random reference for that context.  The same selected
channels were queried either only at the terminal token,
$Q=\{L-1\}$, or jointly over the complete token stream,
$Q=Q_{\mathrm{all}}$.  The existing c4096 all-shared maps were read without
refitting.

The evaluation windows were sampled independently of calibration.  None of
their complete 17-token context-plus-next-token intervals overlapped a c4096
calibration interval, and the 128 evaluation intervals were pairwise
non-overlapping.  The corresponding independent intervention zeroed the same
channels either at the terminal position or at all 16 positions and recomputed
the terminal top-1--top-2 logit margin.

\begin{table*}[h!]
  \centering
  \caption{\textbf{GPT-2 joint full-coordinate channel-subset control.}
  Margin drop is measured by an independent intervention on the original
  attention-output representation.  Paired differences report
  $Q_{\mathrm{all}}-\{L-1\}$ as means with 95\% context-bootstrap intervals
  over 128 strictly disjoint evaluation windows.}
  \label{tab:gpt2_full_coordinate_query}
  \scriptsize
  \setlength{\tabcolsep}{8.0pt}
  \begin{tabular}{lccc}
    \toprule
    Channel group & Terminal-only margin drop & Full-coordinate margin drop
      & Paired difference \\
    \midrule
    Top positive & 0.447 & 0.677 & 0.230 (0.201--0.263) \\
    Top negative & $-0.442$ & $-0.675$ & $-0.233$ ($-0.270$--$-0.200$) \\
    Random median & 0.0019 & 0.0024 & 0.0006 ($-0.0030$--0.0042) \\
    \bottomrule
  \end{tabular}
\end{table*}

Expanding the top-positive query to $Q_{\mathrm{all}}$ increased the signed
margin-drop effect in all 128 contexts, whereas the top-negative group changed
in the opposite direction and the random contrast remained centred near zero
(\cref{tab:gpt2_full_coordinate_query}).  Holding $S$ fixed while changing
$Q$ also changed the inverse: the mean token-energy-profile cosine was 0.489
(95\% interval 0.444--0.535), the mean continuous-inverse cosine was 0.309,
and normalised profile entropy increased by 0.144 (0.116--0.170).  For the
top-positive group, terminal-position energy mass decreased descriptively from
0.202 to 0.112.  These measurements show that the joint full-coordinate target
distributes the recovered input-embedding support over the context rather than
reproducing the terminal-only query.  They do not rank the correctness of two
different targets.

\begin{figure*}[t]
  \centering
  \includegraphics[width=\textwidth]{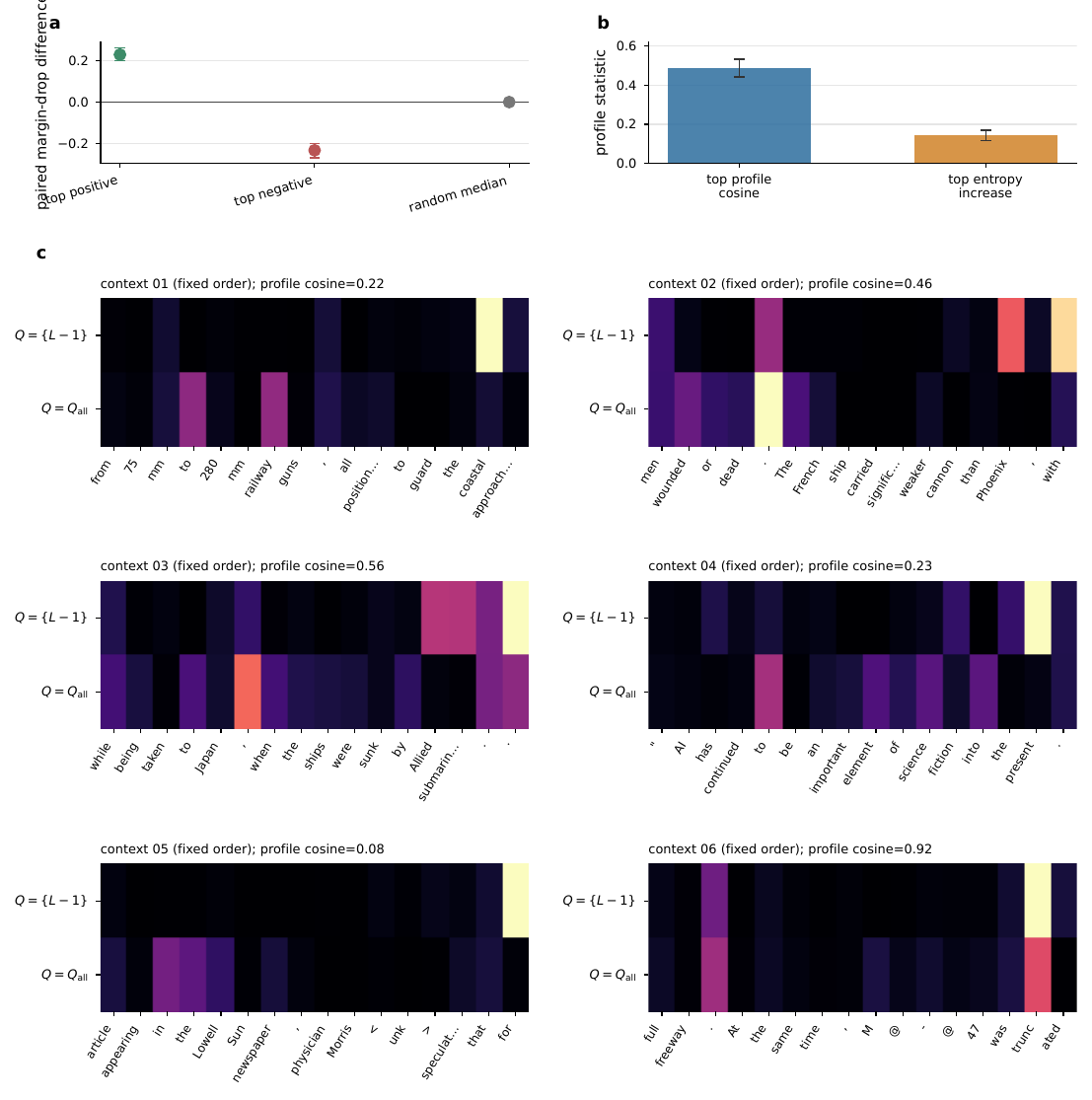}
  \caption{\textbf{GPT-2 terminal and joint full-coordinate channel-subset queries.}
  \textbf{a}, Paired intervention difference between $Q_{\mathrm{all}}$ and
  $Q=\{L-1\}$ for top-positive, top-negative, and representative random
  32-channel groups.  Points are means and bars are 95\% context-bootstrap
  intervals over 128 strictly disjoint WikiText-2 windows.
  \textbf{b}, Token-energy-profile cosine and entropy increase for the
  top-positive group, with the same interval definition.
  \textbf{c}, Terminal-only and joint full-coordinate inverse-energy profiles
  for the first six evaluation contexts in fixed corpus order.  Panels b--c use
  continuous input-embedding inverses and no text decoder; panel a reports an
  independent intervention on the original representation.}
  \label{fig:language_supp_full_coordinate}
\end{figure*}

\paragraph{Terminal-query scope.}
We additionally evaluated the complete $16\times16$ target-position matrix on
24 controlled contexts using the seed-123 map family without refitting.
Raw VJP and attention assigned zero mean inverse energy to positions later than
the queried token, consistent with the causal forward graph.
The deployed closed-form surrogate assigned mean future-position energy of
0.426 after $G_v$ and 0.427 after Final.
Because the effect was already present after $G_v$, it is not specific to
JVP-FC.
We therefore do not use the deployed surrogate for arbitrary isolated
nonterminal-position queries, and the formal semantic and syntactic routing
claims remain restricted to terminal-token targets, for which no future
positions exist.  The joint $Q_{\mathrm{all}}$ channel-subset control above
selects the complete token stream as one target and does not validate any
individual nonterminal position.
This diagnostic bounds the current deployment approximation; it does not alter
the terminal-token results or the general source-local DAG formulation.

\subsection{Continuous Non-Residual Inversion and the Display Interface}
\label{app:language_nonresidual}

The token-routing experiment uses no text renderer.
We separately tested whether the complete sequence of a genuinely transformed
non-residual feature could be inverted in the continuous embedding domain.
The target was the full layer-6 MLP-output sequence.
On 512 calibration-disjoint WikiText windows, mean input-embedding cosine was
0.030 for Raw VJP, 0.663 after $G_v$, and 0.683 after Final JVP-FC.
Independent Euclidean nearest-vocabulary lookup recovered the exact source
token at 2.3\%, 39.7\%, and 44.5\% of positions, respectively.
A calibration-centred matched-versus-cross re-encoding diagnostic gave AUC
0.526, 0.803, and 0.908.
Re-encoding is reported as a source-specific representation-consistency check,
not as the definition of a unique true inverse.

\begin{figure*}[t]
  \centering
  \includegraphics[width=\textwidth]{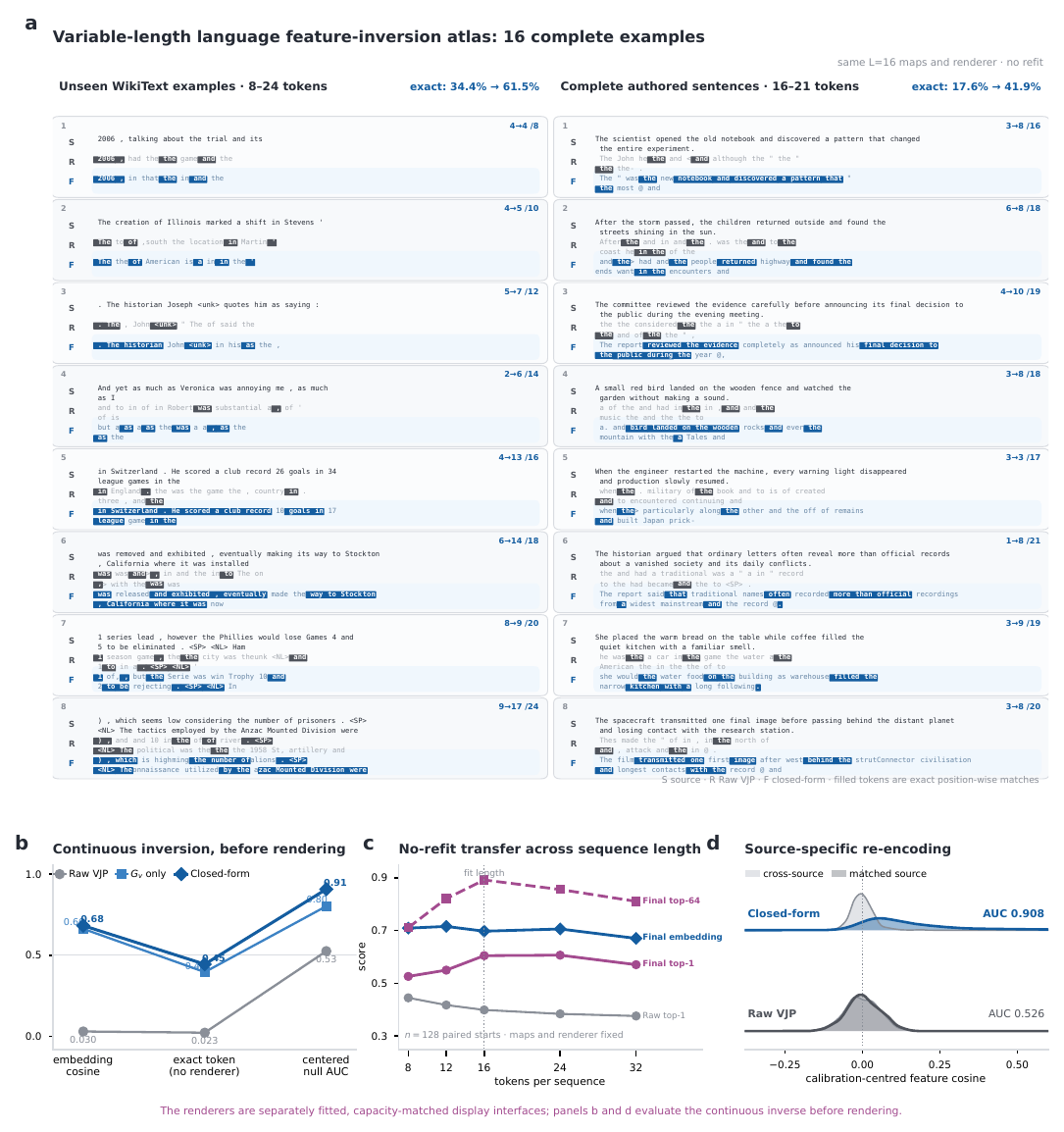}
  \caption{\textbf{Closed-form inversion of a non-residual GPT-2 feature.}
  \textbf{a}, Qualitative atlas over unseen WikiText and authored contexts using
  matched-capacity display interfaces.
  \textbf{b}, Continuous embedding cosine, pre-renderer exact-token recovery,
  and calibration-centred null AUC.
  \textbf{c}, No-refit transfer across sequence lengths 8--32.
  \textbf{d}, Calibration-centred matched- and cross-source re-encoding
  distributions.  All quantitative inversion and re-encoding metrics are
  computed before the display interface.}
  \label{fig:language_supp_nonresidual}
\end{figure*}

A zero-intercept rank-64 residual display interface contained 98,304 trainable
parameters while GPT-2 and the inverse maps remained frozen.
Under matched-capacity interfaces, Raw VJP reached 38.5\% top-1 and 67.3\%
top-64 token recovery, whereas Final reached 59.0\% and 88.1\%.
The same $L=16$ inverse maps and interface transferred without refitting to
lengths $L\in\{8,12,16,24,32\}$ by deterministic interpolation on the
normalised rFFT grid.
Final continuous embedding cosine remained between 0.670 and 0.716.
These experiments demonstrate reusable continuous inversion and provide a
readable display, but neither decoded fluency nor exact-token recovery is used
as evidence for the token-resolved interpretability claim.
Figure~\ref{fig:language_supp_nonresidual} separates the quantitative
continuous inverse from the optional display interface.

\paragraph{Supplementary boundary.}
The formal language claim is that a source-grounded closed-form inverse operates
on a canonical language Transformer and exposes controlled token evidence from
a selected internal representation.
The complete-state and display experiments broaden feasibility but do not
convert feature inversion into sentence generation.
Evidence outside GPT-2 small, substantially longer contexts, and other
language-model architectures is outside the present claim.
Isolated nonterminal-position queries are also excluded because the deployed
surrogate can allocate inverse energy to later positions, as disclosed above;
the joint full-coordinate channel-subset query does not remove this boundary.

\clearpage

\begingroup
\small
\setlength{\bibsep}{0pt plus 0.3ex}
\bibliographystyle{unsrtnat}
\bibliography{references}
\endgroup

\end{document}